\documentclass[twoside,11pt]{article}

\RequirePackage[titletoc]{appendix}
\RequirePackage{multicol}
\RequirePackage[utf8]{inputenc}
\RequirePackage{amsmath}
\RequirePackage{amsfonts}
\RequirePackage{bm}
\RequirePackage{titlesec}
\RequirePackage{marginnote}
\RequirePackage{mathtools}
\RequirePackage{algorithm}
\RequirePackage[noend]{algpseudocode}
\RequirePackage{caption}
\RequirePackage{subcaption}
\RequirePackage{booktabs}
\RequirePackage{multirow}
\usepackage[show]{chato-notes}
\usepackage{jmlr2e}

\newcommand{\spara}[1]{\smallskip\noindent{\bf #1}}

\newenvironment{squishlist}
{\begin{list}{$\bullet$}
  {\setlength{\itemsep}{0pt}
   \setlength{\parsep}{3pt}
   \setlength{\topsep}{3pt}
   \setlength{\partopsep}{0pt}
   \setlength{\leftmargin}{2em}
   \setlength{\labelwidth}{1.5em}
   \setlength{\labelsep}{0.5em} } }
{\end{list}}

\jmlrheading{}{}{}{7/22}{}{Gabriele D'Acunto, Paolo Di Lorenzo and Sergio Barbarossa}

\ShortHeadings{Multiscale Causal Structure Learning}{D'Acunto, Di Lorenzo and Barbarossa}
\firstpageno{1}

\begin{document}

\title{Multiscale Causal Structure Learning}

\author{\name Gabriele D'Acunto \email gabriele.dacunto@uniroma1.it \\
       \addr Department of Computer, Control, and Management Engineering\\
       Sapienza University of Rome\\
       Rome, 00185, Italy
       \AND
       \addr CENTAI\\
       Turin, 10138, Italy
       \AND
       \name Paolo Di Lorenzo \email paolo.dilorenzo@uniroma1.it \\
       \name Sergio Barbarossa \email sergio.barbarossa@uniroma1.it \\
       \addr Department of Information Engineering, Electronics,  and Telecommunications\\
       Sapienza University of Rome\\
       Rome, 00184, Italy}

\editor{tbd}

\maketitle

\begin{abstract}%
The inference of causal structures from observed data plays a key role in unveiling the underlying dynamics of the system. This paper exposes a novel method, named Multiscale-Causal Structure Learning (MS-CASTLE), to estimate the structure of linear causal relationships occurring at different time scales. Differently from existing approaches, MS-CASTLE takes explicitly into account instantaneous and lagged inter-relations between multiple time series, represented at different scales, hinging on stationary wavelet transform and non-convex optimization. MS-CASTLE incorporates, as a special case, a single-scale version named SS-CASTLE, which compares favorably in terms of computational efficiency, performance and robustness with respect to the state of the art onto synthetic data. We used MS-CASTLE to study the multiscale causal structure of the risk of 15 global equity markets, during covid-19 pandemic, illustrating how MS-CASTLE can extract meaningful information thanks to its multiscale analysis, outperforming SS-CASTLE. We found that the most persistent and strongest interactions occur at mid-term time resolutions. Moreover, we identified the stock markets that drive the risk during the considered period: Brazil, Canada and Italy. The proposed approach can be exploited by financial investors who, depending to their investment horizon, can manage the risk within equity portfolios from a causal perspective.
\end{abstract}


\section{Introduction}\label{sec:intro}
The inference of causal relationships plays a fundamental role in our understanding of complex systems. 
The ability to unravel causal structures from the observed data, also known as \emph{causal structure learning}, is an attractive technology that has received a growing attention in the last years, also thanks to the ever increasing volume of available data, see e.g., \cite{pearl2009causality}, \cite{peters2017elements}, \citet{glymour2019review}, and \cite{scholkopf2021toward}. 
The causal structure learning problem can be formalized as follows. Given $N$ random variables $\left\{ y_i \right\}_{i=1}^{N}$, the goal is to learn a proper \emph{directed acyclic graph} (DAG) such that the joint distribution factorizes in terms of \emph{causal modules} $P(y_i | \mathcal{P}_{i})$, i.e.,
\begin{equation*}
P(y_1,y_2,\ldots,y_N) = \prod_{i=1}^{N}P(y_i | \mathcal{P}_{i}),   
\end{equation*}
\noindent
where $\mathcal{P}_{i}$ denotes the set of \textit{parents} of variable $y_i$. In particular, the acyclicity requirement represents a necessary condition in order to set causes apart from effects; a result that cannot be accomplished in the presence of feedback loops among variables. 

If the random variables are sampled by a set of time series, the causal relations need to be consistent with time ordering. In such a case, in addition to lagged causal relationships, there might be instantaneous interactions among different time series that need to be carefully studied to unravel possible causal relations (\citealt{elcainf}). To be more specific, let us consider a data set $\mathbf{Y} \in \mathbb{R}^{T \times N}$ composed by $N$ time series of length $T$. Let $y_i[t]$ be the value assumed by the $i$-th time series at time $t$ and let $\mathcal{P}_{i,l}$ denote the set of parents of $y_i[t]$, with lag $l \in \mathbb{Z}^{+}$. We are interested in understanding whether the considered time series admits a \emph{functional representation} in which $y_i[t]$ depends on a set of parent variables, up to a finite lag $L$:
\begin{equation}\label{eq:funcrepr}
y_i[t]=f^i\left(\mathcal{P}_{i,L}, \ldots, \mathcal{P}_{i,0}, \epsilon_i[t]\right), \quad i \in \{1, \ldots, N \}
\end{equation}
where $\epsilon_i[t]$ represents either additive noise, statistically independent of the $i$-th time series, or a possible model mismatch, occurring at time $t$.
It is worth noticing that  the set of parents $\mathcal{P}_{i,l}$ can vary with $l$. To distinguish causes from effects, to be compliant with the causal inference problem, the set of Equations \eqref{eq:funcrepr} must admit a representation based on a DAG. 
However, as far as lagged interactions are concerned, since we cannot observe causal effects from present to past, $\mathcal{P}_{i,l}$ may contain $y_i[t-l]$, with $l>0$. 
In other words, time ordering provides lagged causal connections with implicit causal direction. However, when we look at instantaneous interactions, if we represent each $\mathcal{P}_{i, 0}$ over a graph, then the graph must be acyclic, otherwise it would be impossible to define the direction of the causal relation. 

\begin{figure}[t]
    \centering
    \begin{subfigure}[b]{.496\textwidth}
        \centering
        \includegraphics[width=\textwidth]{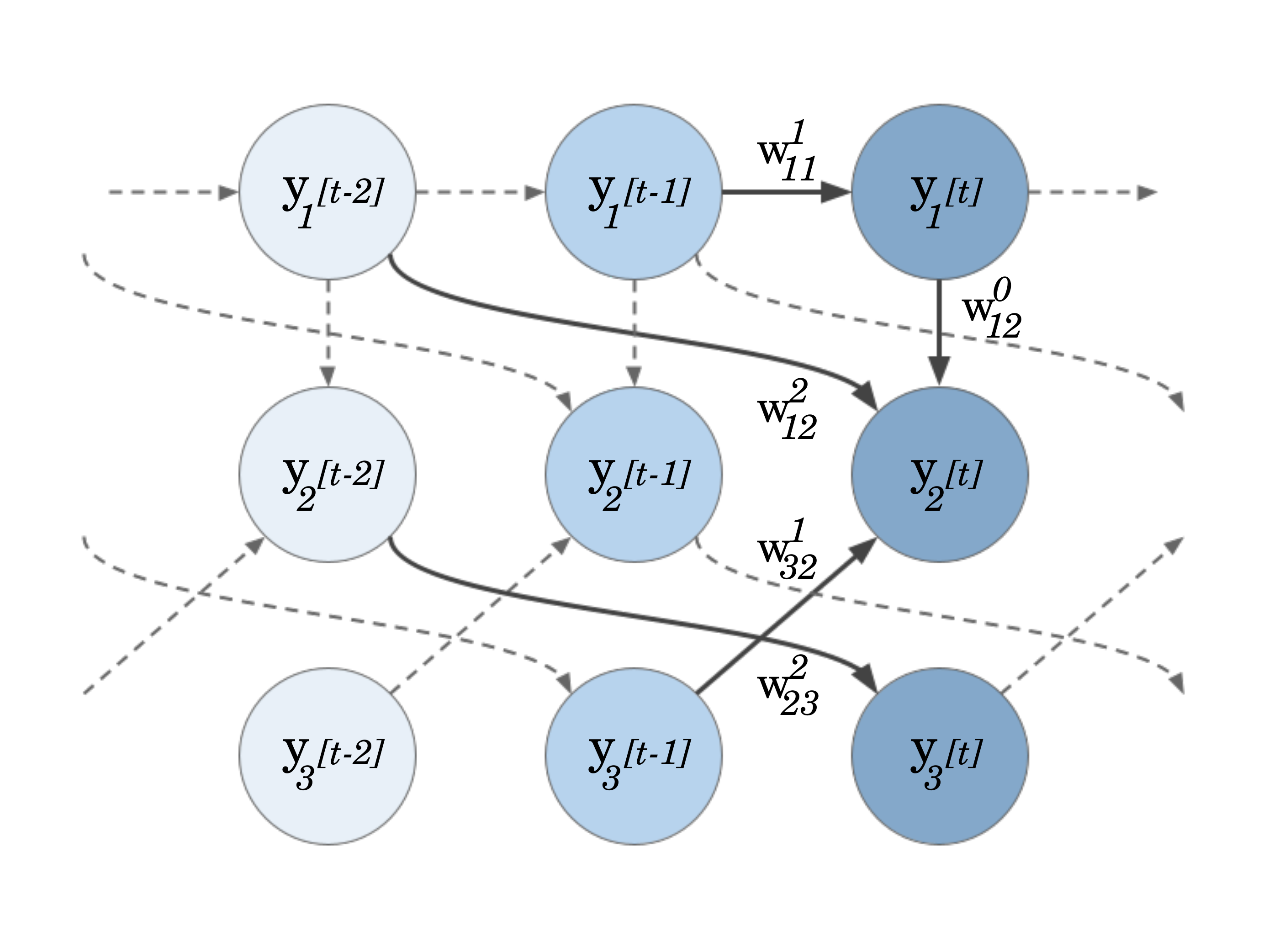}
        \caption{}
        \label{fig:SSCG}
    \end{subfigure}
    \hfill
    \begin{subfigure}[b]{.496\textwidth}
        \centering
        \includegraphics[width=\textwidth]{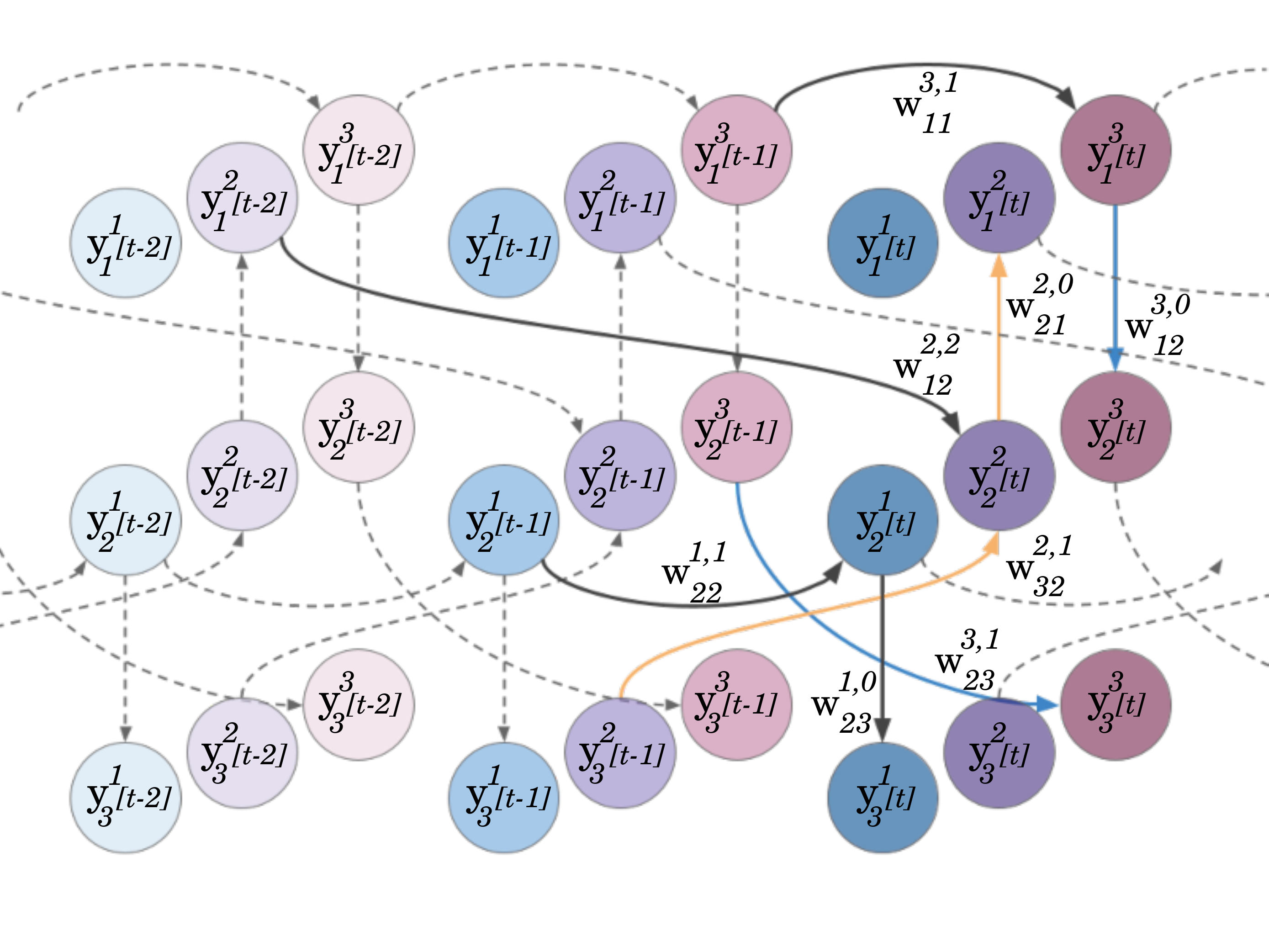}
        \caption{}
        \label{fig:MSCG}
    \end{subfigure}
    \caption{(a) SSCG with $N=3$ and maximum lag order $L=2$; (b) MSCG with $N=3$, $D=3$ time scales and maximum lag order $L=2$.}
    \label{fig:freqvstime}
\end{figure}

If we limit our attention to linear dependencies, the causal inference model can be expressed as
\begin{equation}\label{eq:SVARM}
   \mathbf{y}[t] = \sum_{l=0}^L \mathbf{y}[t-l]\mathbf{W}^{l} + \boldsymbol{\epsilon}[t],
\end{equation}
which coincides with the so called \emph{Structural Vector Autoregressive Model} (SVARM).
In Equation \eqref{eq:SVARM}, $\mathbf{y}[t]:=(y_1[t], \ldots, y_n[t])\in \mathbb{R}^{1 \times N}$ is the row vector containing the values assumed by $N$ time series, at time $t$, whereas
$\mathbf{W}^{l} \in \mathbb{R}^{N \times N}$, with $l=0, \ldots, L$, where $L$ is the maximum lag,  is the matrix representing the causal relation at lag $l$, so that $w^{l}_{ij} \neq 0$ if $y_i[t-l] \in \mathcal{P}_{j,l}$. In particular, $\mathbf{W}^{0}$ represents instantaneous interactions and its structure is such that, if we map the coefficients of $\mathbf{W}^{0}$ over the edges of a graph of size $N$, the resulting graph must be {\it acyclic}. Finally, $\boldsymbol{\epsilon}[t] \in \mathbb{R}^{1 \times N}$ is a row vector representing a random disturbance or model mismatch at time $t$. Equation \eqref{eq:SVARM} is said to be \emph{structural} since it allows us to express variables (effects) as linear functions of other endogenous variables (causes), by taking into consideration instantaneous as well as lagged relations, also referred to as {\it intra-} and {\it inter-layer} connections, respectively.
As an example, \Cref{fig:SSCG} shows the \emph{single-scale causal graph} (SSCG) associated to Equation \eqref{eq:SVARM}, in case of $N=3$ and $L=2$. In \Cref{fig:SSCG}, the subscript represents the node index while the time lag is given within the square brackets. As mentioned above, causal interactions occur from the past to the present and the instantaneous effects given at time $t$ do not involve any cycle. 

The model represented in Equation \eqref{eq:SVARM} and sketched in \Cref{fig:SSCG}, although very well known and applied, is however limited because it implicitly assumes that the time scale at which important dependencies show up is that associated to the observation task. However, in reality, there might be different dependencies occurring at different scales and, in general, there is no prior knowledge about the most suitable time scales to unravel important causal dependencies (\citealt{runge2019inferring}). Hence it makes sense to combine SVARM with a multiscale representation.
The goal of this paper is to propose a {\it linear causal inference algorithm based on a multiscale representation}, in order to capture the most relevant causal dependencies across multiple scales. More precisely, we propose a new methodology that infers the DAG describing linear causal relations among time series across multiple scales. 

The multiscale causal inference model can be built introducing a wavelet decomposition applied to the observed data set  $\mathbf{Y}$.  We used a  \emph{stationary wavelet transform} (SWT, \cite{nason1995stationary}). The wavelet decomposition is computed by letting the input signal to pass through a filterbank, where the output of each filter is associated to a different scale. Let us denote by $\mathbf{y}^d[t] \in \mathbb{R}^{N}$ the output vector representing the value assumed by the $N$ time series, at time $t$ and scale $d$, with $d=1, \ldots, D$.
Notice that the SWT returns non-decimated detail coefficients $\mathbf{y}^d[t]$ at each scale, where the latter is proportional to the variations of averages of the input signal over time resolution $2^{d-1}$ and is associated with the frequency interval $[1/2^{d+1}, 1/2^d]$~(\citealt{percival2000wavelet}). 
A key characteristic of the SWT is that, differently from the standard \emph{discrete wavelet transform} (DWT), it provides a  \emph{translation invariant} representation.
This is a useful property to capture relevant information present in the input data, without worrying about the position of the analysis time window.

Furthermore, our approach aims to retain both odd and even decimations at each decomposition level of the transformation without adding unnecessary redundancies.
To achieve this property, we use an orthogonal filters family, i.e., Daubechies wavelets, within the SWT framework, so that the time series ${y}_i^d[t]$ associated to different time scales are orthogonal to each other.
Moreover, since (i) the squared $\ell_{2}$ norm of the coefficients is equal to the energy of the data and (ii) the filters family is orthogonal, the variance of the input signal is preserved and it is partitioned across the scales (\citealt{percival2000wavelet}). 
According to our notation, $d=1$ is the finest scale while $d=D$ is the coarsest.
Putting $D$ different scale details in a single vector $\tilde{\mathbf{y}}[t]:=[\mathbf{y}^D[t], \mathbf{y}^{D-1}[t], \ldots, \mathbf{y}^1[t]]$, and stacking all these row vectors, for $t=1, \ldots, T,$ on top of each other, we build an augmented data set $\mathbf{\widetilde{Y}} \in \mathbb{R}^{T \times \bar{N}}$ (with $\bar{N}=DN$), where the row vector at timestamp $t$ contains the $t$-th details values of scale $d$, for all the $N$ input signals, at indexes $(d-1) \cdot N, \ldots, d \cdot N$. 
Then we build the block diagonal matrix $\mathbf{\widetilde{W}}^l:={\rm block}[\mathbf{W}^D_l, \mathbf{W}^{D-1}_l, \ldots,  \mathbf{W}^1_l]$ of size $\bar{N} \times \bar{N}$, where the $d$-th block $\mathbf{W}^{d}_l$ incorporates the causal interactions $w^{d,l}_{ij}$ occurring at the $d$-th scale with lag $l\geq0$ (up to a maximum lag $L$), and $w^{d,l}_{ij} \neq 0$ iff $y_{i}^d[t-l] \in \mathcal{P}_{j,l}^d$.
Here, $\mathcal{P}_{j,l}^d$ represents the parent set for time series $\mathbf{y}_j$ at lag $l$ and scale $d$, which in this case may vary along both graph layers and pages. 
The resulting MSCG is then modeled as follows:
\begin{equation}\label{eq:MSCG}
    \mathbf{\widetilde{y}}[t] = \sum_{l=0}^L \mathbf{\widetilde{y}}[t-l]\,\mathbf{\widetilde{W}}^{l} + \boldsymbol{\widetilde{\epsilon}}[t],
\end{equation}
where $\boldsymbol{\widetilde{\epsilon}}[t] \in \mathbb{R}^{1 \times \bar{N}}$ denotes the additive noise term.
In the above model, the matrix $\mathbf{\widetilde{W}}^0$ must respect the acyclicity requirement. 
As an example,  \cref{fig:MSCG} depicts a \emph{multiscale causal graph} (MSCG), in case of $N=3$ time series, $D=3$ time scales and maximum lag $L=2$. In \cref{fig:MSCG}, each layer refers to a specific time lag, whereas different pages refer to different time scales.
In terms of notation, in \cref{fig:MSCG}, the node superscript refers to the scale index (the page of the graph), while the subscript indicates the node index. The time lag, i.e., the layer of the graph, is given within the square brackets. Focusing on the causal interactions within each page, we observe both inter and intra-layer directed edges, where the latter, as in case of SSCG, must not involve any cycle. However, variables can interact differently at each time resolution. Therefore, we could observe reverse causal relationships between variables when looking across pages as, e.g., in the case of the blue and orange arcs given in \cref{fig:MSCG}. Finally, arcs between pages are not present due to the application of Daubechies orthogonal wavelets family.
\cref{fig:MSCG} represents then a multiscale DAG incorporating both instantaneous and lagged linear causal relations at different time scales: each page of the graph represents a SSCG at a certain time resolution.

The rest of the paper is organized as follows. In \Cref{sec:relw} we review the relevant state of the art on causal structure learning, with special attention to the analysis of time series and to the need of incorporating multiscale representations. 
Then, \Cref{sec:Contribution of this paper} highlights the main contributions of this paper. The problem formulation is formally stated in \Cref{sec:OptProb} and the proposed solution framework is described in \Cref{sec:MS-CASTLE}. 
The performance of the single scale version of the exposed method is evaluated in \Cref{sec:Numerical results} and compared to the state of the art algorithms suitable for solving Equation \eqref{eq:SVARM}.
Afterwards, we apply the developed multiscale causal model to unveil the multiscale causal structure of the risk of 15 global equity markets, during covid-19 pandemic. 
The results of this study, together with a comparison with those returned by the single scale approach, are given in \cref{sec:real-data}.
Finally, \cref{sec:conclusion} draws conclusions and sketches out future work directions.

\section{Related Work}\label{sec:relw}
Causal structure learning algorithms can be classified in accordance with the approach used to infer the associated DAG. 
In particular, we can identify three main different classes: 
(i) \emph{constraint-based approaches}, which run conditional independence tests to validate the presence of an edge between two variables (\citealt{spirtes2000, huang2020});
(ii) \emph{score-based methods}, that measure the goodness of fit of graphs according to a given criterion and then use search procedures to explore the solution space (\citealt{heckerman1995, chickering2002, huang2018generalized});
(iii) \emph{functional methods}, which model a variable in terms of a function of its parents (\citealt{lingam, hoyer, hyvarinen, peters2014, buhlmann2014cam}). 
Furthermore, a recent important contribution came by reformulating the problem of learning a DAG using a suitable continuous non-convex penalty (\citealt{zheng2018, zheng2020}). 
This enabled the usage of neural networks (\citealt{yu2019dag, lachapelle2019gradient}) and reinforcement learning (\citealt{zhu2019causal}) in causal discovery problems.

As pointed out in the previous section, whenever we deal with causal inference for time series analysis, we need to take into account time ordering as well. Considering linear models, this leads to the formulation of Equation \eqref{eq:SVARM}, which can be thought of a combination of a \emph{structural equation model} (SEM,~\cite{elcainf}) and a \emph{vector autoregressive model} (VAR,~\cite{sims1980macroeconomics}).
To estimate the parameters of Equation \eqref{eq:SVARM}, a stream of research assumes $\boldsymbol{\epsilon}_{t}$ to be non-normally distributed (\citealt{hyvarinen,moneta2013causal}). This allows us to apply \textit{independent component analysis} (ICA,~\citealt{hyvarinen1999fast}) to infer the causal structure from observations. Then, leveraging on non-convex optimization, DYNOTEARS showed promising results in the task of causal structure learning for time series (\citealt{pamfil2020dynotears}). 

All previous works refer to a single time scale representation. However, the introduction of  multiscale analysis is of paramount importance since it represents a key feature of complex systems, as shown in \citet{calvet2001forecasting, kwapien2012physical}. In particular, wavelet analysis, together with network analysis, has been already employed in the study of financial risk contagion~(\citealt{loh2013co, khalfaoui2015analyzing, wang2017stock}). More recently, the integration of machine learning methods and multiscale representations has been proved to provide significant advancements in biological and behavioral sciences, as reported in \citet{alber2019integrating} and \citet{peng2021multiscale}. However, the combination of multiscale representation with causal learning is still an open problem and this motivates our work in this paper, as detailed in the next section.

\section{Contribution of this Paper}
\label{sec:Contribution of this paper}
The main goal of this paper is to propose a {\it linear causal inference algorithm based on a multiscale representation},  able to capture the most relevant causal dependencies across multiple scales. 
More specifically, we expose an algorithm, termed \emph{Multiscale-Causal Structure Learning} (MS-CASTLE), enabling an efficient estimate of the causal matrices in Equation \eqref{eq:MSCG}, taking into account sparsity and acyclicity constraints. In particular, we formulate the optimization problem as non-convex, because of the acyclicity constraint and, subsequently, we derive an efficient algorithmic solution (i.e., MS-CASTLE) based on (a linearized version of) the \emph{Alternating Direction Method of Multipliers} (ADMM, \citealt{boyd2011distributed}). MS-CASTLE includes, as a particular case,  the single time-scale version, which we term \emph{Single Scale-Causal Structure Learning} (SS-CASTLE), to estimate the coefficients present in  Equation \eqref{eq:SVARM}, where only causality in the time domain is considered. The major contributions of this paper are the following.\\

\spara{Single-scale structure learning.} First of all, we compare SS-CASTLE with the following alternative techniques: (i) DYNOTEARS (\citealt{pamfil2020dynotears}), which shares with SS-CASTLE the same problem formulation, but it relies upon a different optimization procedure; (ii) two major linear non-Gaussian methods, i.e., VAR-ICALiNGAM and VAR-DirectLiNGAM (\citealt{hyvarinen, hyvarinen1999fast, 10.5555/1953048.2021040}). 
In particular, even if our SS-CASTLE method exploits the same dagness function proposed by \citet{zheng2018} to ensure the acyclicity property of the causal structure, differently from DYNOTEARS, we linearize the acyclicity constraint and we solve the resulting problem by leveraging on the \textit{Alternating Direction Method of Multipliers} (ADMM,~\citealt{boyd2011distributed}).
As shown in Section \ref{sec:Numerical results}, SS-CASTLE method outperforms DYNOTEARS and compares favorably with all other single-scale alternatives.
More specifically, the numerical results reported in  \Cref{sec:Numerical results} show that SS-CASTLE:
\begin{squishlist}
\item requires a lower computational cost than DYNOTEARS for solving Equation \eqref{eq:SVARM}, while preserving the inference accuracy;
\item provides better performance and greater robustness with respect to the considered linear non-Gaussian algorithms when we sample $\boldsymbol{\epsilon}[t]$ from a $p$-generalized normal distribution, with $p \in \left\{1, 1.5, 2, 2.5, 100  \right\}$. 
\end{squishlist}

\spara{Multi-scale structure learning.} Secondly, the major contribution of this paper is the proposal of a multiscale causal inference algorithm that allows the extraction of causal links at different scales, without requiring any prior knowledge of the scale where causal relations are most effective.
As highlighted in \Cref{sec:relw}, while the analysis of causality in the time domain has received significant attention during the past years, multiscale causal analysis has not been deeply investigated yet. We applied our methods to the analysis of financial systems, with focus on the inference of the graph representing inter-relations among different time series, a typical methodology used to study the risk spreading among financial institutions (\cite{bardoscia2021physics}). 
Our analysis provides novel results at both methodological as well as application levels, with respect to the current stream of research, known as \emph{Econophysics} (\citealt{mantegna1999introduction}).

At the methodological level, we propose a multiscale machine learning causal model that, differently from existing work~(\cite{billio2012econometric}), allows us to analyse both instantaneous and lagged causal interactions at distinct time scales. Furthermore, since the inference problem is non-convex, MS-CASTLE is only guaranteed to converge to stationary points. To check the robustness of the proposed method, we carried out a persistence analysis of the inferred causal relationships by varying the strength of the parameter used in the penalty term to enforce a sparse solution.

At the application level, we apply MS-CASTLE to infer the causal dynamics of risk contagion among $15$ global equity markets during covid-19 pandemic, from January $2020$, the 2nd to April $2021$, the 30th, rather than focusing on financial institutions.
As far as the risk measurement is concerned, in \cref{sec:resultsMS} we show the multiscale causal structure inferred from the previous risk series, made only by highly persistent edges (see \cref{sec:methodology}), and we compare the resulting graph with the graph obtained by the estimation of Equation \eqref{eq:SVARM}, presented in \cref{sec:resultsTS}. 

The results of our multiscale analysis show that:
\begin{squishlist}
\item causal connections are characterized by positive weights and are denser at mid-term time resolution (scales 3 and 4, i.e., 8-16 and 16-32 days, respectively);
\item the strongest connections are lagged and they appear at scale 3 and 4;
\item the markets injecting the majority of risk in the network are Brazil, Canada and Italy;
\item the multiscale approach provides information regarding the causal structure of the system that cannot be understood by looking only at the estimated SSCG.
\end{squishlist}
Further discussions concerning the obtained results and the richness of information gained through the multiscale causal analysis are given in \cref{sec:discussion}. 

\section{Problem Formulation}\label{sec:OptProb}
Let us consider as input a data set $\mathbf{Y} \in \mathbb{R}^{T \times N}$ constituted by 
sequences of length $T= 2^{D}$, for some integer value of $D \in \mathbb{N}$. 
Using the SWT, we decompose each of the $N$ time series ${y}_i[t] \in \mathbb{R}^{T}$ in $D$ non-decimated detail coefficients, as mentioned in \cref{sec:intro}.
We store the resulting samples $\{y^d_i[t]\}^{i=\{1,\ldots,N\}}_{d=\{1,\ldots,D\}}$ in an augmented data set $\mathbf{\widetilde{Y}} \in \mathbb{R}^{T \times \bar{N}}$, where the row vector at timestamp $t$ contains the $t$-th details values of scale $d$ for all the $N$ input signals from index $(d-1) \cdot N$ to $d \cdot N$.

Our goal is to estimate the causal matrices $\mathbf{\widetilde{W}}^{l}$, $l=0,\ldots,L$, in Equation \eqref{eq:MSCG}.
To ensure the acyclicity of the estimated MSCG, the inferred matrices of causal effects $\mathbf{\widetilde{W}}^{l}$, $l=0,\ldots,L$, must entail a DAG. However, learning DAGs from observational data is a combinatorial problem and, without any restrictive assumption, it has been shown to be NP-hard (\citealt{chickering2004large}). In our case, since we cannot observe edges coming from the present to the past, lagged causal relationships encompassed in the matrices $\mathbf{\widetilde{W}}^{l}$ (with $l>0$) are acyclic by definition. Therefore, the main issue concerns the inference of the matrix $\mathbf{\widetilde{W}}^{0}$ representing instantaneous causal effects. To handle the acyclicity of $\mathbf{\widetilde{W}}^{0}$, similarly to DYNOTEARS, we exploit the \textit{dagness} matrix function $h(\mathbf{M}):\mathbb{R}^{N \times N} \rightarrow \mathbb{R}$ proposed by \citet{zheng2018}, who proved that a matrix $\mathbf{\widetilde{W}}^{0}$ of size $\bar{N} \times \bar{N}$ can be represented as a DAG {\it if and only if} 
\begin{equation}
    h(\mathbf{\widetilde{W}}^{0})=\textrm{Tr}\left(e^{\mathbf{\widetilde{W}}^{0} \circ \mathbf{\widetilde{W}}^{0}}\right)-\bar{N} = 0,
\end{equation}
where $\circ$ represents the Hadamard product.
Using this function as a penalty term, we are now able to formulate the learning task as a continuous, albeit non-convex, optimization problem. 

First, let us introduce the matrix $\mathbf{\bar{Y}} \coloneqq [\mathbf{\widetilde{Y}}^{0}, \mathbf{\widetilde{Y}}^{1}, \ldots, \mathbf{\widetilde{Y}}^{L} ] \in \mathbb{R}^{T \times V}$ (with $V=\bar{N}(L+1)$) containing the matrices of $l$-shifted observations $\mathbf{\widetilde{Y}}^{l} \in \mathbb{R}^{T \times \bar{N}}$.
Similarly, we build $\mathbf{\bar{W}} \coloneqq [\mathbf{\widetilde{W}}^{0^{T}}, \mathbf{\widetilde{W}}^{1^{T}}, \ldots, \mathbf{\widetilde{W}}^{L^{T}}]^{T} \in \mathbb{R}^{V \times \bar{N}}$.
For convenience, let us indicate with $\mathbb{\bar{B}}$ the set of matrices having the same structure as $\mathbf{\bar{W}}$, i.e., made up of stacked block diagonal matrices.

Then, the proposed multiscale causal structure learning problem is mathematically cast as
\begin{equation}\label{eq:OptMSCASTLE}
    \begin{aligned}
        &\min_{\mathbf{\bar{W}} \in \mathbb{\bar{B}}} \quad  \dfrac{1}{2} || \mathbf{\widetilde{Y}} - \mathbf{\bar{Y}}\mathbf{\bar{W}}||_{F}^{2} + \lambda ||\mathbf{\bar{W}}||_{1}\\
       &\textrm{subject to}  \quad  h(\mathbf{\widetilde{W}}^{0})=\textrm{Tr}\left(e^{\mathbf{\widetilde{W}}^{0} \circ \mathbf{\widetilde{W}}^{0}}\right)-\bar{N} = 0,
    \end{aligned}
\end{equation}
where the subscript $F$ stands for Frobenius norm. The $\ell_1$ norm penalty is used in Problem \eqref{eq:OptMSCASTLE} to enforce sparsity of the aggregated causal matrix $\mathbf{\bar{W}}$, with a tunable parameter $\lambda>0$. 
Here, differently from already proposed ICA-based estimation procedures~(\citealt{hyvarinen, moneta2013causal}), the matrices of causal coefficients are learnt simultaneously. Despite the convexity of the objective function, Problem \eqref{eq:OptMSCASTLE} is non-convex due to the presence of the acyclicity constraint $h(\mathbf{\widetilde{W}}^{0}) = 0$. In the next section, we will derive an efficient method to solve  Problem \eqref{eq:OptMSCASTLE}.

\section{The MS-CASTLE Algorithm}\label{sec:MS-CASTLE}
To find a local solution of Problem \eqref{eq:OptMSCASTLE}, we exploit the computational efficiency of ADMM (\cite{boyd2011distributed}). In particular, we recast Problem \eqref{eq:OptMSCASTLE} in the following equivalent manner, introducing the auxiliary matrix $\mathbf{Z} \in \mathbb{R}^{V \times \bar{N}}$, and obtaining
\begin{equation}\label{eq:ADMM0}
    \begin{aligned}
        &\min_{\mathbf{\bar{W}} \in \mathbb{\bar{B}}} \quad  \dfrac{1}{2} || \mathbf{\widetilde{Y}} - \mathbf{\bar{Y}}\mathbf{\bar{W}}||_{F}^{2} + \lambda ||\mathbf{Z}||_{1}\\
        &\textrm{subject to} \quad  h(\mathbf{\widetilde{W}}^{0})=\textrm{Tr}\left(e^{\mathbf{\widetilde{W}}^{0} \circ \mathbf{\widetilde{W}}^{0}}\right)-\bar{N} = 0,\\
        &\qquad \qquad \quad\; \mathbf{\bar{W}}-\mathbf{Z}=\mathbf{0}_{V \times \bar{N}}\text{ .}
    \end{aligned}
\end{equation}
Now, following the scaled ADMM approach (\cite{boyd2011distributed}), and letting $\alpha$ and $\boldsymbol{\beta}$ be the Lagrange multipliers associated with the equality constraints of Problem \eqref{eq:ADMM0}, we introduce the following \emph{augmented lagrangian} (AUL) function
\begin{equation}\label{eq:AUL}
    \begin{aligned}
        \mathcal{L}_{\rho}\left(\mathbf{\bar{W}}, \mathbf{z}, \alpha, \boldsymbol{\beta}\right) = & \dfrac{1}{2} || \mathbf{\widetilde{Y}} - \mathbf{\bar{Y}}\mathbf{\bar{W}}||_{F}^{2} + \alpha h\left( \mathbf{\widetilde{W}}^0 \right) + \lambda ||\mathbf{z}||_{1} + \dfrac{\rho}{2} ||\mathbf{\bar{w}}-\mathbf{z} + \boldsymbol{\beta}||_2^{2}-\dfrac{\rho}{2}\|\boldsymbol{\beta}\|^2,
    \end{aligned}
\end{equation}
where $\mathbf{\bar{w}}=vec(\mathbf{\bar{W}})\in \mathbb{R}^{V\bar{N}}$,  $\mathbf{z}=vec(\mathbf{Z})\in \mathbb{R}^{V\bar{N}}$, and $\rho>0$ is a tunable positive coefficient. The ADMM algorithm proceeds by iteratively minimizing the AUL function with respect to the primal variables $\mathbf{\bar{W}}$,  $\mathbf{z}$, while maximizing it with respect to the dual variables $\alpha$ and $\boldsymbol{\beta}$. However, while the AUL function is strongly convex w.r.t. $\mathbf{z}$, and naturally concave w.r.t. $\alpha$ and $\boldsymbol{\beta}$, it is non-convex w.r.t. $\mathbf{\bar{W}}$, due to the presence of the non-convex dagness function $h\left( \mathbf{\widetilde{W}}^0 \right)$. To handle this non-convexity issue, following the idea of linearized ADMM methods (\cite{yang2013linearized, goldfarb2013fast}), we substitute the non-convex dagness function $h\left( \mathbf{\widetilde{W}}^0 \right)$ in the AUL with its linearization around the current value $\mathbf{\widetilde{W}}_k^0$ assumed at each iteration $k$, i.e., 
\begin{equation}\label{eq:linear}
    \overline{h}\left( \mathbf{\widetilde{W}}^0; \mathbf{\widetilde{W}}_k^0\right)  = h\left( \mathbf{\widetilde{W}}_k^0 \right) + \textrm{Tr}\left(G^T(\mathbf{\widetilde{W}}_k^0) (\mathbf{\widetilde{W}}^0-\mathbf{\widetilde{W}}_k^0) \right),
\end{equation}
where $G(\mathbf{\widetilde{W}}^0)$ represents the matrix-gradient of function $h\left( \mathbf{\widetilde{W}}^0 \right)$. Then, substituting Equation \eqref{eq:linear} into $h\left( \mathbf{\widetilde{W}}^0 \right)$, we obtain the following approximated AUL:
\begin{equation}\label{eq:AUL2}
    \begin{aligned}
        \overline{\mathcal{L}}_{\rho}\left(\mathbf{\bar{W}}, \mathbf{z}, \alpha, \boldsymbol{\beta}; \mathbf{\widetilde{W}}_k^0\right) = &\; \dfrac{1}{2} || \mathbf{\widetilde{Y}} - \mathbf{\bar{Y}}\mathbf{\bar{W}}||_{F}^{2} + \alpha h\left( \mathbf{\widetilde{W}}_k^0 \right) + \alpha \textrm{Tr}\left(G^T(\mathbf{\widetilde{W}}^0)(\mathbf{\widetilde{W}}^0-\mathbf{\widetilde{W}}_k^0)  \right) \\
        &+ \lambda ||\mathbf{z}||_{1}+ \dfrac{\rho}{2} ||\mathbf{\bar{w}}-\mathbf{z} + \boldsymbol{\beta}||_2^{2}-\dfrac{\rho}{2}\|\boldsymbol{\beta}\|^2,
    \end{aligned}
\end{equation}
which is now strongly convex w.r.t. $\mathbf{\widetilde{W}}^0$, while preserving the first-order optimality conditions of the AUL in Equation \eqref{eq:AUL} around the current approximation point $\mathbf{\widetilde{W}}_k^0$. As a result, any point satisfying the Karush-Kuhn-Tucker (KKT) conditions using the approximated AUL in Equation \eqref{eq:AUL2}, satisfies also the KKT conditions of the original Problem \eqref{eq:ADMM0}. Hinging on this fact, we now apply ADMM to the approximated AUL in Equation \eqref{eq:AUL2}. Then, letting $\mathbf{\widetilde{W}}_k^0$, $\mathbf{z}_k$, $\alpha_k$, and $\boldsymbol{\beta}_k$ be the current guesses of the primal and dual variables at time $k$, we obtain the following set of recursions:
\begin{equation}\label{ADMM}
    \begin{aligned}
        \mathbf{\bar{W}}_{k+1} = &\; \argmin_{\mathbf{\bar{W}} \in \mathbb{\bar{B}}} \dfrac{1}{2} || \mathbf{\widetilde{Y}} - \mathbf{\bar{Y}}\mathbf{\bar{W}}||_{F}^{2} + \alpha_{k} \textrm{Tr}\left(G^T(\mathbf{\widetilde{W}}_k^0) \mathbf{\widetilde{W}}^0 \right)  + \dfrac{\rho}{2} || \mathbf{\bar{w}}-\mathbf{z}_k + \boldsymbol{\beta}_{k}||_2^{2}\\
        \mathbf{z}_{k+1}  = &\; \argmin_{\mathbf{z}} \lambda ||\mathbf{z}||_{1} + \dfrac{\rho}{2} ||\mathbf{\bar{w}}_{k+1}-\mathbf{z} + \boldsymbol{\beta}_k||_2^{2}\\
        \alpha_{k+1} = &\; \alpha_{k} + \gamma\, h\left(\mathbf{\widetilde{W}}_{k+1}^0 \right)\\
        \boldsymbol{\beta}_{k+1} = &\; \boldsymbol{\beta}_{k} + \mathbf{\bar{w}}_{k+1}-\mathbf{z}_{k+1}
    \end{aligned}
\end{equation}
The first step in  \eqref{ADMM} is the minimization of a strongly convex quadratic function, subject to structure constraints $\mathbf{\bar{W}} \in \mathbb{\bar{B}}$, i.e., simple linear constraints on the elements of $\mathbf{\bar{W}}$. We perform this minimization using the L-BFGS-B algorithm (\cite{byrd1995limited}), i.e., a variation of the Limited-memory Broyden–Fletcher–Goldfarb–Shanno method that handles box constraints. The second step in Procedure \eqref{ADMM} can instead be computed in closed form as (\cite{boyd2011distributed})
\begin{equation}
\mathbf{z}_{k+1} = \mathcal{S}_{\left( \lambda/\rho \right)}\left(\mathbf{\bar{w}}_{k+1} + \boldsymbol{\beta}_{k}\right),
\end{equation}
where $\mathcal{S}_{\delta}(x)= {\rm sign}(x)\cdot \max(x-\delta,0)$ is the soft-thresholding function, used to enforce sparsity of the causal matrix representations. The third step in Procedure \eqref{ADMM} performs a gradient ascent step to maximize Function \eqref{eq:AUL2} with respect to $\alpha$, using a (possibly time-varying) step-size $\gamma$. Similar arguments then hold for the fourth step of Procedure \eqref{ADMM}. All the steps are then summarized in Algorithm 1, which we term as MS-CASTLE.
\begin{algorithm}[t]
\caption{MS-CASTLE}\label{algo:Algo1}
\begin{algorithmic}[1]
    \Procedure{MS-CASTLE}{$\mathbf{Y}, L, \lambda, \rho, \gamma, r, t, \gamma^{\text{max}}, \text{maxiter}$}
    \State $\mathbf{\widetilde{Y}} \gets$ Apply SWT to $\mathbf{Y}$
    \State $\mathbf{\bar{Y}} \gets [\mathbf{\widetilde{Y}}^{0}, \mathbf{\widetilde{Y}}^{1}, \ldots, \mathbf{\widetilde{Y}}^{L} ]$
    \State Initialize $\mathbf{\bar{W}}, \mathbf{Z}, \alpha, \boldsymbol{\beta}$ 
    \While{k $<$ maxiter $\And h_{k}>t$}
        \State Find $\mathbf{\bar{W}}_{k+1}$ using the L-BFGS-B algorithm to solve
        \State $\;\;\mathbf{\bar{W}}_{k+1} =  \displaystyle\argmin_{\mathbf{\bar{W}} \in \mathbb{\bar{B}}} \dfrac{1}{2} || \mathbf{\widetilde{Y}} - \mathbf{\bar{Y}}\mathbf{\bar{W}}||_{F}^{2} + \alpha_{k} \textrm{Tr}\left(G^T(\mathbf{\widetilde{W}}_k^0) \mathbf{\widetilde{W}}^0 \right)  + \dfrac{\rho}{2} || \mathbf{\bar{w}}-\mathbf{z}_k + \boldsymbol{\beta}_{k}||_2^{2}$
        \State $h_{k+1} \gets h\left(\mathbf{\bar{W}}_{k+1}\right)$
        \If{$h_{k+1}/h_{k}>r$}
            \State $\gamma \gets 10 \cdot \gamma$ \Comment{$\gamma \in \left(0, \gamma^{\text{max}} \right)$}
        \EndIf
        \State $\mathbf{z}_{k+1} \gets \mathbf{S}_{\left( \lambda/\rho \right)}\left(\mathbf{\bar{w}}_{k+1} + \boldsymbol{\beta}_{k}\right)$\Comment{Soft-thresholding operator}
        \State $\alpha_{k+1} \gets \alpha_{k} + \gamma\cdot h_{k+1}$
        \State $\boldsymbol{\beta}_{k+1} \gets \boldsymbol{\beta}_{k} + \mathbf{\bar{w}}_{k+1}-\mathbf{z}_{k+1}$
    \EndWhile
    \State \textbf{return} $\mathbf{\bar{W}}$
    \EndProcedure
\end{algorithmic}
\end{algorithm}

\textbf{Remark:} Algorithm \ref{algo:Algo1} can be easily customized to solve Equation \eqref{eq:SVARM}, where we simply ignore the multiresolution analysis. With regards to Algorithm 1, it simply means to skip line 2. This leads to the aforementioned SS-CASTLE algorithm, which applies to a particular sub-case of Problem (\ref{eq:OptMSCASTLE}), in which we have: (i) $\mathbf{\widetilde{Y}}=\mathbf{Y}$; (ii) $\mathbf{\bar{Y}} \coloneqq [\mathbf{Y}^{0}, \mathbf{Y}^{1}, \ldots, \mathbf{Y}^{L} ] \in \mathbb{R}^{T \times N(L+1)}$; (iii) $\mathbf{\bar{W}} \coloneqq [\mathbf{W}^{0^{T}}, \mathbf{W}^{1^{T}}, \ldots, \mathbf{W}^{L^{T}}]^{T} \in \mathbb{R}^{N(L+1) \times \bar{N}}$, where $\mathbf{W}^{l} \in \mathbb{R}^{N \times N}$ are the matrices of causal coefficients of Equation (\ref{eq:SVARM}). 

The numerical results achievable with SS-CASTLE and MS-CASTLE are reported in the next section and compared with alternative methods. 

\section{Numerical Results}
\label{sec:Numerical results}
This section shows the advantages of SS-CASTLE (i.e., the customization of the proposed MS-CASTLE method to temporal causal structure analysis) over existing alternative methods in solving Equation \eqref{eq:SVARM}.
More specfically, \Cref{sec:DYNO} shows that, when compared to DYNOTEARS, which aims to solve the same optimization problem, SS-CASTLE benefits from the linearization procedure described above to lower the computational cost of each iteration while preserving performance.
In addition, \Cref{sec:LiNGAM} prove SS-CASTLE to outperform both VAR-ICALiNGAM and VAR-DirectLiNGAM when we sample $\boldsymbol{\epsilon}_t$ from a $p$-generalized normal distribution, with $p \in \left\{1, 1.5, 2, 2.5, 100  \right\}$.

\subsection{Comparison with DYNOTEARS}
\label{sec:DYNO}

Here we provide a comparison between SS-CASTLE  and DYNOTEARS (\citealt{pamfil2020dynotears}). We test the two methods over synthetic data, so that the ground truth is known. Our goal is to compare the computational time needed to the two alternative methods to estimate the causal matrices in Equation \eqref{eq:SVARM} with a similar accuracy. 

\subsubsection{Data Generating Process}\label{sec:DYNO-datagen}
We generate synthetic data according to Equation \eqref{eq:SVARM}.
More in details, we set $L=1$ and we assume that each $\epsilon_{i}[t] \sim N\left(0,\sigma^2_i\right)$ with $\sigma^2_i \in [1,2]$. 
Moreover, we set the number of samples $T=1000$ and we pick $N \in \left\{10, 30, 50, 100\right\}$. 
For each of the four possible values of the number of nodes, we simulated $100$ data sets.

Regarding the causal matrices, we generate them by adopting the same procedure illustrated by \citet{hyvarinen}. 
In order to manage the level of sparsity of $\mathbf{W}^0$ and $\mathbf{W}^1$, we introduce the parameter $s \in (0,1)$.
The latter is used as a parameter of a Bernoulli distribution, more precisely $\mathcal{B}(1-s)$, that controls the number of nonzero coefficients of the causal matrices. 
As the number of nodes $N$ grows, we increase the sparsity of the causal structure.
More specifically, the combinations $(N,s)$ used in the experiments below are $\{(10, .80), (30, .85), (50, .90), (100, .95)\}$.

\subsubsection{Results}\label{sec:DYNO-results}

\begin{figure}
    \centering
    \includegraphics[width=\textwidth]{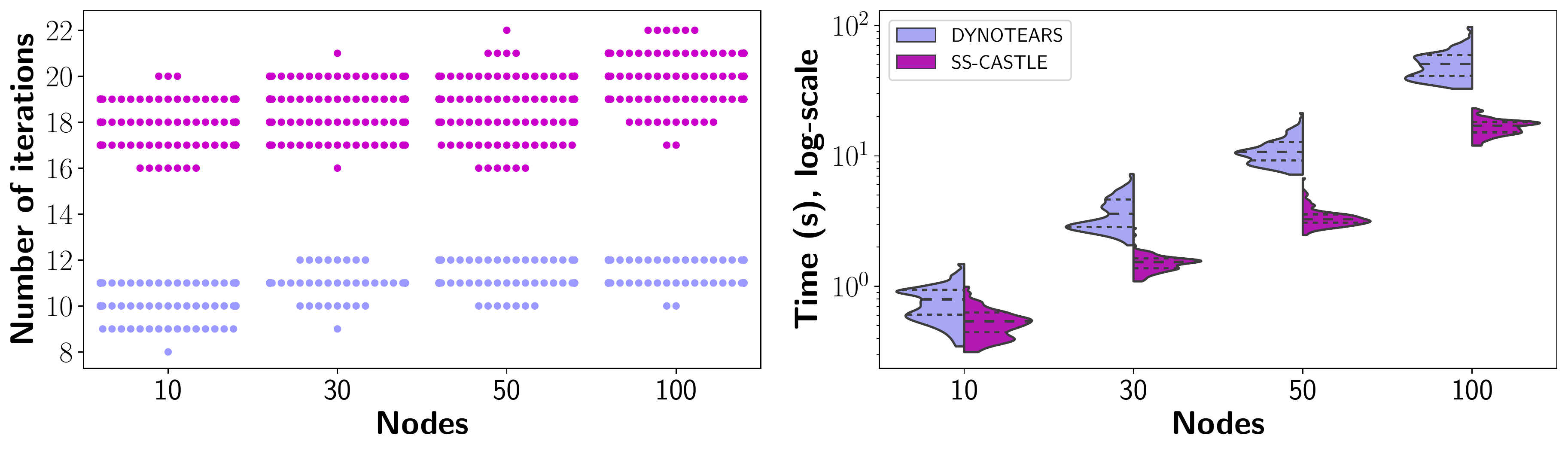}
    \caption{Number of iterations (left) and computational time (right) to solve Equation \eqref{eq:SVARM} shown by DYNOTEARS (purple) and SS-CASTLE (pink).}
    \label{fig:Computational Efficiency}
\end{figure}

\Cref{fig:Computational Efficiency} displays the number of iterations (left) and computational time (right) needed to solve Equation \eqref{eq:SVARM} as a function of the number of nodes, shown by DYNOTEARS (purple) and SS-CASTLE (pink).
On the left of \Cref{fig:Computational Efficiency}, we report a swarm plot in which, given a certain number of nodes, each point represents the number of iterations required by each algorithm to retrieve the solution. 
In accordance with \cref{sec:DYNO-datagen}, for each value of $N$ we have 100 points per algorithm.
Therefore, given a certain value of $N$ and a specific number of iterations $n$, the number of points reported in horizontal represents the number of data set (composed by $N$ time series) in which the algorithm has required $n$ iterations to solve the problem.
On the right of \Cref{fig:Computational Efficiency}, we provide a violin plot that depicts, for each value of $N$, the histogram of the computational time (measured in seconds) needed by each algorithm to solve the problem.
Moreover, dashed lines within the histogram represent quartiles.

From Fig. \ref{fig:Computational Efficiency} (left), we see that, even though SS-CASTLE needs more iterations to converge, SS-CASTLE significantly reduces the overall computational time to converge. Furthermore, we also observe that the higher is the network size, the greater is the gain. 
This result is due to a decrease of the computational cost associated to each iteration, as a consequence of the linearization of the dagness function (see \Cref{sec:MS-CASTLE}).  

\begin{figure}[t]
    \centering
    \begin{subfigure}[b]{\textwidth}
        \centering
        \includegraphics[width=\textwidth]{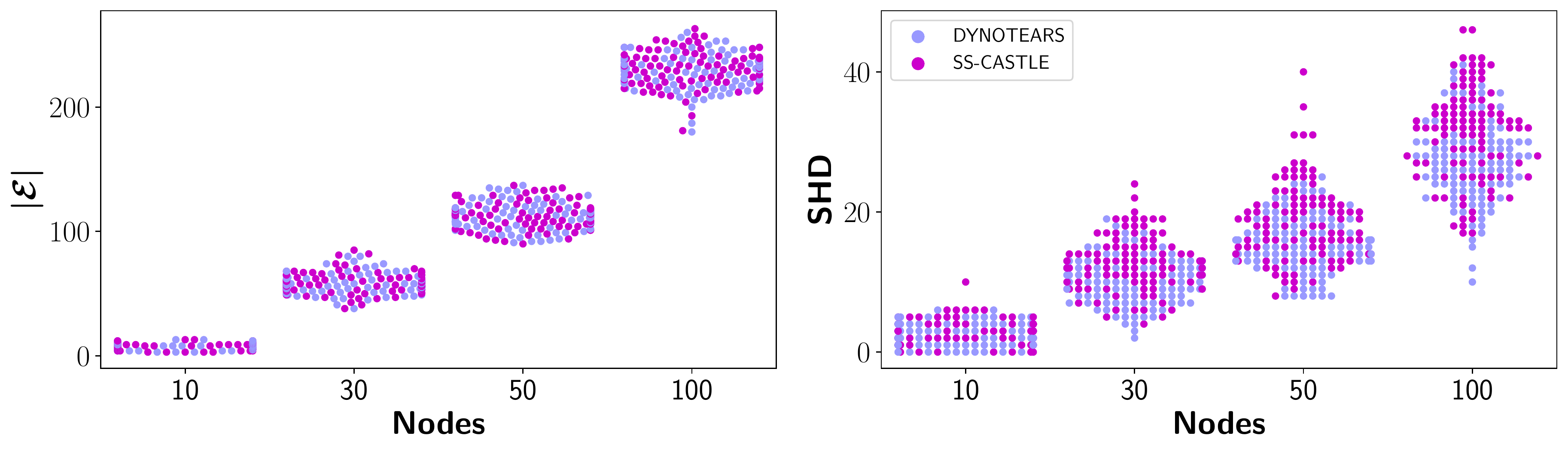}
        \caption{$\mathbf{W}^{0}$}
        \label{fig:Performance Comparison W0}
    \end{subfigure}
    \vfill
    \begin{subfigure}[b]{\textwidth}
        \centering
        \includegraphics[width=\textwidth]{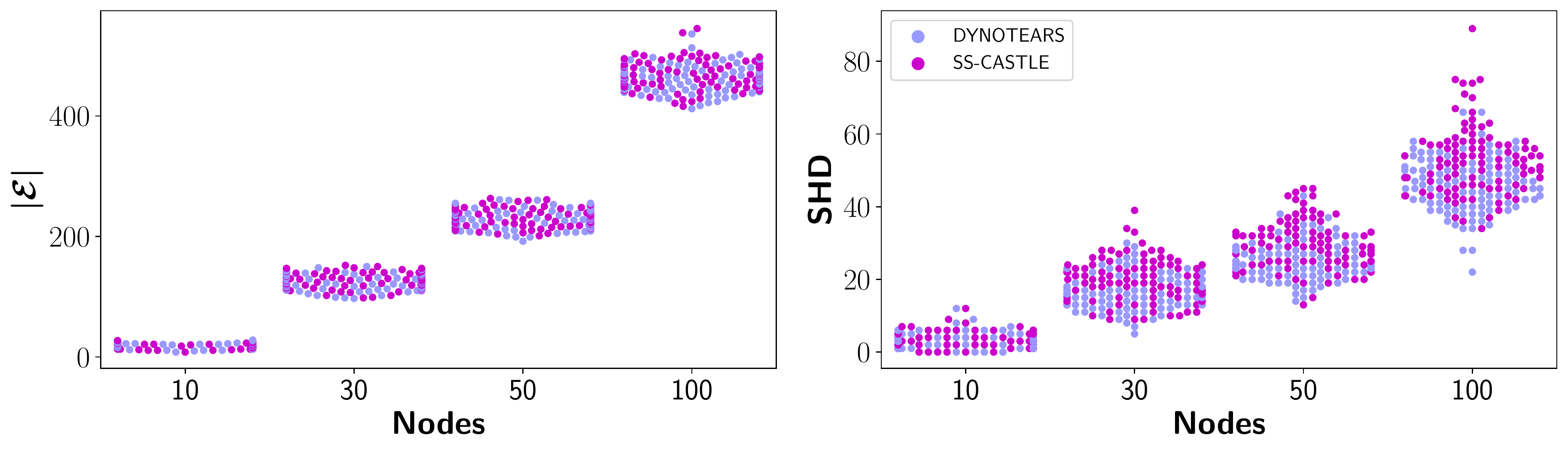}
        \caption{$\mathbf{W}^{1}$}
        \label{fig:Performance Comparison W1}
    \end{subfigure}
    \caption{Swarm plots regarding the number of edges (left) and \textit{Structural Hamming Distance} (SHD, right) of the estimated causal matrices.}
    \label{fig:Performance Comparison}
\end{figure}

In addition, \Cref{fig:Performance Comparison} depicts the swarm plot concerning the number of edges and the \textit{Structural Hamming Distance} (SHD) of the estimated matrices of causal coefficients.
The latter metric indicates the number of modifications needed to retrieve the ground truth from the estimated causal graph (the lower, the better).
First, we observe that both models converge to causal networks of similar size. In addition, by looking at SHD, we notice that dagness function linearization does not cause a worsening in estimation accuracy.

\begin{figure}[p]
    \centering
    \begin{subfigure}[b]{\textwidth}
        \centering
        \includegraphics[width=\textwidth]{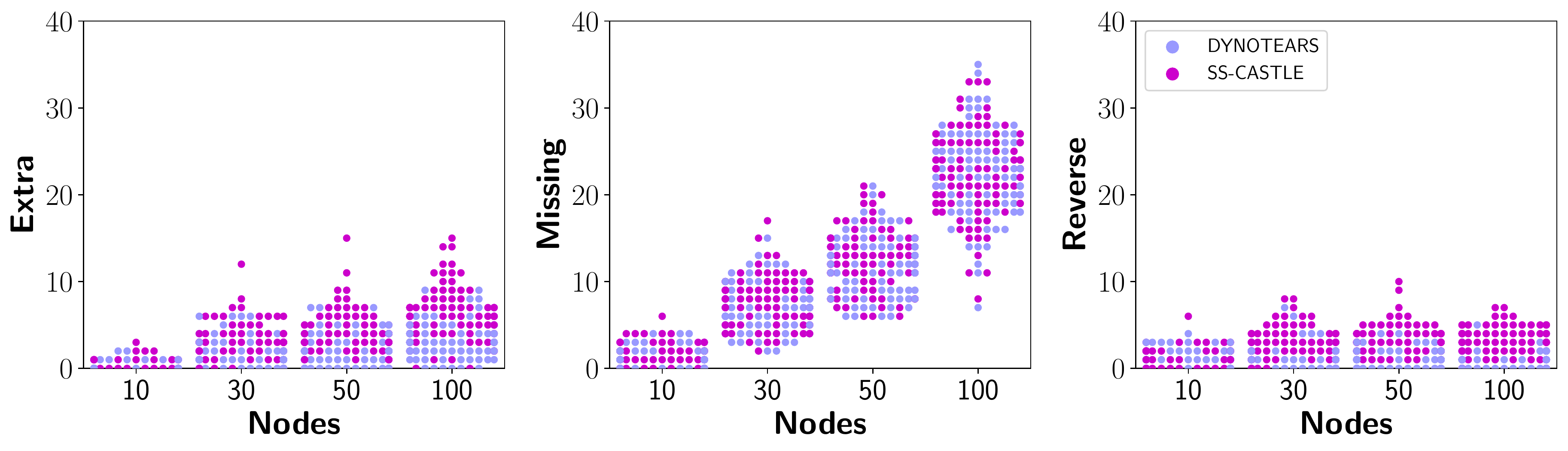}
        \caption{$\mathbf{W}^{0}$}
        \label{fig:Performance Comparison W0 Deepening}
    \end{subfigure}
    \vfill
    \begin{subfigure}[b]{\textwidth}
        \centering
        \includegraphics[width=\textwidth]{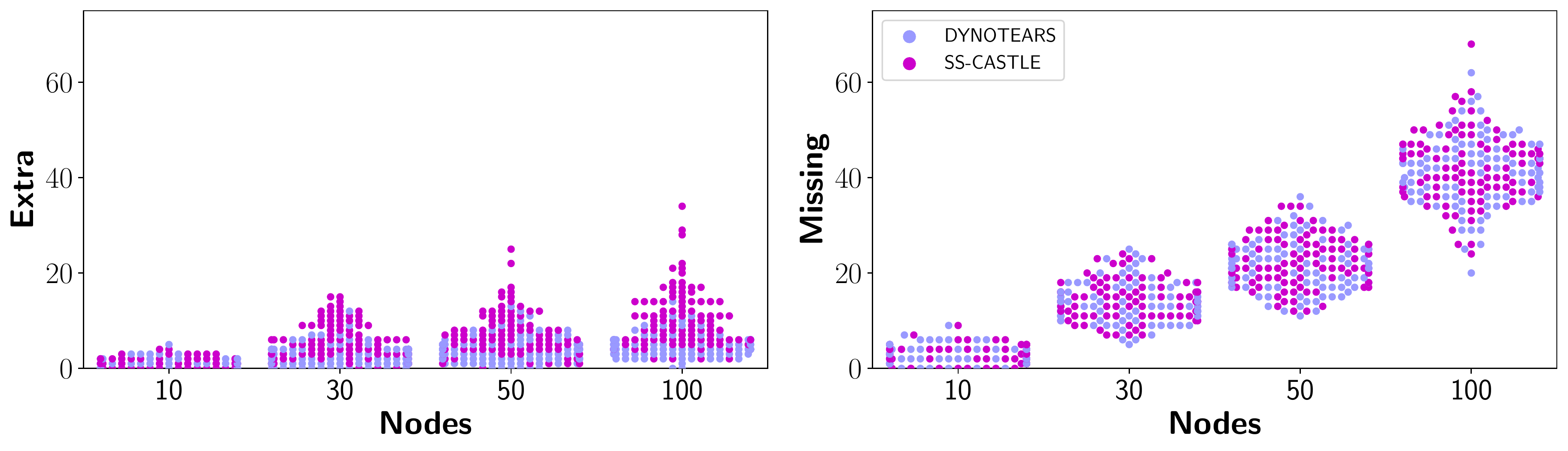}
        \caption{$\mathbf{W}^{1}$}
        \label{fig:Performance Comparison W1 Deepening}
    \end{subfigure}
    \caption{Swarm plots regarding the building blocks of SHD (from the left: extra, missing and reverse edges wrt the ground truth) associated with the estimated causal matrices $\mathbf{W}^{0}$ (a) and $\mathbf{W}^{1}$ (b). Please notice that in case of lagged causal interactions, we cannot observe reverse edges.}
    \label{fig:Performance Comparison Deepening}
\end{figure}

The comparison in terms of accuracy between DYNOTEARS and SS-CASTLE is further detailed below.
In particular, \Cref{fig:Performance Comparison Deepening} shows the contribution of \textit{extra}, \textit{missing} and \textit{reverse} edges to the SHD.
More precisely:
\begin{squishlist}
    \item \textit{extra edges} are estimated edges not encompassed in the causal graph skeleton;
    \item \textit{missing edges} are causal connections present in the ground truth that have not been retrieved, neither with a wrong direction;
    \item \textit{reverse edges} are those connection estimated with a wrong direction.
\end{squishlist}
We observe that the number of missing edges is by far the most dominant component (especially in case of larger networks).
Overall, the models perform similarly across the three components.
It is worth noticing that, since $\mathbf{W}^{1}$ is acyclic by definition, we cannot have reverse edges.

\subsection{Comparison with Linear Non-Gaussian Methods}\label{sec:LiNGAM}

We compared the performance of SS-CASTLE and of two major linear non-Gaussian methods, VAR-ICALiNGAM and VAR-DirectLiNGAM, on synthetic data sets as well.
More in details, the latter two models rely upon the assumption of non-gaussianity of $\boldsymbol{\epsilon}_t$ in Equation \eqref{eq:SVARM}.
VAR-ICALiNGAM belongs to the family of ICA-based methods: first it fits a VAR model to recover lagged causal interactions and then it employs FastICA (\citealt{hyvarinen1999fast}) on VAR residuals to uncover instantaneous relationships. 
In the past, several ICA-based algorithms have been developed.
However, as shown by \citet{moneta2020identification}, previous models are equivalent in terms of performance.
Regarding VAR-DirectLiNGAM, it was proposed in order to solve the possible convergence issues of ICA-based methods (\citealt{himberg2004validating}) and it is guaranteed to retrieve the right solution of the problem if the model assumptions are satisfied and the sample size is very large.
In the experiments below, in order to fit the aforementioned models, we use the \texttt{lingam} Python package\footnote{The package is available at \url{https://github.com/cdt15/lingam}.} made available from the authors.

\subsubsection{Data Generating Process}\label{sec:LiNGAM-datagen}

We generated synthetic data by using Equation \eqref{eq:SVARM}, in which we set $L=1$.
Moreover, we conducted an extensive simulation study in order to asses the robustness of all the methods in different settings. 
In particular, we varied the features of the generated data sets as follows. 

Firstly, we use different data set sizes, $T \in \{100, 500, 1000\}$. 
By varying the number of samples, we can inspect the sensitivity with respect to the data set size of the tested algorithms.
The latter aspect is relevant in several fields, especially when the system at a hand shows nonstationarity.
For instance, this is the case of finance, where practitioners usually deal with a small number of historical observations due to the continuous evolution of financial markets.
Secondly, we vary the network size, $N \in \{10, 30, 50\}$. 
Concerning the level of sparsity and the generation of the causal matrices, we adopt the same methodology described in \cref{sec:DYNO-datagen}.
Last but not least, we sample $\boldsymbol{\epsilon}_t$ from a $p$-generalized normal distribution, with $p \in \left\{1, 1.5, 2, 2.5, 100  \right\}$. 
The $p$-generalized normal distribution is defined as follows (\citealt{kalke2013simulation}).
\begin{definition}[$p$-generalized normal distribution]
Let us consider $x \in \mathbb{R}$, $p \in \mathbb{R}^{+}$. Therefore, the $p$-generalized normal distribution has density function equal to
\begin{equation}\label{eq:pgen}
    f_{p}(x) = \dfrac{p^{1-1/p}}{2\Gamma \left(1/p \right)}\exp\left[-\dfrac{|x|^p}{p}\right],  
\end{equation}
where $\Gamma$ is the gamma function.
\end{definition}
The parameter $p$ plays a key role. 
In particular, it determines the rate of decay of Equation \eqref{eq:pgen}.
More in detail: (i) $p=1$ corresponds to a Laplace distribution; (ii) $p=1.5$ is the super-Gaussian case; (iii) for $p=2$ we get the normal distribution; (iv) $p=2.5$ is the sub-Gaussian case; (v) for $p=100$ we obtain approximately a uniform distribution. 
Therefore, as $p$ diverges from 2, the non-normality of $\boldsymbol{\epsilon}_t$ is enhanced. 
For each combination of the parameters above, we generate 100 data sets.

\subsubsection{Results}\label{sec:LiNGAM-results}

\begin{table}
\centering
\begin{tabular}{ccr|r|r}
     &       &  \multicolumn{3}{c}{$\lambda$} \\
\hline
Nodes & p &  T=100 & T=500 & T=1000 \\
\midrule
\multirow{5}{*}{10} & 1.0   &  0.50  &  0.10   &  0.10 \\
     & 1.5   &  0.10   &  0.05   &  0.05 \\
     & 2.0   &  0.50   &  0.10   &  0.10   \\
     & 2.5   &  0.10   &  0.05   &  0.01  \\
     & 100.0 &  0.10   &  0.05   &  0.01 \\
\cline{1-5}
\multirow{5}{*}{30} & 1.0   &  0.50    &  0.10   &  0.10\\
     & 1.5   &  0.10   &  0.05   &  0.05 \\
     & 2.0   &  0.50   &  0.10   &  0.10 \\
     & 2.5   &  0.10   &  0.05   &  0.01 \\
     & 100.0 &  0.10   &  0.05   &  0.01 \\
\cline{1-5}
\multirow{5}{*}{50} & 1.0   &  0.50   &  0.10   &  0.10  \\
     & 1.5   &  0.50   &  0.05   &  0.05 \\
     & 2.0   &  0.50   &  0.10   &  0.10 \\
     & 2.5   &  0.10   &  0.05   &  0.05 \\
     & 100.0 &  0.10   &  0.05   &  0.01 \\
\bottomrule
\end{tabular}
\caption{Selected values for $\lambda$ for each of the considered parameters combinations $(T, N, p)_i$.}
\label{table:lambd}
\end{table}

Before testing SS-CASTLE on the generated data, we fine-tune the sparsity strength $\lambda$ onto separate data sets generated according to the procedure explained in \Cref{sec:LiNGAM-datagen}.
More precisely, we let $\lambda$ assume values in the set $\{.001, .005, .01, .05, .1, .5\}$ and, for each combination $(T, N, p)_i$, we chose the best value according to F1-score and SHD. 
Due to the needed computational time, for the case $N=50$ we restrict the possible values of $\lambda$ to $\{.01, .05, .1, .5\}$. 
The latter restriction does not impact the analysis since our objective is not to find the optimal value of the hyper-parameter, rather to set the latter in a data-driven manner. 
The chosen values for $\lambda$ are shown in Table \ref{table:lambd}.

\begin{figure}[t]
    \centering
    \begin{subfigure}[b]{.496\textwidth}
        \centering
        \includegraphics[width=\textwidth]{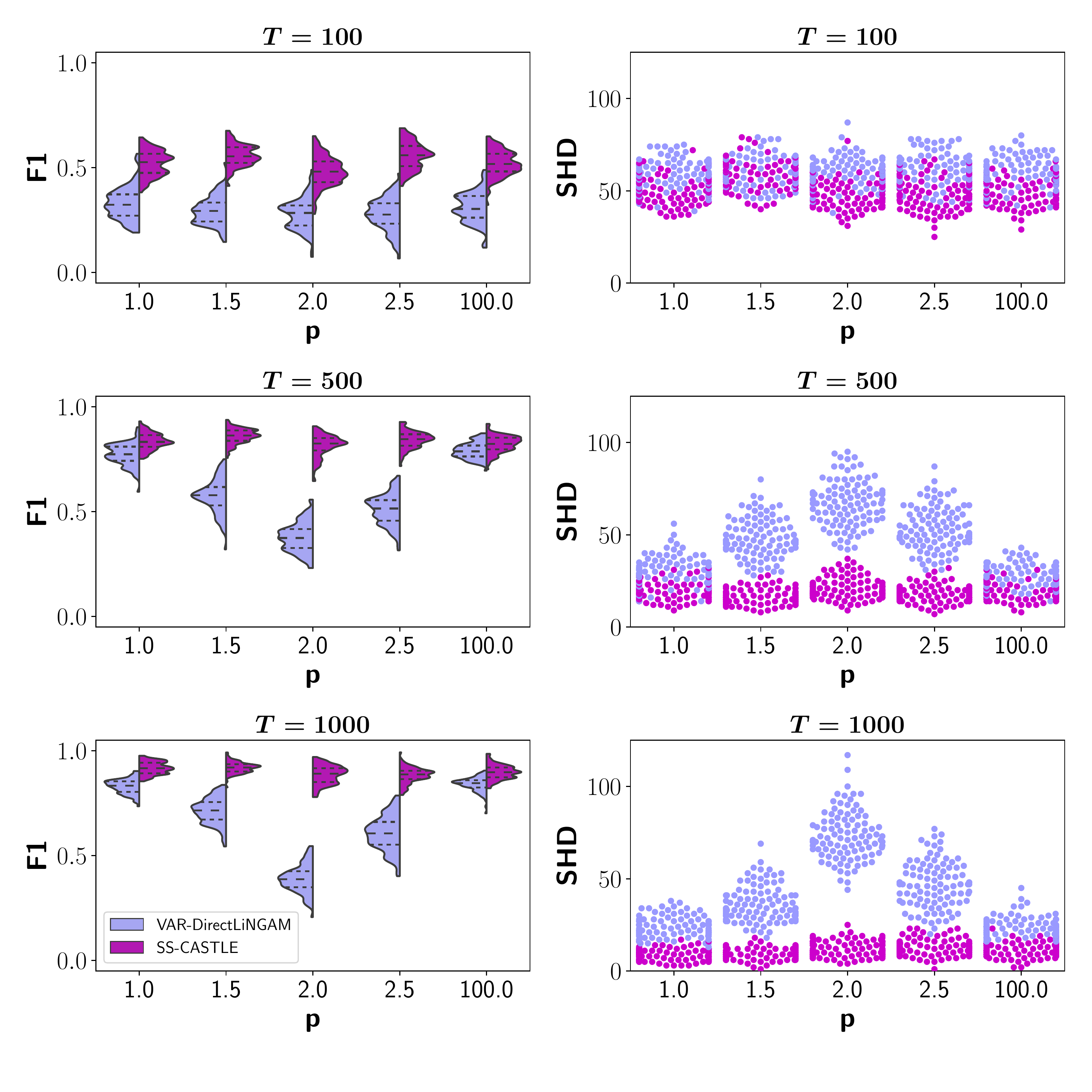}
        \caption{SS-CASTLE vs VAR-DirectLiNGAM}
        \label{fig:metricsTW0_d}
    \end{subfigure}
    \hfill
    \begin{subfigure}[b]{.496\textwidth}
        \centering
        \includegraphics[width=\textwidth]{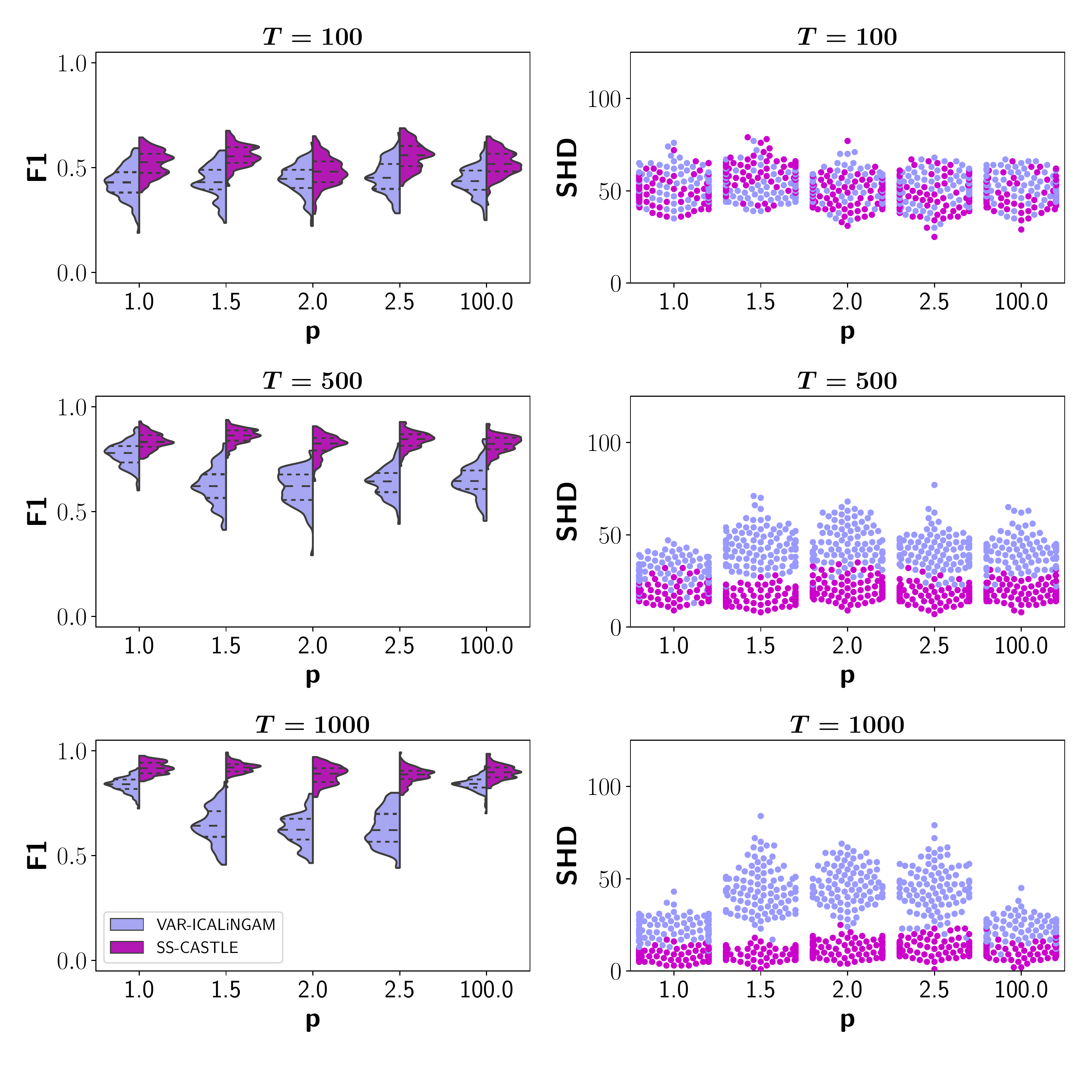}
        \caption{SS-CASTLE vs VAR-ICALiNGAM}
        \label{fig:metricsTW0_i}
    \end{subfigure}
    \caption{Comparison with VAR-DirectLiNGAM (a) and with VAR-ICALiNGAM (b) in the estimation of the matrix  $\mathbf{W}^{0}$. Each subfigure depicts the F1-score (violin plots on the left) and SHD (swarm plots on the right) when $N=30$ and the number of samples $T$ varies in $\{100, 500, 1000\}$.}
    \label{fig:metricsTW0}
\end{figure}

\spara{Sensitivity to data set size.}
\Cref{fig:metricsTW0} depicts the performance comparison in the estimation of $\mathbf{W}^{0}$.
For readability, we provide only the results in case of $N=30$ nodes. 
Results are qualitatively equivalent in the other two cases.
Overall, we can notice the outperformance of SS-CASTLE against VAR-DirecLiNGAM (\Cref{fig:metricsTW0_d}) and VAR-ICALiNGAM (\Cref{fig:metricsTW0_i}) in terms of both F1-score (violin plots on the left) and SHD (swarm plots on the right). 
Even though VAR-DirectLiNGAM shows a slightly better performance than VAR-ICALiNGAM in case of strongly non-Gaussian settings ($p=1$ and $p=100$) and larger data sets ($T=500$ and $T=1000$), it tends to suffer more when the non-gaussianity assumption becomes violated and a lower number of samples is available. 
The latter behaviour is consistent with DirectLiNGAM model assumptions. 
In accordance with the problem formulation (see \Cref{sec:MS-CASTLE}), SS-CASTLE does not show any dependence on $p$. 
Our findings show that it requires a smaller number of data to converge towards a more accurate solution. 

\begin{figure}[p]
    \centering
    \begin{subfigure}[b]{.76\textwidth}
        \centering
        \includegraphics[width=\textwidth]{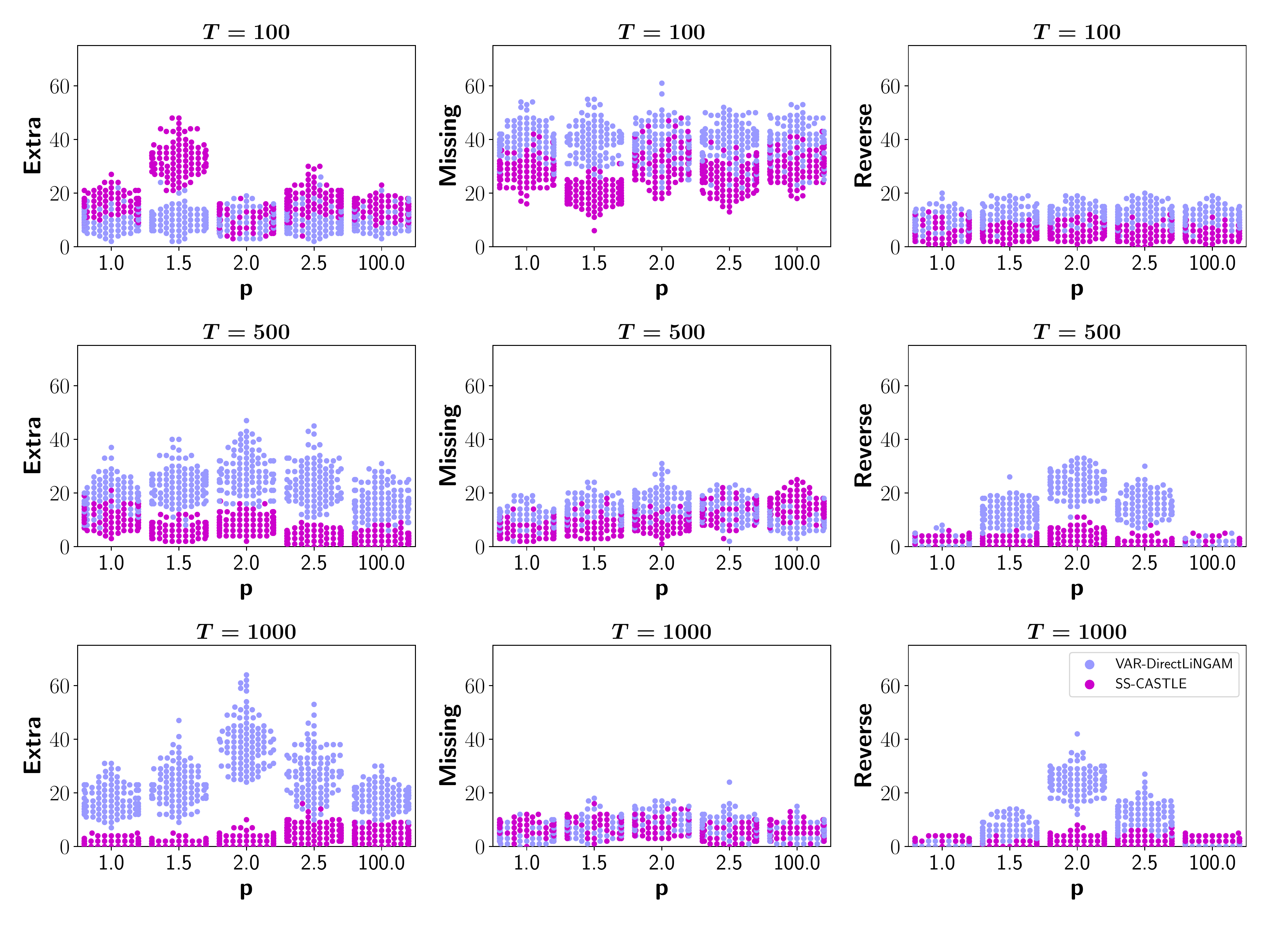}
        \caption{SS-CASTLE vs VAR-DirectLiNGAM}
        \label{fig:swarm_detTW0_d}
    \end{subfigure}
    \vfill
    \begin{subfigure}[b]{.76\textwidth}
        \centering
        \includegraphics[width=\textwidth]{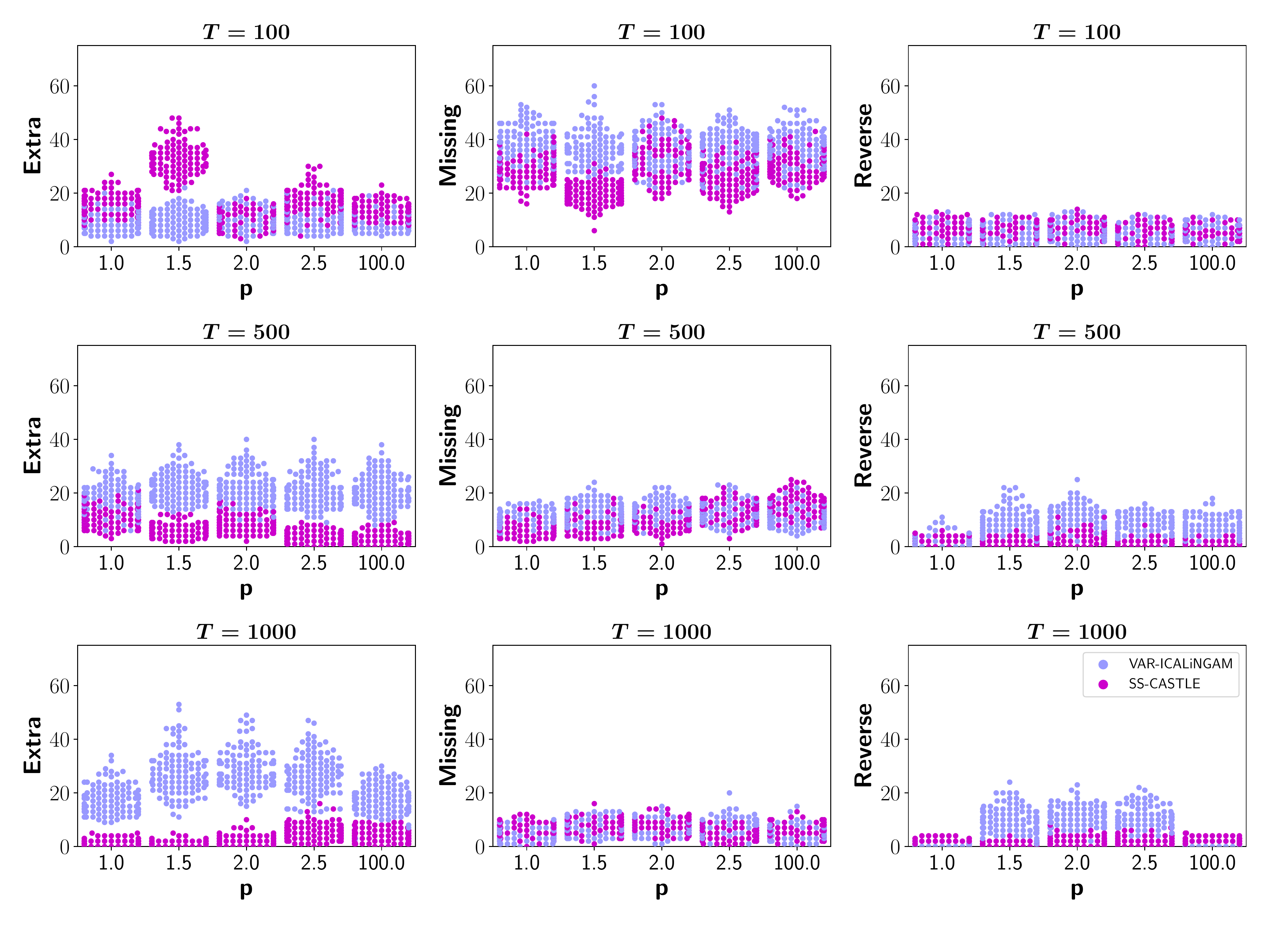}
        \caption{SS-CASTLE vs VAR-ICALiNGAM}
        \label{fig:swarm_detTW0_i}
    \end{subfigure}
    \caption{Comparison with VAR-DirectLiNGAM (a) and with VAR-ICALiNGAM (b) in the estimation of the matrix  $\mathbf{W}^{0}$. Each subfigure reports extra, missing and reverse edges composing SHD, when $N=30$ and the number of samples $T$ varies in $\{100, 500, 1000\}$.}
    \label{fig:swarm_detTW0}
\end{figure}

Moreover, \Cref{fig:swarm_detTW0} provides additional details concerning the structural mistakes made by the models. 
In particular, \Cref{fig:swarm_detTW0_d} provides the comparison with VAR-DirectLiNGAM in terms of extra, missing and reverse edges composing SHD, whereas \Cref{fig:swarm_detTW0_i} that with VAR-ICALiNGAM.
We notice how non-Gaussian methods tend to estimate a greater number of extra and reverse edges, even when $T=1000$.
With regards to missing edges, all the models tend to perform similarly as $T$ grows.
Again, the results show that non-Gaussian methods display a dependence on the value of $p$. 
In order to better interpret the values of SHD, consider that in case $N=30$ and $s=.85$, on average only $65$ entries of $\mathbf{W}^{0}$ are different from zero due to the acyclicity requirement.

\begin{figure}
    \centering
    \begin{subfigure}[b]{.496\textwidth}
        \centering
        \includegraphics[width=\textwidth]{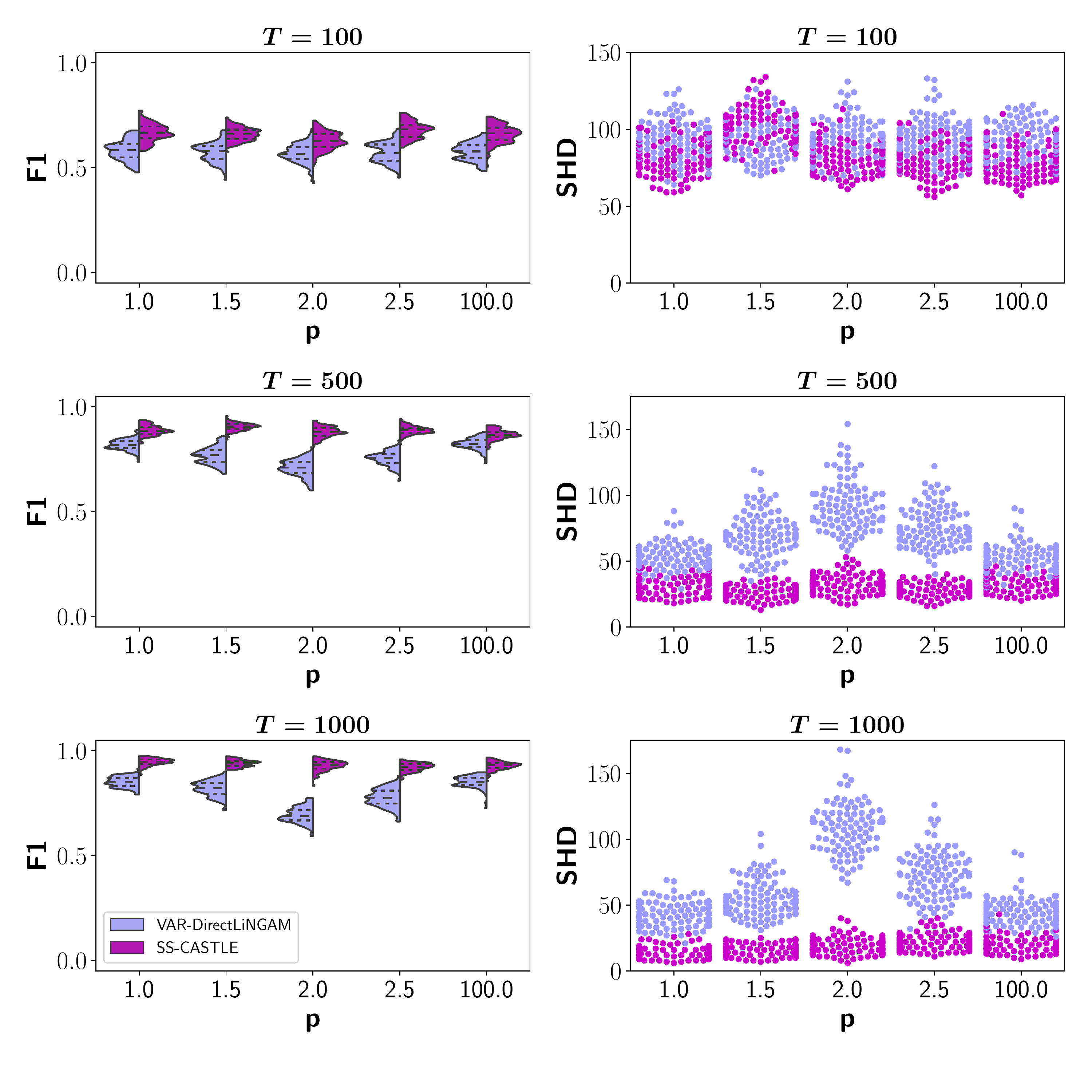}
        \caption{SS-CASTLE vs VAR-DirectLiNGAM}
    \end{subfigure}
    \hfill
    \begin{subfigure}[b]{.496\textwidth}
        \centering
        \includegraphics[width=\textwidth]{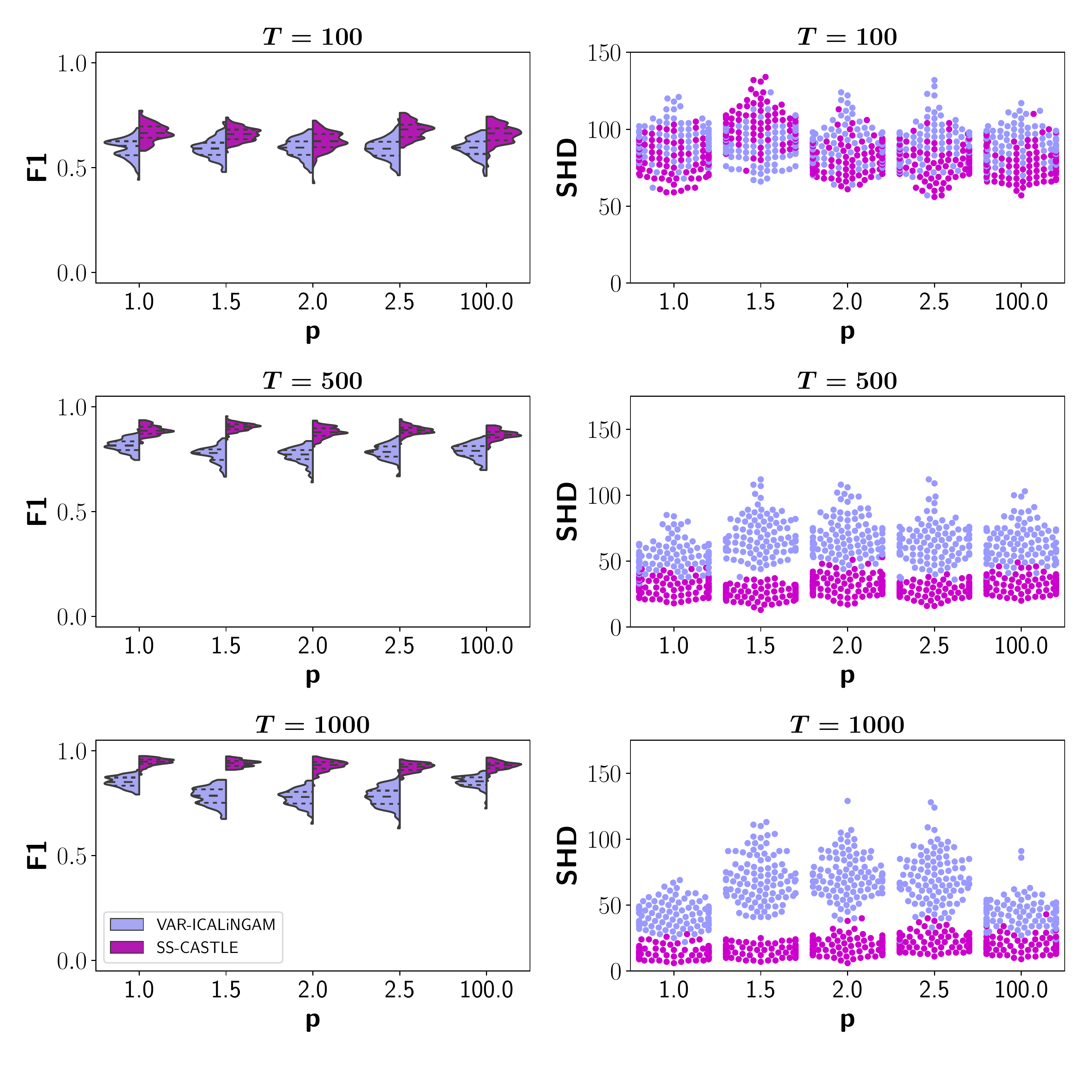}
        \caption{SS-CASTLE vs VAR-ICALiNGAM}
    \end{subfigure}
    \caption{Comparison with VAR-DirectLiNGAM (a) and with VAR-ICALiNGAM (b) in the estimation of the matrix  $\mathbf{W}^{0}$. Each subfigure reports extra, missing and reverse edges composing SHD, when $N=30$ and the number of samples $T$ varies in $\{100, 500, 1000\}$.}
    \label{fig:metricsTW1}
\end{figure}

Regarding the matrix of lagged causal effects $\mathbf{W}^{1}$, \Cref{fig:metricsTW1} shows the same metrics analysed above. 
Overall, we observe again the outperformance of SS-CASTLE that, differently from the considered non-Gaussian methods, contemporaneously estimates inter and intra-layer connections.
In addition, we see that all models tend to be more accurate in retrieving the lagged interactions. Please notice that, given the time ordering, $\mathbf{W}^{1}$ is acyclic by definition. Therefore, all entries could be different from zero. This means that in case of $N=30$ and $s=.85$, on average we have 135 nonzero coefficients.

\begin{figure}
    \centering
    \begin{subfigure}[b]{.496\textwidth}
        \centering
        \includegraphics[width=\textwidth]{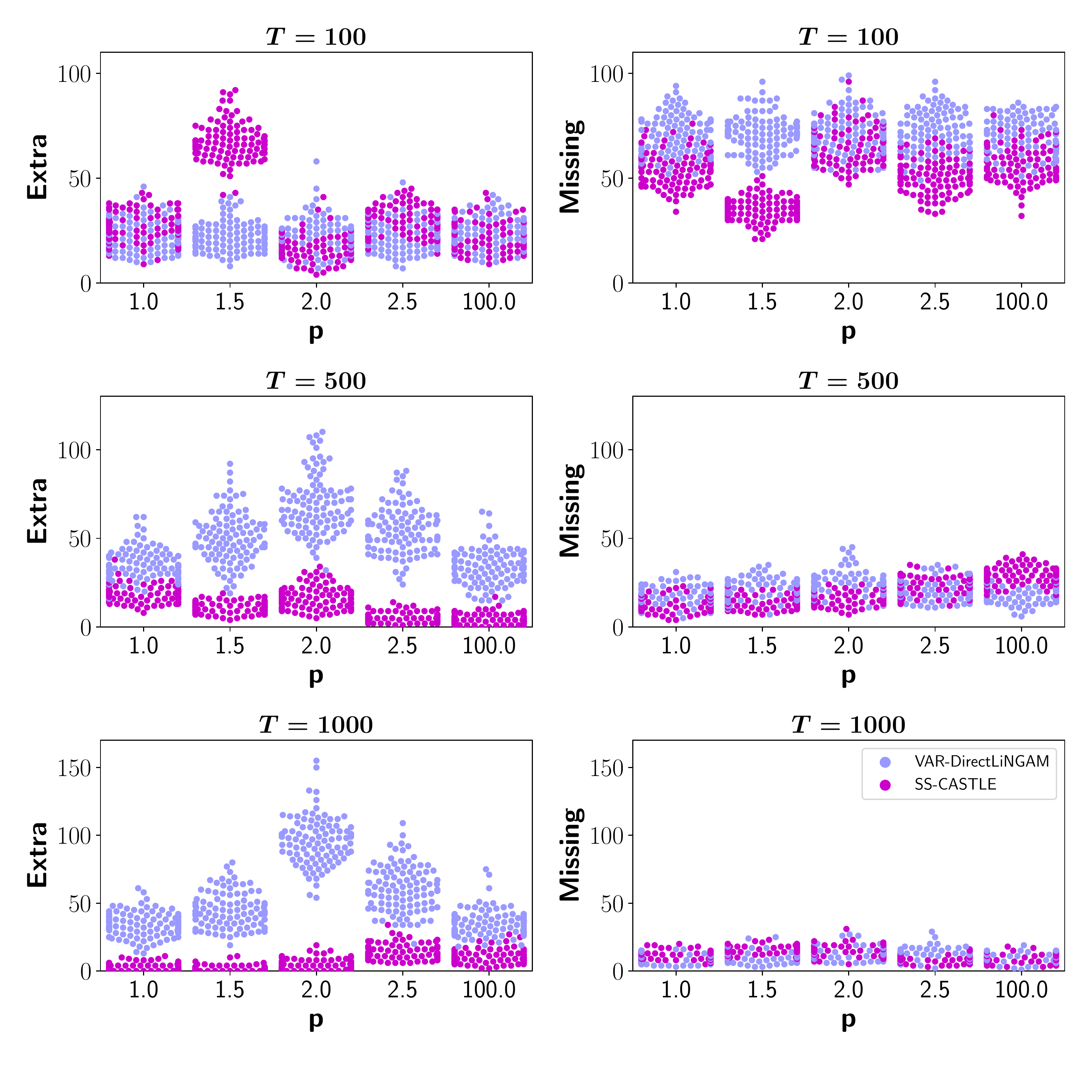}
        \caption{SS-CASTLE vs VAR-DirectLiNGAM}
    \end{subfigure}
    \hfill
    \begin{subfigure}[b]{.496\textwidth}
        \centering
        \includegraphics[width=\textwidth]{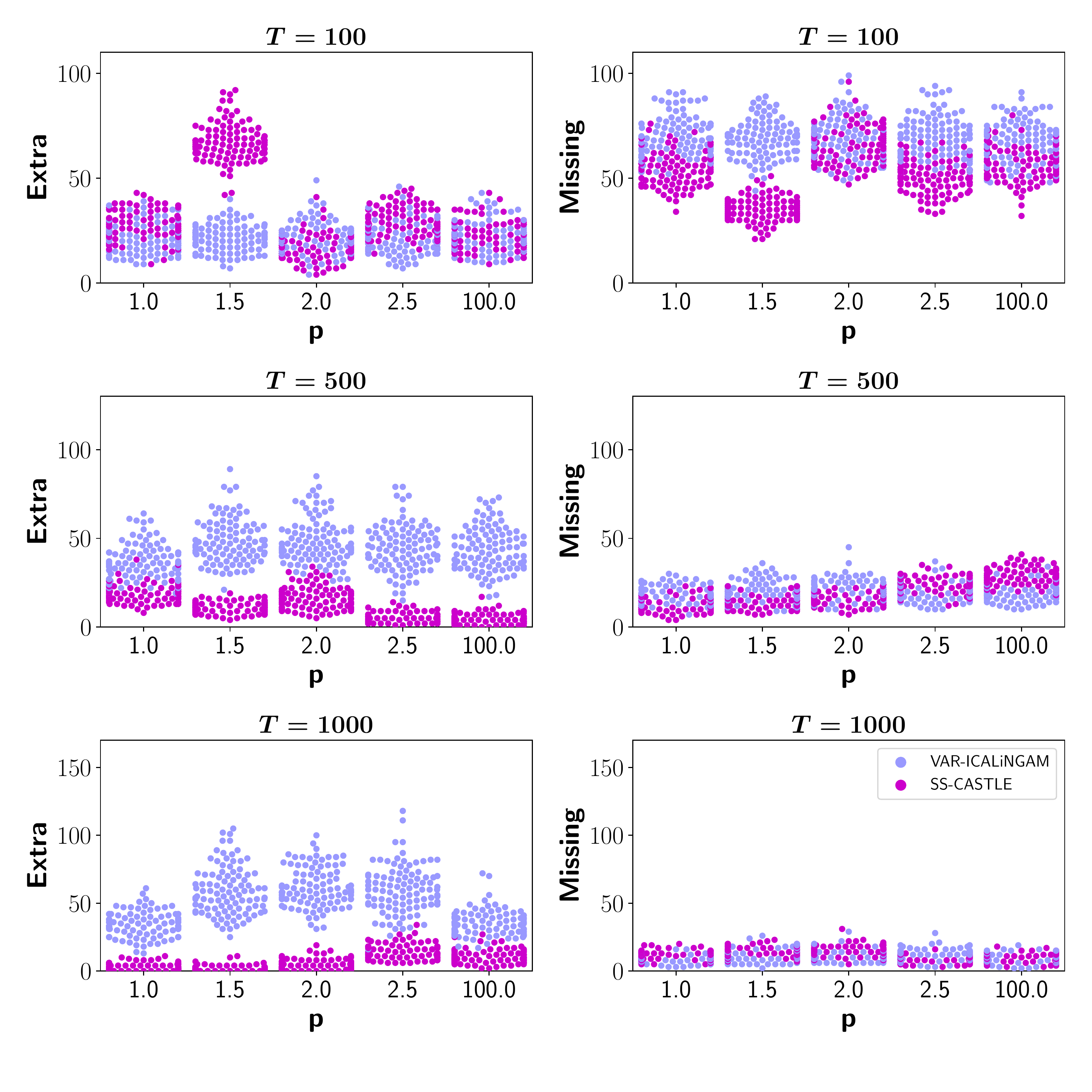}
        \caption{SS-CASTLE vs VAR-ICALiNGAM}
    \end{subfigure}
    \caption{Comparison with VAR-DirectLiNGAM (a) and with VAR-ICALiNGAM (b) in the estimation of the matrix  $\mathbf{W}^{1}$. Each subfigure reports extra and missing edges composing SHD, when $N=30$ and the number of samples $T$ varies in $\{100, 500, 1000\}$.}
    \label{fig:swarm_detTW1}
\end{figure}

Besides, \Cref{fig:swarm_detTW1} depicts additional structural information concerning the estimates.
We see that the non-Gaussian methods are prone to return solutions characterized by a large number of extra edges. 
Regarding SS-CASTLE, the number of missing edges turns out to be the major contributor to SHD in case of small data sets ($T=100$).
With the increase of $T$, extra and missing edges start to contribute similarly to the aforementioned structural metric.

\begin{figure}[t]
    \centering
    \begin{subfigure}[b]{.496\textwidth}
        \centering
        \includegraphics[width=\textwidth]{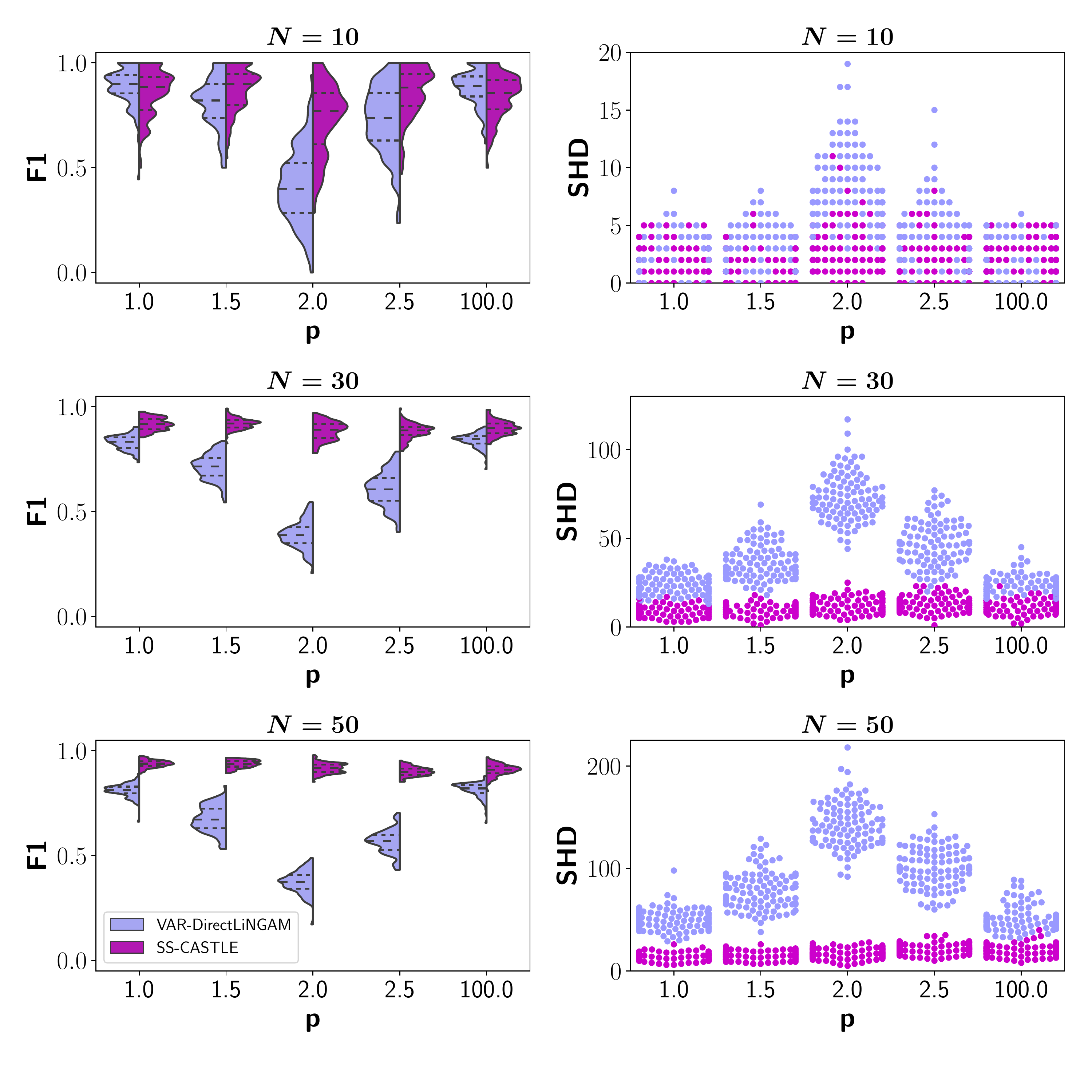}
        \caption{SS-CASTLE vs VAR-DirectLiNGAM}
    \end{subfigure}
    \hfill
    \begin{subfigure}[b]{.496\textwidth}
        \centering
        \includegraphics[width=\textwidth]{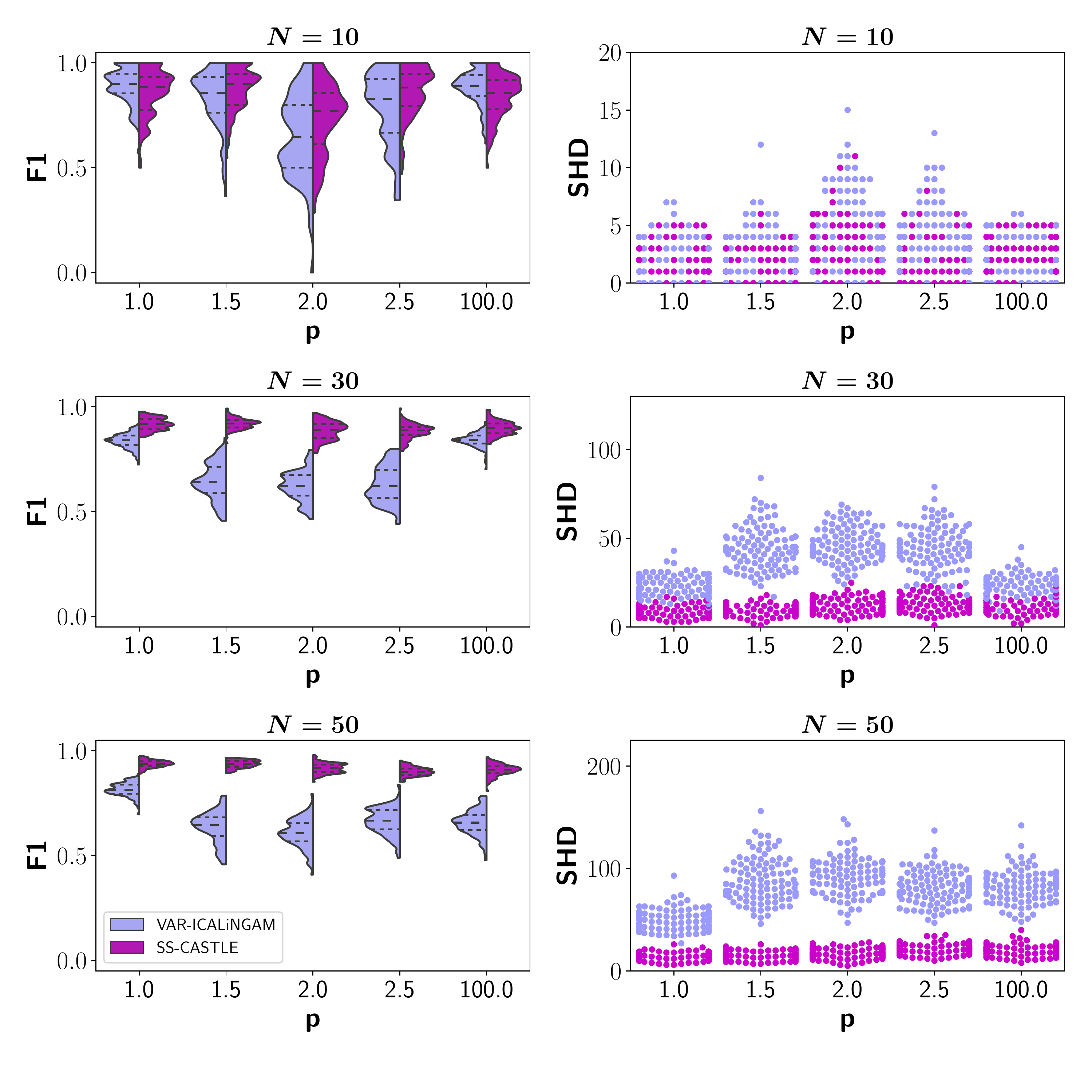}
        \caption{SS-CASTLE vs VAR-ICALiNGAM}
    \end{subfigure}
    \caption{Comparison with VAR-DirectLiNGAM (a) and with VAR-ICALiNGAM (b) in the estimation of the matrix  $\mathbf{W}^{0}$. Each subfigure depicts the F1-score (violin plots on the left) and SHD (swarm plots on the right) when $T=1000$ and the number of time series $N$ varies in $\{10, 30, 50\}$.}
    \label{fig:violinNW0}
\end{figure}

\spara{Sensitivity to network size.}
\Cref{fig:violinNW0} shows the comparison of models performance in the estimation of the instantaneous causal effects.
For readability, we provide only the results in case of $T=1000$.
The results are qualitatively equivalent in the remaining two cases.
At a first glance, we notice the higher error variance in the violin plots related to F1-score, when $N=10$, across all models. 
However, even though F1-score is a normalized metric, this is an effect of the small number of instantaneous causal connections.
Indeed, in case of $N=10$ and $s=.80$, on average the ground truth $\mathbf{W}^{0}$ has only 9 entries different from zero.
As a consequence, a single mistake weighs more. 
Overall, we see that SS-CASTLE outperforms the other methods. 
Moreover, we do not appreciate a decrease in performance when the number of nodes increases.
In addition, SS-CASTLE proves to be robust to changes in the value of $p$.

\begin{figure}[p]
    \centering
    \begin{subfigure}[b]{.76\textwidth}
        \centering
        \includegraphics[width=\textwidth]{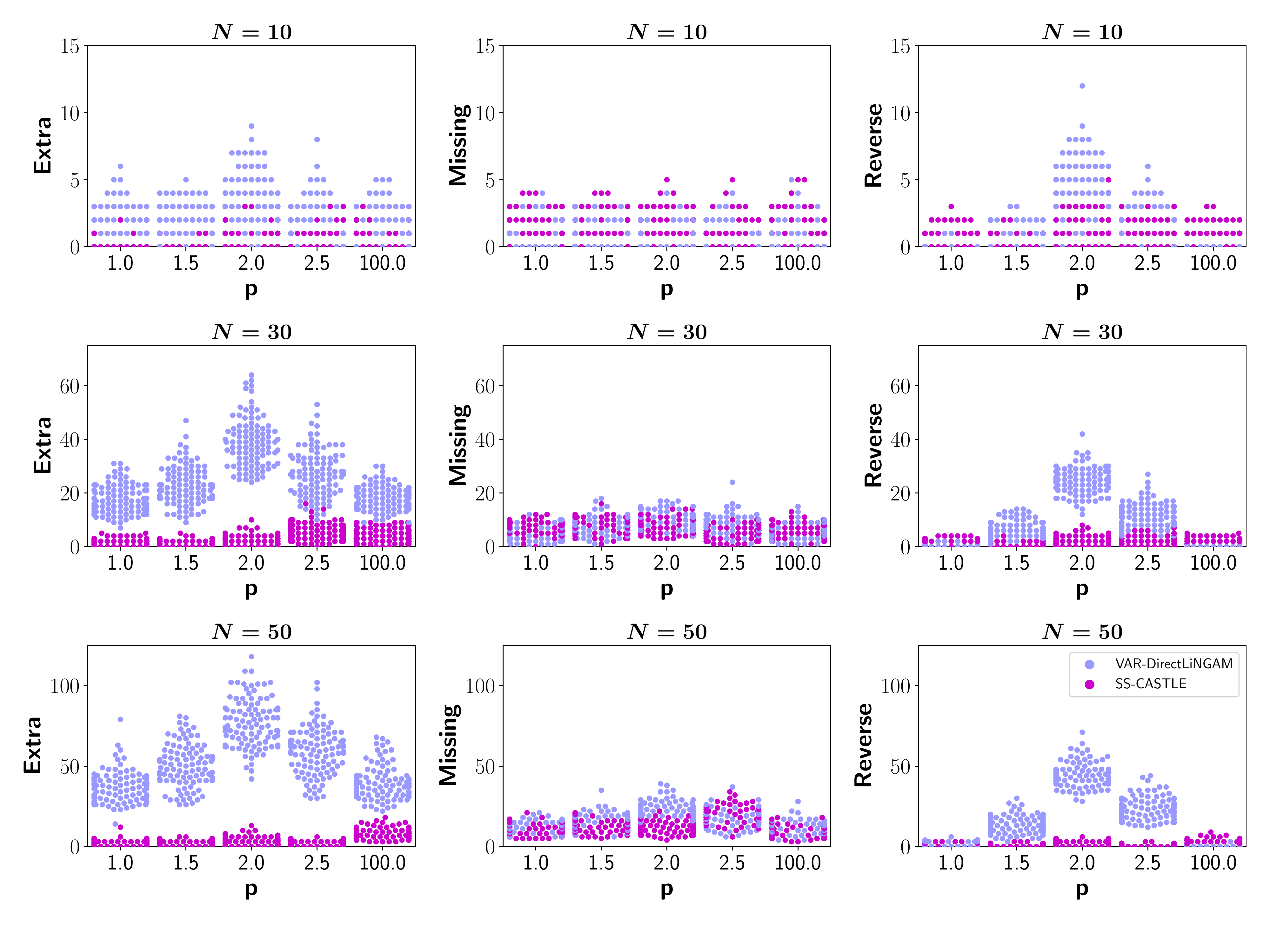}
        \caption{SS-CASTLE vs VAR-DirectLiNGAM}
    \end{subfigure}
    \vfill
    \begin{subfigure}[b]{.76\textwidth}
        \centering
        \includegraphics[width=\textwidth]{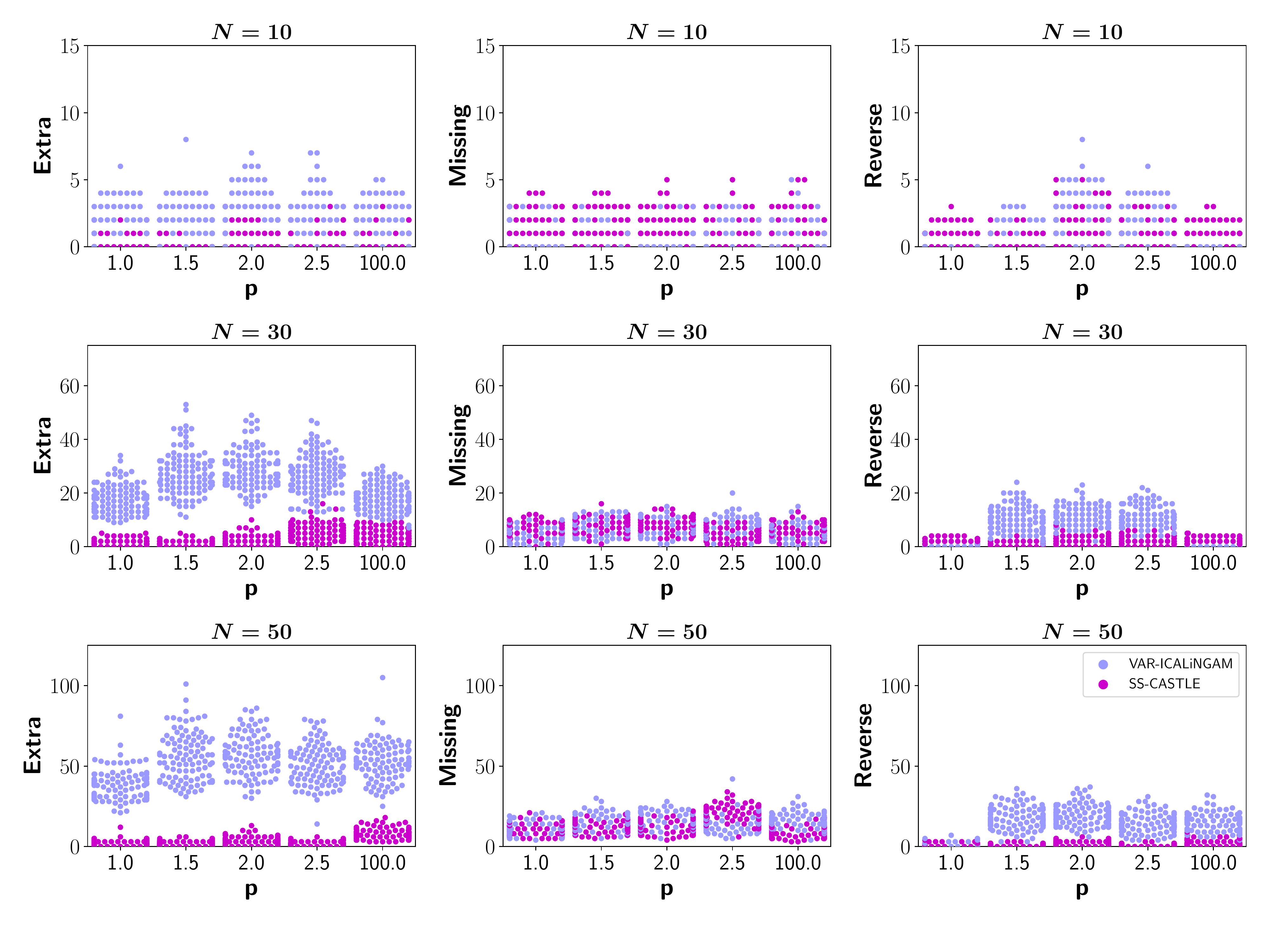}
        \caption{SS-CASTLE vs VAR-ICALiNGAM}
    \end{subfigure}
    \caption{Comparison with VAR-DirectLiNGAM (a) and with VAR-ICALiNGAM (b) in the estimation of the matrix  $\mathbf{W}^{0}$. Each subfigure reports extra, missing and reverse edges composing SHD, when $T=1000$ and the number of time series $N$ varies in $\{10, 30, 50\}$.}
    \label{fig:swarm_detNW0}
\end{figure}

\Cref{fig:swarm_detNW0} provides further information regarding the estimated structure of instantaneous effects.
Again, we see that non-Gaussian methods are prone to estimate a greater number of extra and reverse edges.
With regards SS-CASTLE, the main component of SHD is the number of missing edges.

\begin{figure}
    \centering
    \begin{subfigure}[b]{.496\textwidth}
        \centering
        \includegraphics[width=\textwidth]{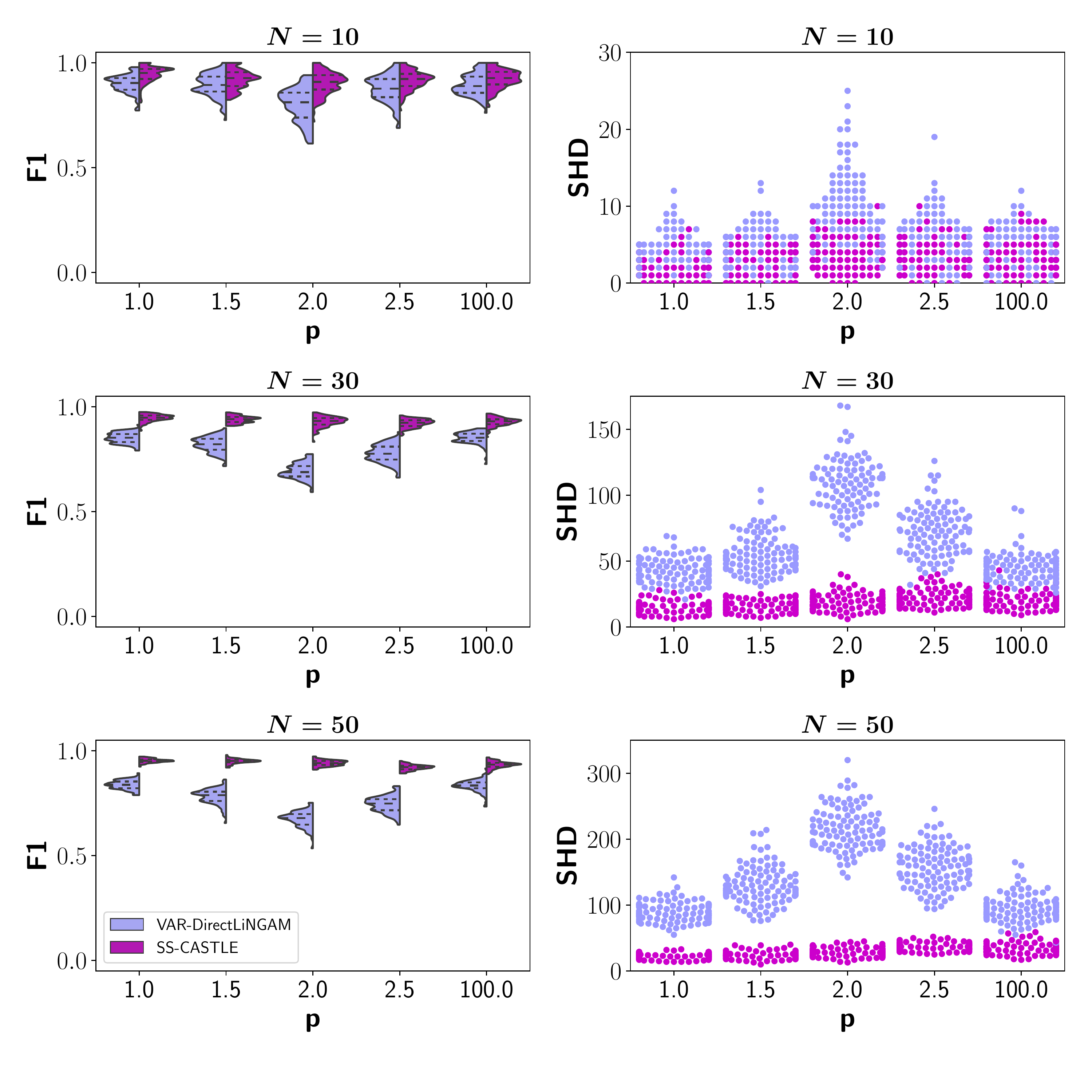}
        \caption{SS-CASTLE vs VAR-DirectLiNGAM}
    \end{subfigure}
    \hfill
    \begin{subfigure}[b]{.496\textwidth}
        \centering
        \includegraphics[width=\textwidth]{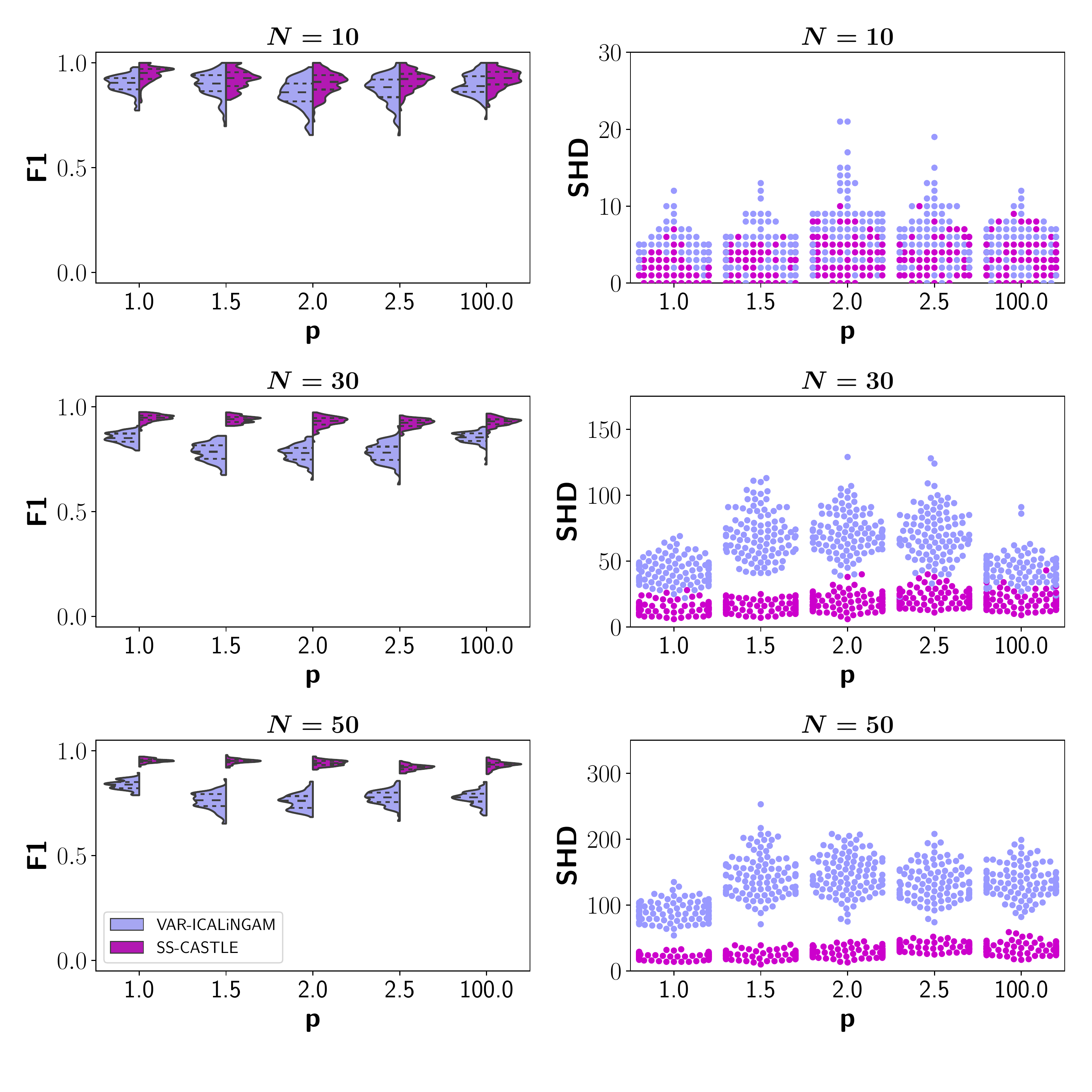}
        \caption{SS-CASTLE vs VAR-ICALiNGAM}
    \end{subfigure}
    \caption{Comparison with VAR-DirectLiNGAM (a) and with VAR-ICALiNGAM (b) in the estimation of the matrix  $\mathbf{W}^{1}$. Each subfigure depicts the F1-score (violin plots on the left) and SHD (swarm plots on the right) when $T=1000$ and the number of time series $N$ varies in $\{10, 30, 50\}$.}
    \label{fig:violinNW1}
\end{figure}

Regarding lagged causal connections, \Cref{fig:violinNW1} shows that the models tend to perform better than in case of instantaneous effects. 
We underline that, in case of $N=10$, we do not observe the same error variance as above.
Indeed, with the same level of sparsity $s$, on average $\mathbf{W}^{1}$ has approximately twice as many non-zero entries than $\mathbf{W}^{0}$.
Overall, we notice that non-Gaussian algorithms tend to suffer when the number of nodes increases. Our results show that VAR-ICALiNGAM tends to be more robust than VAR-DirectLiNGAM as non-gaussianity assumption turns out to be violated. 
Also in this case, SS-CASTLE does not show any decrease in performance while varying $p$. 
In addition, it achieves high performance in case of larger networks as well.

\begin{figure}
    \centering
    \begin{subfigure}[b]{.496\textwidth}
        \centering
        \includegraphics[width=\textwidth]{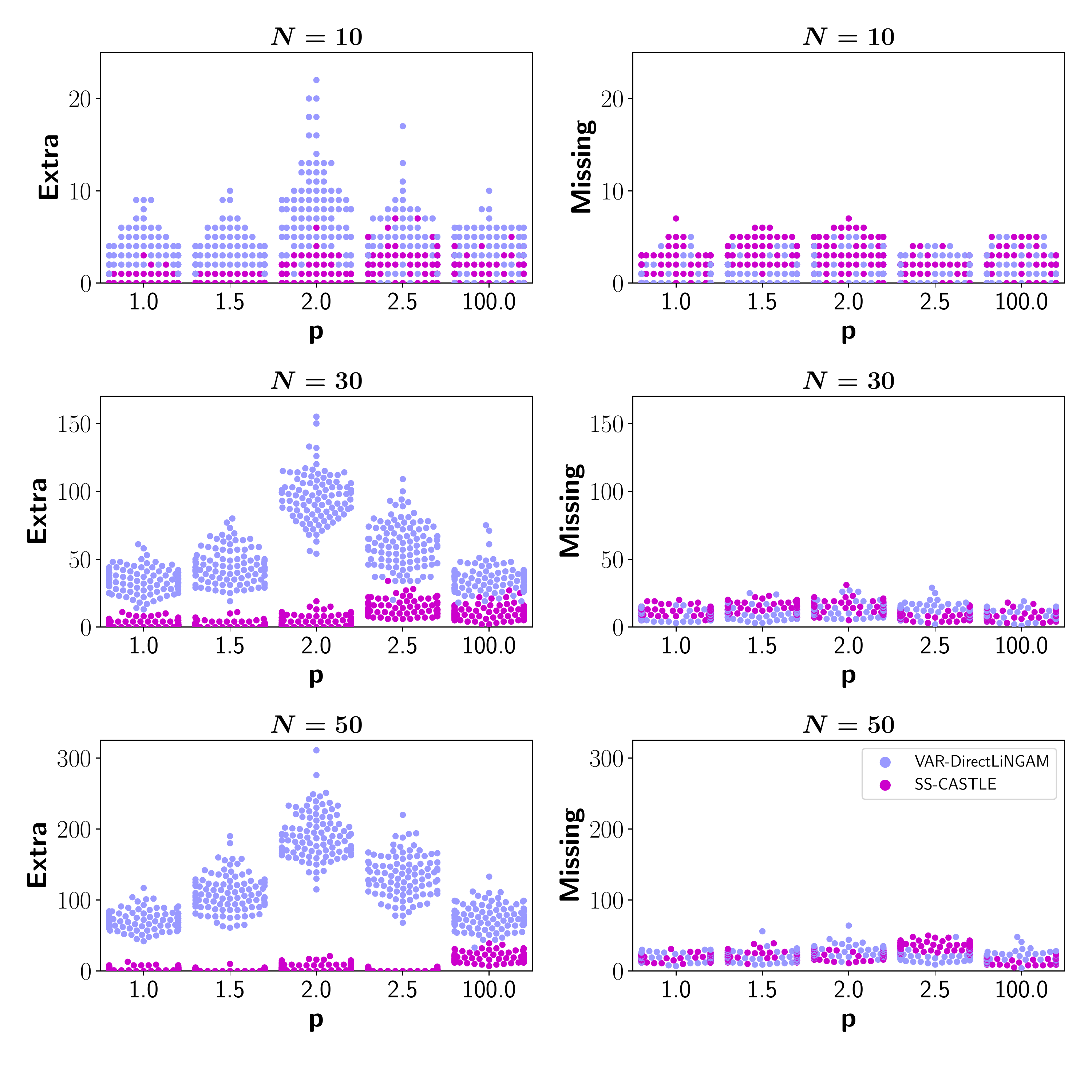}
        \caption{SS-CASTLE vs VAR-DirectLiNGAM}
    \end{subfigure}
    \hfill
    \begin{subfigure}[b]{.496\textwidth}
        \centering
        \includegraphics[width=\textwidth]{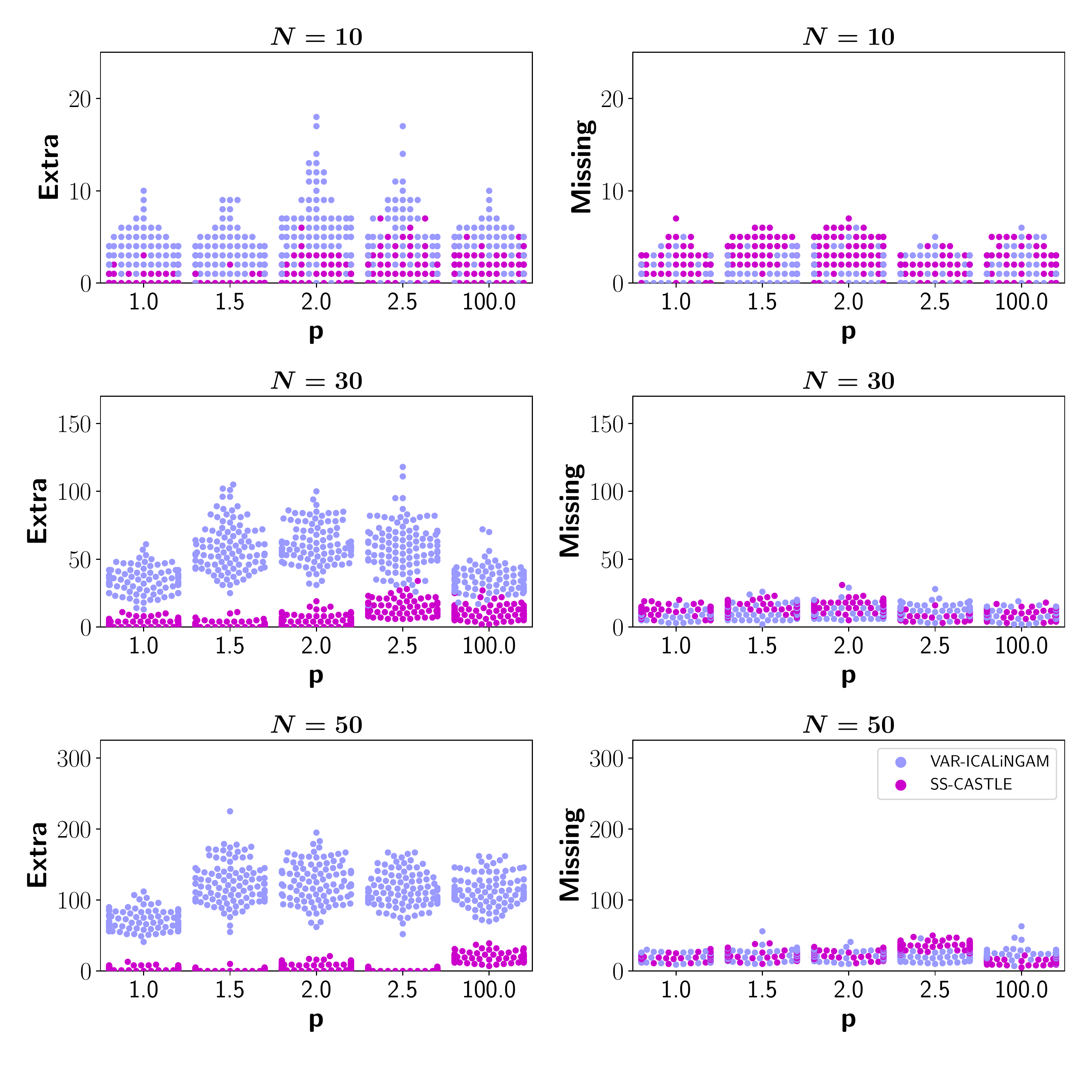}
        \caption{SS-CASTLE vs VAR-ICALiNGAM}
    \end{subfigure}
    \caption{Comparison with VAR-DirectLiNGAM (a) and with VAR-ICALiNGAM (b) in the estimation of the matrix  $\mathbf{W}^{1}$. Each subfigure reports extra and missing edges composing SHD, when $T=1000$ and the number of time series $N$ varies in $\{10, 30, 50\}$.}
    \label{fig:swarm_detNW1}
\end{figure}

As above, \Cref{fig:swarm_detNW1} depicts the building blocks of SHD metric. 
The results show that, even though the models display a similar number of missing arcs, overall SS-CASTLE is more robust to extra edges.

\clearpage

\section{Causal Structure Analysis of Financial Markets}\label{sec:real-data}

\begin{figure}[t]
    \centering
    \includegraphics[width=.8\textwidth]{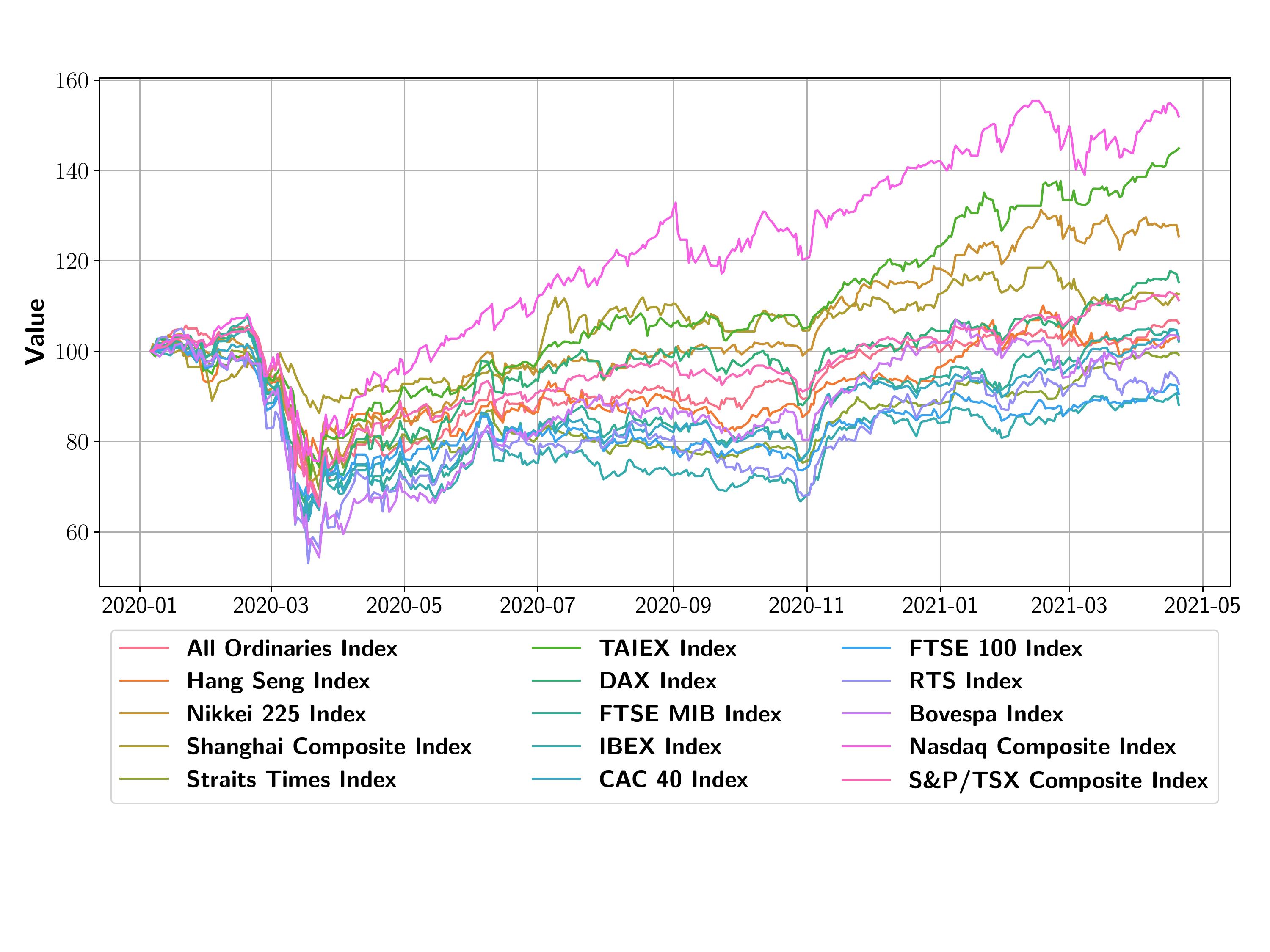}
    \vspace{-15mm}
    \caption{Behavior of equity markets during the considered period. All indexes are rebased to $100$.}
    \label{fig:ts}
    \vspace{-2mm}
\end{figure}

In this section, we apply the proposed technique to infer the causal structure of financial markets. We consider data concerning 15 global equity markets at daily frequency. To focus on covid-19 pandemic period, we restrict our attention to observations from January 2020, the 2nd to April 2021, the 30th. In our analysis, we deal with the following markets: \emph{All Ordinaries Index} (AOR, Australia), \emph{Hang Seng Index} (HSI, Hong Kong), \emph{Nikkei 225 Index} (NKX, Japan), \emph{Shanghai Composite Index} (SHC, Shanghai), \emph{Straits Times Index} (STI, Singapore), \emph{TAIEX Index} (TWSE, Taiwan), \emph{DAX Index} (DAX, Germany), \emph{FTSE MIB Index} (FMIB, Italy), \emph{IBEX Index} (IBEX, Spain), \emph{CAC 40 Index} (CAC, France), \emph{FTSE 100 Index} (UKX, UK), \emph{RTS Index} (RTS, Russia), \emph{Bovespa Index} (BVP, Brazil), \emph{Nasdaq Composite Index} (NDQ, US), \emph{S\&P/TSX Composite Index} (TSX, Canada). The data has been downloaded from Stooq\footnote{The website is reachable at \url{https://stooq.pl/}.}. \Cref{fig:ts} depicts the behavior of the indexes during the considered time window. In particular, the indexes plummet during the first months of 2020 and, subsequently, they show a second downturn during October 2020.  In addition, \Cref{tab:stats} provides summary statistics.
Overall, according to risk adjusted return\footnote{The risk adjusted return is a performance metric, defined as average compounded return to volatility ratio.}, Sortino ratio\footnote{Sortino ratio evaluates risk adjusted performance of a financial instrument discounting for its downside standard deviation.} and average compounded return to max drawdown ratio (ACR/MDD), TWSE and NDQ outperform the rest of the indexes.
Moreover, we see that annualized average compounded returns largely vary across the considered instruments:
while IBEX and UKX are the worst performing, TWSE and NDQ are the most profitable.
Furthermore, all indexes show a high level of volatility.
Among the others, BVP and RTS are the most volatile indexes.
Last bu not least, all indexes suffer heavy losses during the analysed period, as shown by max drawdown metric (MDD).
Interestingly, SHC shows the lowest value.

\begin{table}
    \centering
    \resizebox{.95\columnwidth}{!}{%
    \begin{tabular}{lrrrrrrrrrrrrrrr}
        \toprule
        {} &  AOR &  HSI &  NKX &  SHC &  STI &  TWSE &  DAX &  FMIB &  IBEX &  CAC &  UKX &  RTS &  BVP &  NDQ &  TSX \\
        \midrule
        Avg Comp. Ret. (\%) &                  3.57 &             2.12 &             15.79 &                      8.40 &                -1.30 &        28.29 &      10.08 &            1.15 &       -9.33 &          1.88 &           -7.43 &      -5.60 &           2.15 &                   31.41 &                     7.76 \\
        Volatility (\%)     &                 25.94 &            22.32 &             23.65 &                     19.26 &                21.07 &        19.74 &      29.25 &           31.12 &       30.36 &         28.79 &           26.24 &      36.12 &          39.69 &                   32.46 &                    29.00 \\
        Risk Adj. Ret. (\%) &                  0.14 &             0.10 &              0.67 &                      0.44 &                -0.06 &         1.43 &       0.34 &            0.04 &       -0.31 &          0.07 &           -0.28 &      -0.16 &           0.05 &                    0.97 &                     0.27 \\
        Sortino (\%)        &                  0.18 &             0.13 &              0.99 &                      0.60 &                -0.08 &         2.03 &       0.47 &            0.05 &       -0.41 &          0.09 &           -0.37 &      -0.20 &           0.07 &                    1.34 &                     0.35 \\
        MDD (\%)            &                 37.09 &            25.33 &             31.27 &                     14.62 &                31.93 &        28.72 &      38.78 &           41.54 &       39.43 &         38.56 &           34.93 &      49.46 &          46.82 &                   30.12 &                    37.43 \\
        ACR/MDD            &                  0.10 &             0.08 &              0.50 &                      0.57 &                -0.04 &         0.98 &       0.26 &            0.03 &       -0.24 &          0.05 &           -0.21 &      -0.11 &           0.05 &                    1.04 &                     0.21 \\
        Skew               &                 -1.10 &            -0.37 &              0.27 &                     -0.76 &                -0.44 &        -0.54 &      -0.63 &           -2.26 &       -1.05 &         -0.96 &           -0.80 &      -1.02 &          -1.04 &                   -0.69 &                    -1.01 \\
        Kurtosis           &                  7.37 &             1.87 &              5.01 &                      6.49 &                 7.22 &         5.40 &      10.27 &           20.19 &       11.14 &          9.49 &            8.94 &       7.01 &          11.32 &                    7.35 &                    18.77 \\
        1st \%-ile (\%)      &                 -6.16 &            -4.18 &             -4.47 &                     -3.60 &                -4.63 &        -3.94 &      -5.00 &           -5.21 &       -4.69 &         -5.38 &           -4.03 &      -7.12 &          -9.37 &                   -5.16 &                    -6.76 \\
        5th \%-ile (\%)      &                 -2.33 &            -2.28 &             -2.16 &                     -1.86 &                -1.75 &        -1.80 &      -3.38 &           -2.81 &       -2.95 &         -3.00 &           -2.79 &      -3.35 &          -3.21 &                   -3.12 &                    -2.12 \\
        Min                &                 -9.52 &            -5.56 &             -6.08 &                     -7.72 &                -7.35 &        -5.83 &     -12.24 &          -16.92 &      -14.06 &        -12.28 &          -10.87 &     -13.02 &         -14.78 &                  -12.32 &                   -12.34 \\
        Max                &                  6.56 &             5.05 &              8.04 &                      5.71 &                 6.07 &         6.37 &      10.98 &            8.93 &        8.57 &          8.39 &            9.05 &       9.23 &          13.91 &                    9.35 &                    11.96 \\
        \bottomrule
    \end{tabular}%
    }
    \caption{Summary statistics of equity markets at daily frequency. Average compounded return, volatility, risk adjusted return, and Sortino Ratio are annualised.}
    \label{tab:stats}
\end{table}

To get the series of markets risk, as measured by conditional volatility, we model the logarithmic returns of indexes by means of GARCH models (\citealt{bollerslev1986generalized}).
We use the latter econometric technique to measure systemic risk of equity markets while capturing stylized facts of equity returns, such as volatility clustering (i.e., large (small) swings in stock prices tend to group together), heteroscedasticity (i.e., time-dependent variance) and fat-tailedness (i.e., kurtosis greater than 3).
With regards to GARCH parameters, we select the best combination according to lowest value of BIC criterion (\citealt{schwarz1978estimating}).
The time series of conditional volatility represent our input data set $\mathbf{Y}$.

\subsection{Methodological Approach}\label{sec:methodology}

\begin{figure}[t]
    \centering
    \begin{subfigure}[b]{.496\textwidth}
        \centering
        \includegraphics[width=\textwidth]{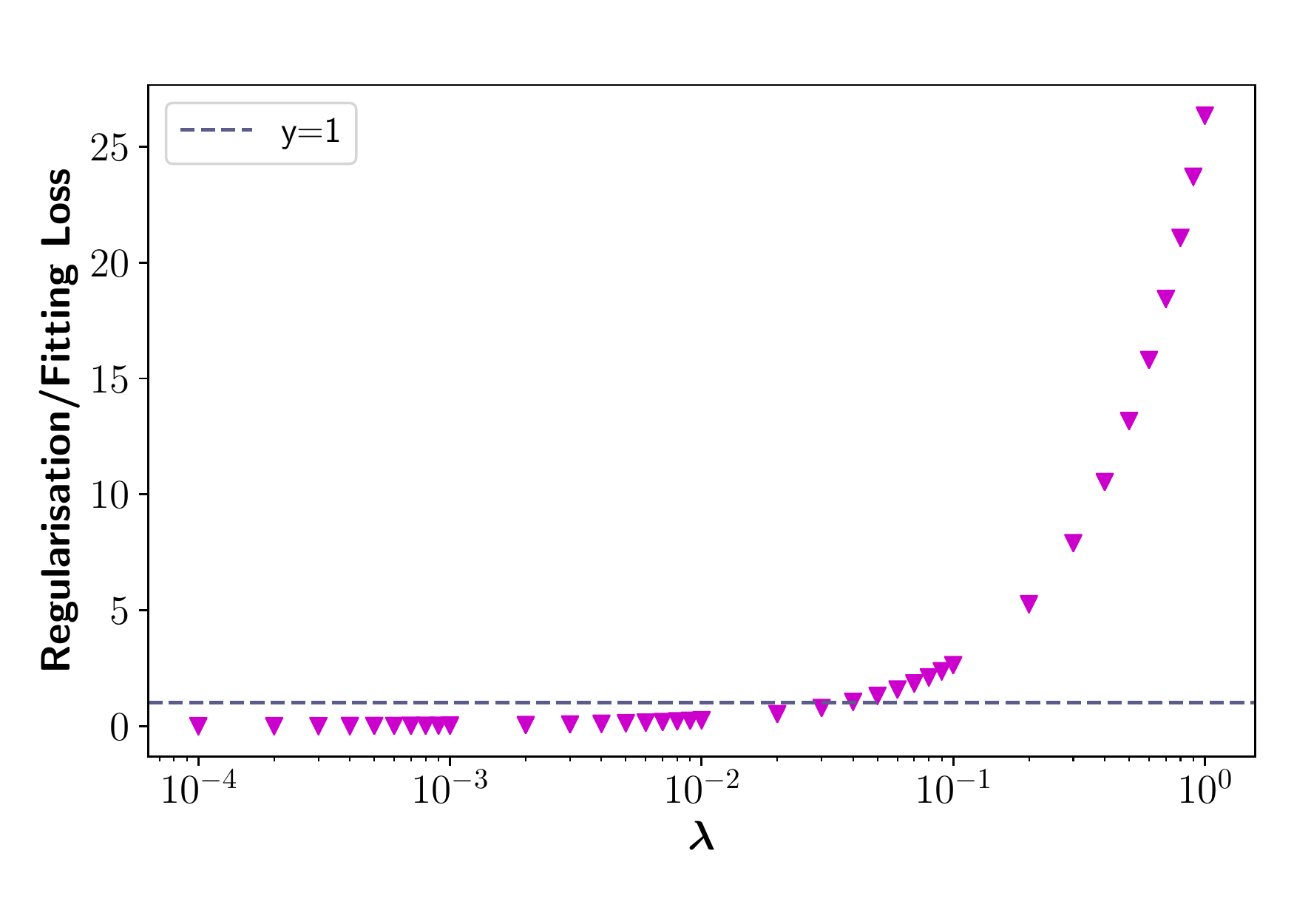}
        \caption{SS-CASTLE}
    \end{subfigure}
    \hfill
    \begin{subfigure}[b]{.496\textwidth}
        \centering
        \includegraphics[width=\textwidth]{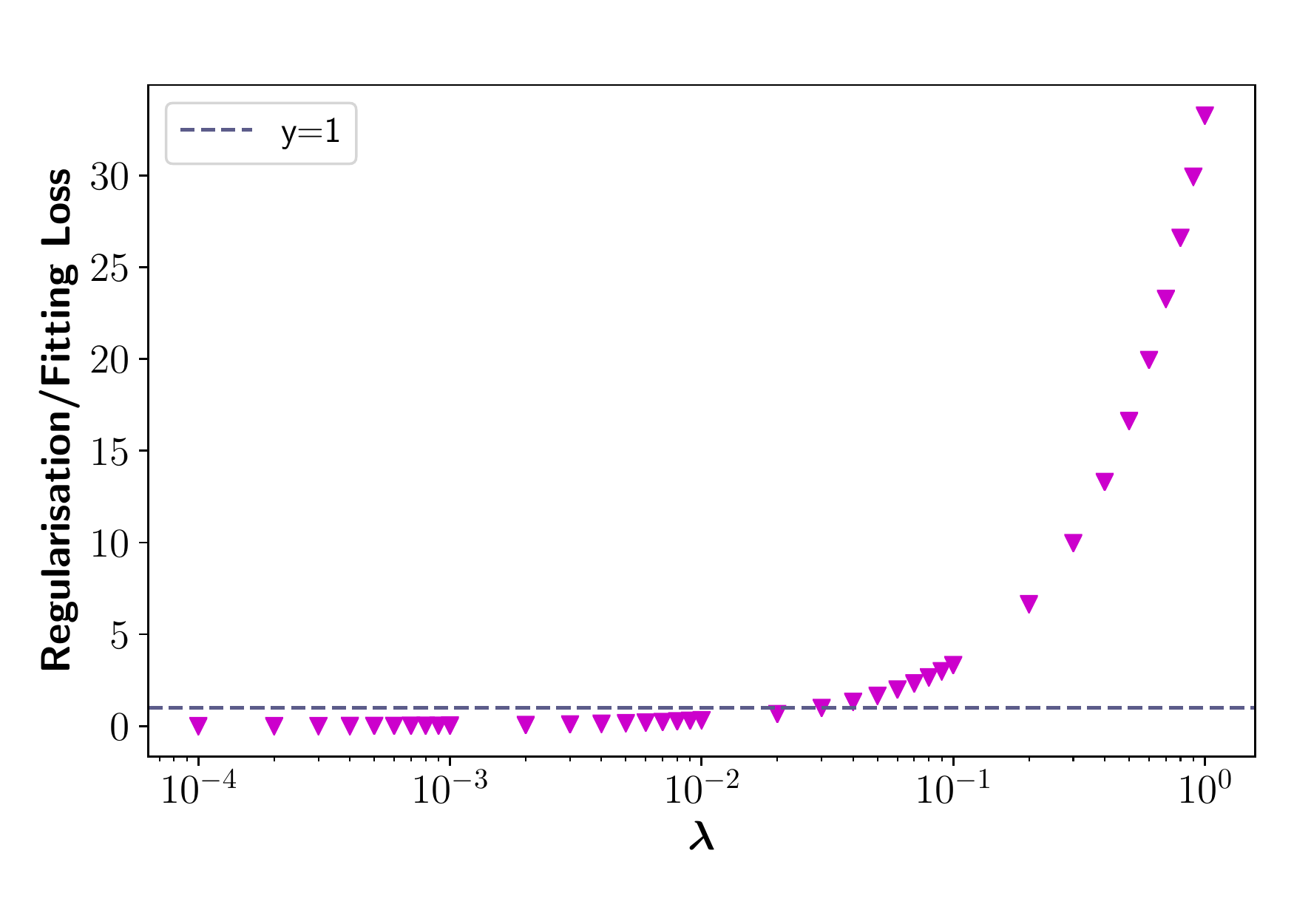}
        \caption{MS-CASTLE}
    \end{subfigure}
\caption{Behaviour of regularization to fitting loss ratio along $\lambda$ for SS-CASTLE (left) and MS-CASTLE (right), where the \emph{x} axis is given in log scale.}
\label{fig:ratio}
\end{figure}

In this section, we deepen the methodology used to retrieve the causal structure underlying the data, coming from the estimation of the causal matrices in both Equations \eqref{eq:SVARM} and \eqref{eq:MSCG}, and constituted by highly persistent edges. In particular, due to non-convexity of Problems (\ref{eq:OptMSCASTLE}), both SS-CASTLE and MS-CASTLE generally converge to stationary points that, possibly, can be very different from each other for diverse values of the sparsity-inducing parameter $\lambda$. Thus, to reduce this ambiguity, in our analysis we look for solutions of SSCG and MSCG that are as persistent as possible with respect to different values of $\lambda$. To this aim, we first choose a suitable range for previous hyper-parameter, looking at the regularization to fitting loss ratio, i.e., the quotient of the division between the second and the first term of the objective function of Problem \eqref{eq:OptMSCASTLE}. \Cref{fig:ratio} shows the behavior of the regularization to fitting loss ratio with respect to $\lambda$, considering both SS-CASTLE (left) and MS-CASTLE (right). As a meaningful range, we select the values that return a ratio from $0.1$ to $1$. In this way, we track the change in causal connections when the sparsity-inducing regularization term becomes as important as the model fitting term. Then, from \Cref{fig:ratio}, we select: (i) $\lambda \in [0.004, 0.04]$ for SS-CASTLE; and (ii) $\lambda \in [0.003, 0.03]$ for MS-CASTLE. For each interval, we pick 10 values for $\lambda$.

Once the range of $\lambda$ is identified, we define the persistence of a causal relation at a threshold $\bar{c}$ as follows. Let us indicate with $\mathbf{r}$ the vector constituted by the regularization to fitting ratios $r_k$ corresponding to the chosen $k$ values of $\lambda$. In addition, consider $w_{ij}^k$ as the causal coefficient from node $i$ to $j$ estimated for $\lambda=\lambda_k$. Then, the persistence of the causal coefficient is
\begin{equation}\label{eq:pers}
    p_{ij} = \dfrac{\sum_{k} {\mathbf{1}_{|w_{ij}^k|>\bar{c}} \cdot r_k}}{\sum_k r_k},
\end{equation}
where $\mathbf{1}_{|w_{ij}^k|>\bar{c}}$ is equal to $1$ iff $|w_{ij}^k|>\bar{c}$, and zero otherwise. Equation \eqref{eq:pers} assigns a higher persistence value to arcs that are present in causal structures estimated from Problem \eqref{eq:OptMSCASTLE} in which $\lambda$ takes on greater values. Also, from Equation \eqref{eq:pers}, it holds $p_{ij} \in [0,1]$. However, the formula does not provide any guarantee regarding the stability of the sign of the causal relation. Indeed, it only considers the presence of an arc and not the value (and therefore the sign) of the causal coefficient associated with the arc. Thus, we define as highly persistent only those edges, with $p_{ij}>0.95$, that show a stable sign of the corresponding causal coefficient for all values of $\lambda$. These edges constitute the causal structures illustrated in the sequel.

\subsection{Experimental Results}\label{sec:results}

To compare temporal and multiscale approaches, we estimate the causal matrices in both Equations \eqref{eq:SVARM} and \eqref{eq:MSCG} by using SS-CASTLE and MS-CASTLE, respectively.

\begin{figure}[t]
    \centering
    \begin{subfigure}[b]{.32\textwidth}
        \centering
        \includegraphics[width=\textwidth]{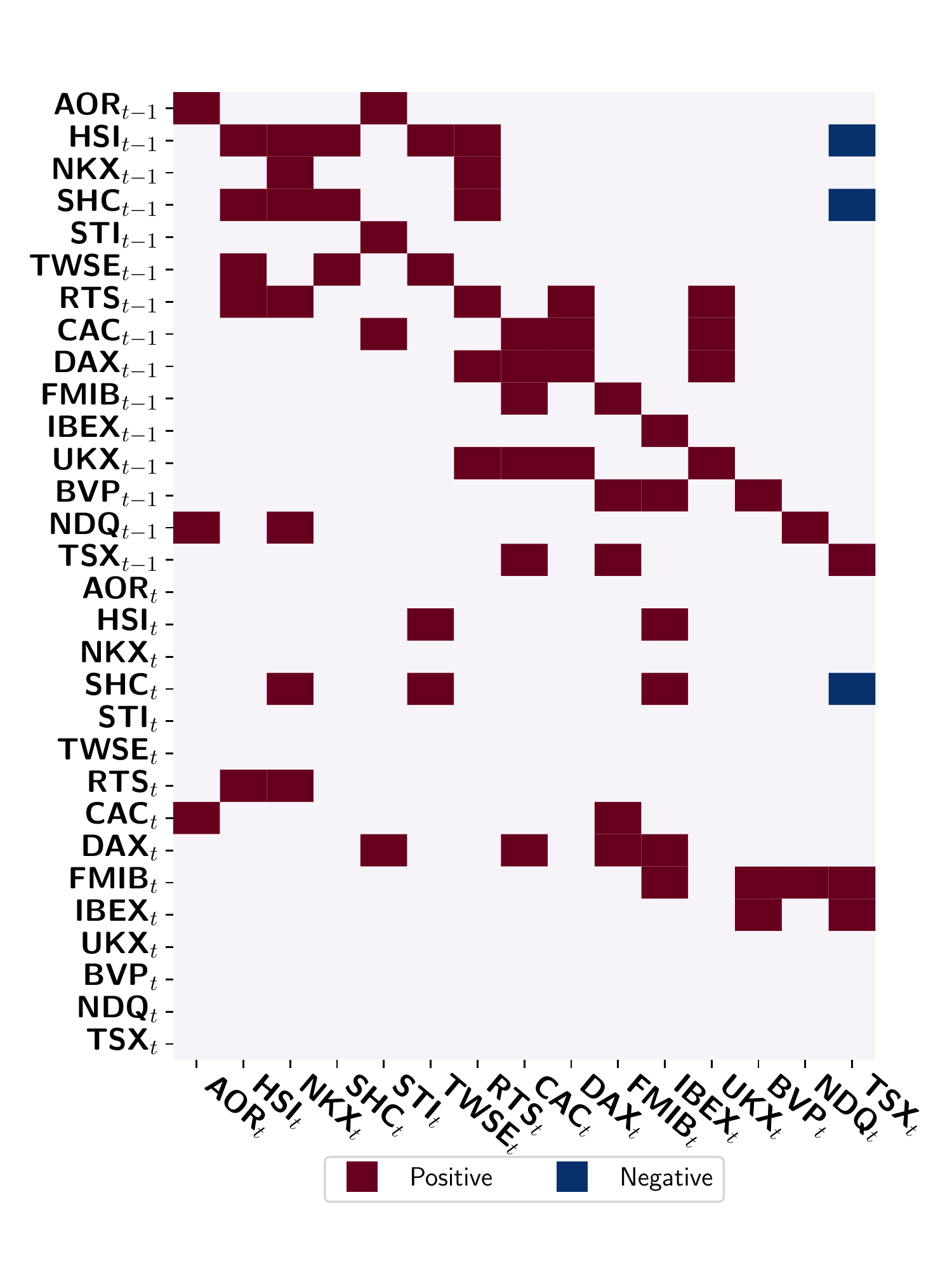}
        \caption{$\bar{c}=0.01$}
    \end{subfigure}
    \hfill
    \begin{subfigure}[b]{.32\textwidth}
        \centering
        \includegraphics[width=\textwidth]{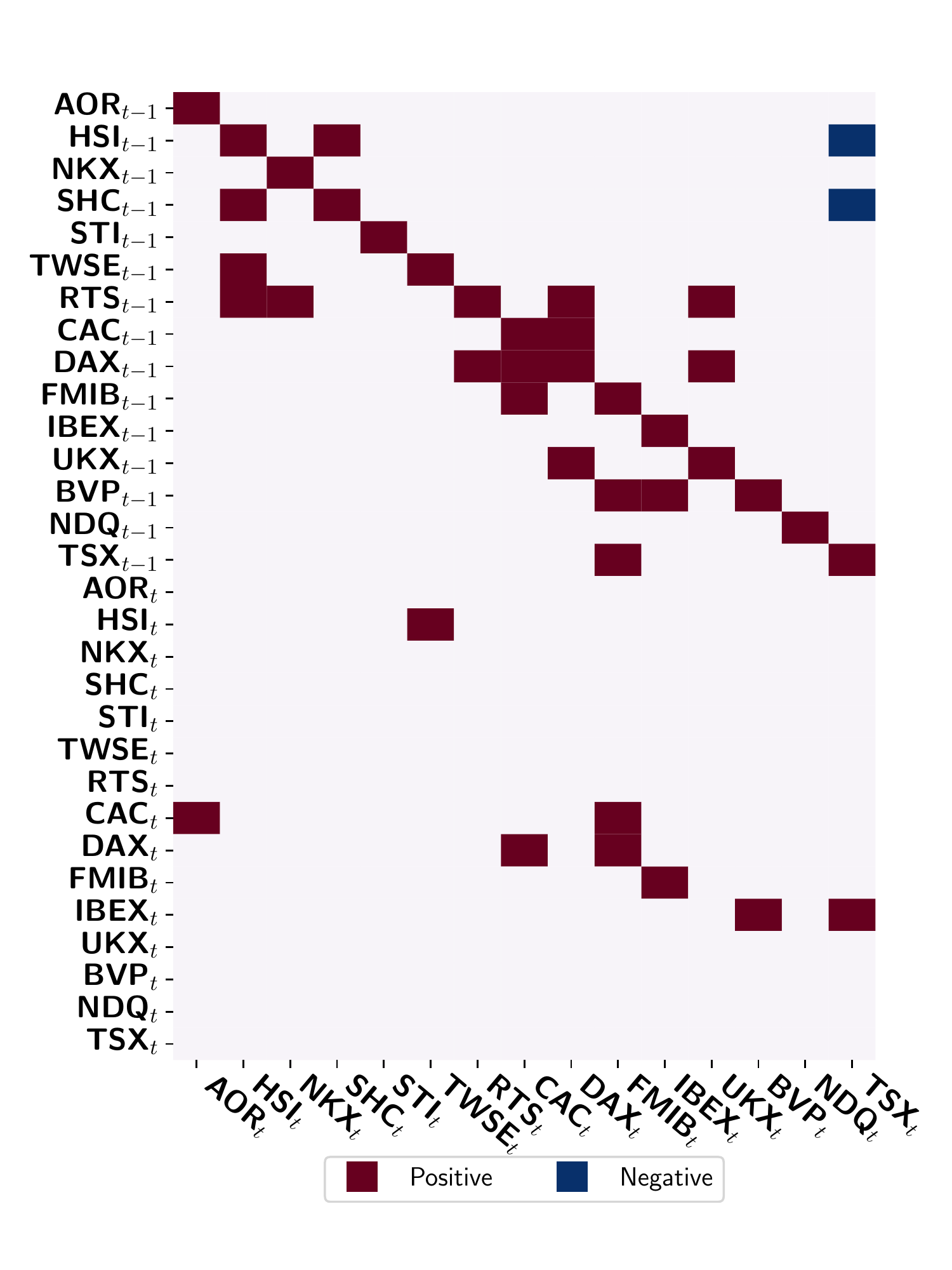}
        \caption{$\bar{c}=0.05$}
    \end{subfigure}
    \hfill
    \begin{subfigure}[b]{.32\textwidth}
        \centering
        \includegraphics[width=\textwidth]{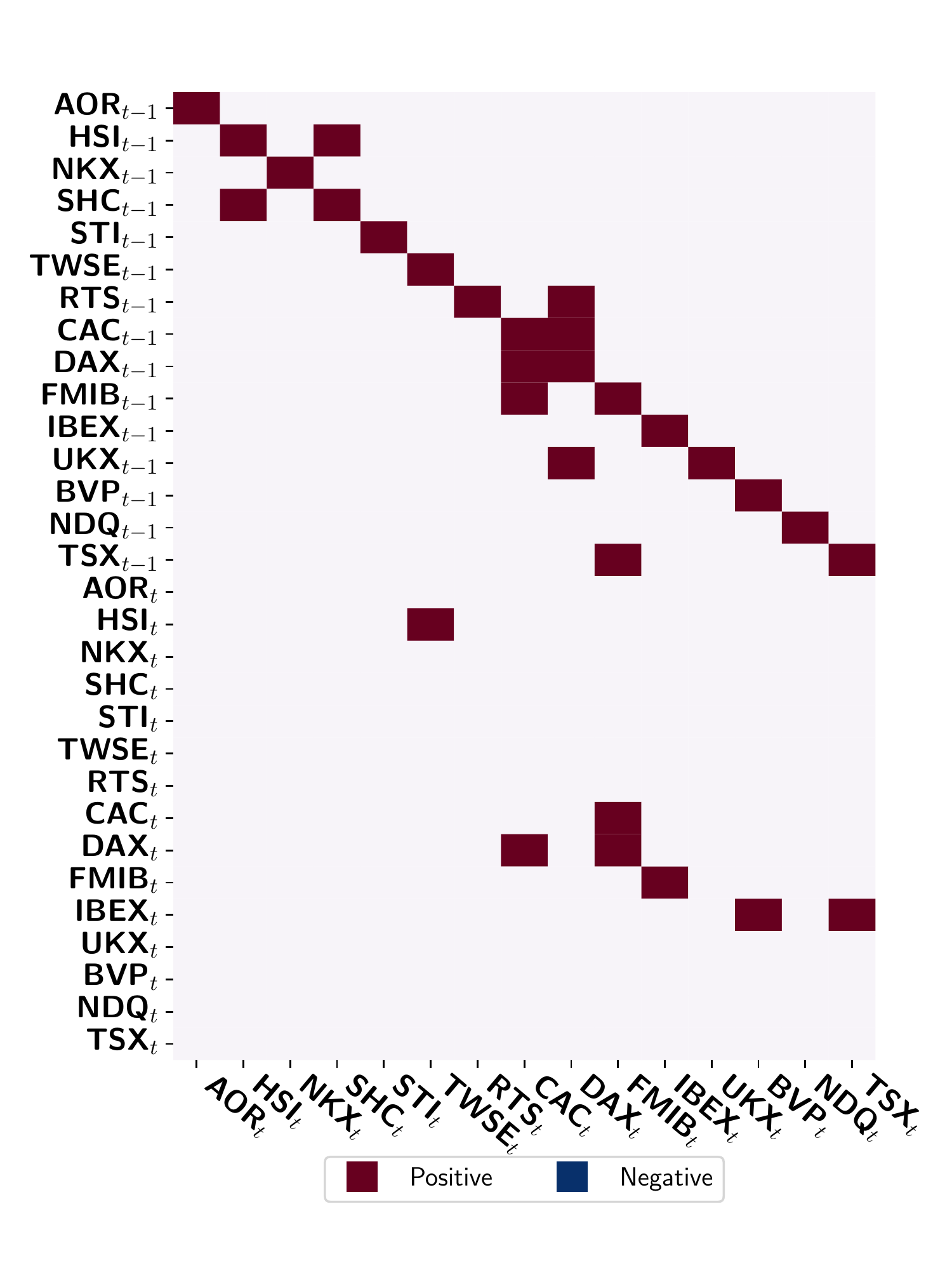}
        \caption{$\bar{c}=0.1$}
    \end{subfigure}
    \vfill
    \begin{subfigure}[b]{.32\textwidth}
        \centering
        \includegraphics[width=\textwidth]{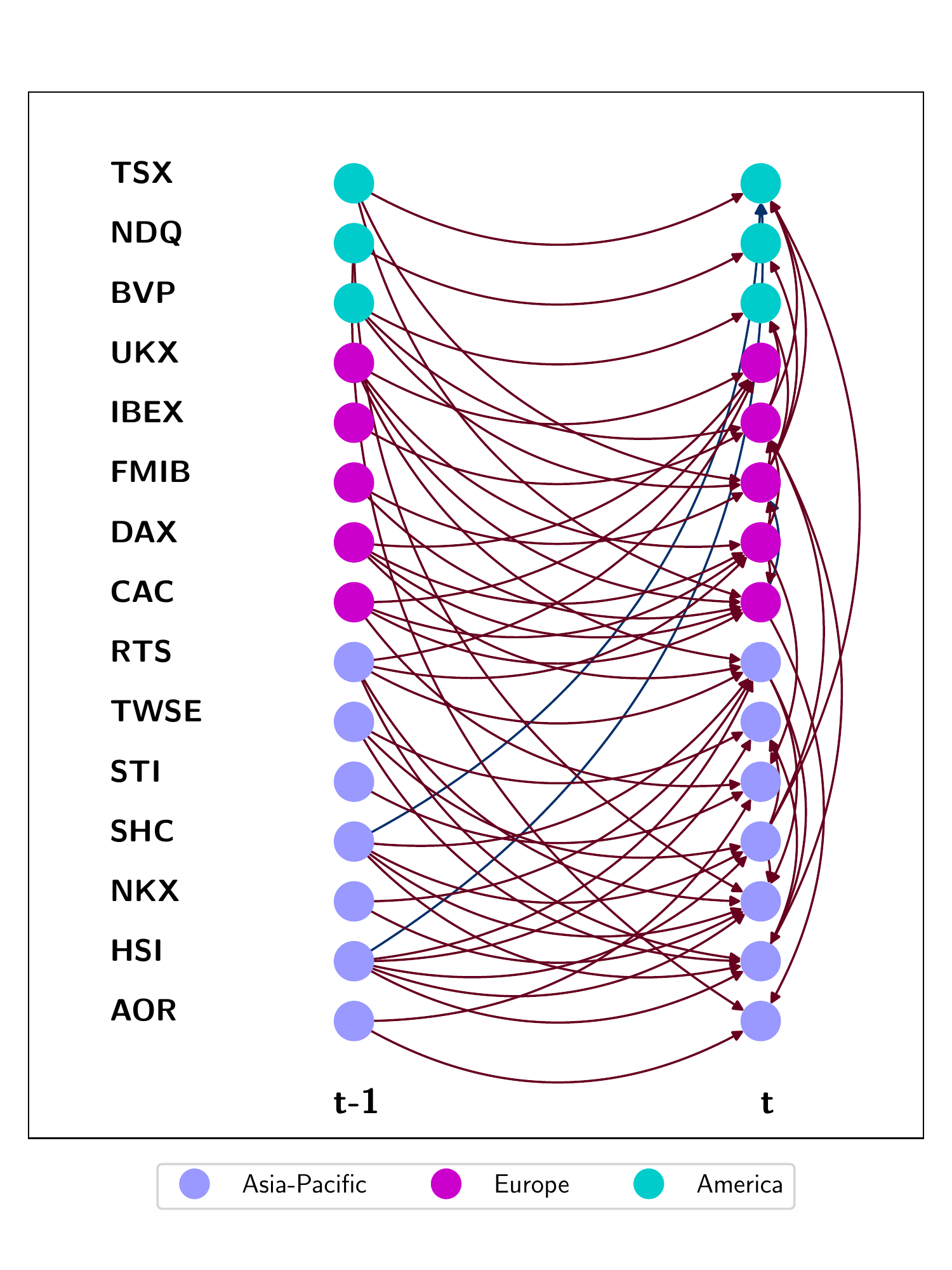}
        \caption{$\bar{c}=0.01$}
    \end{subfigure}
    \hfill
    \begin{subfigure}[b]{.32\textwidth}
        \centering
        \includegraphics[width=\textwidth]{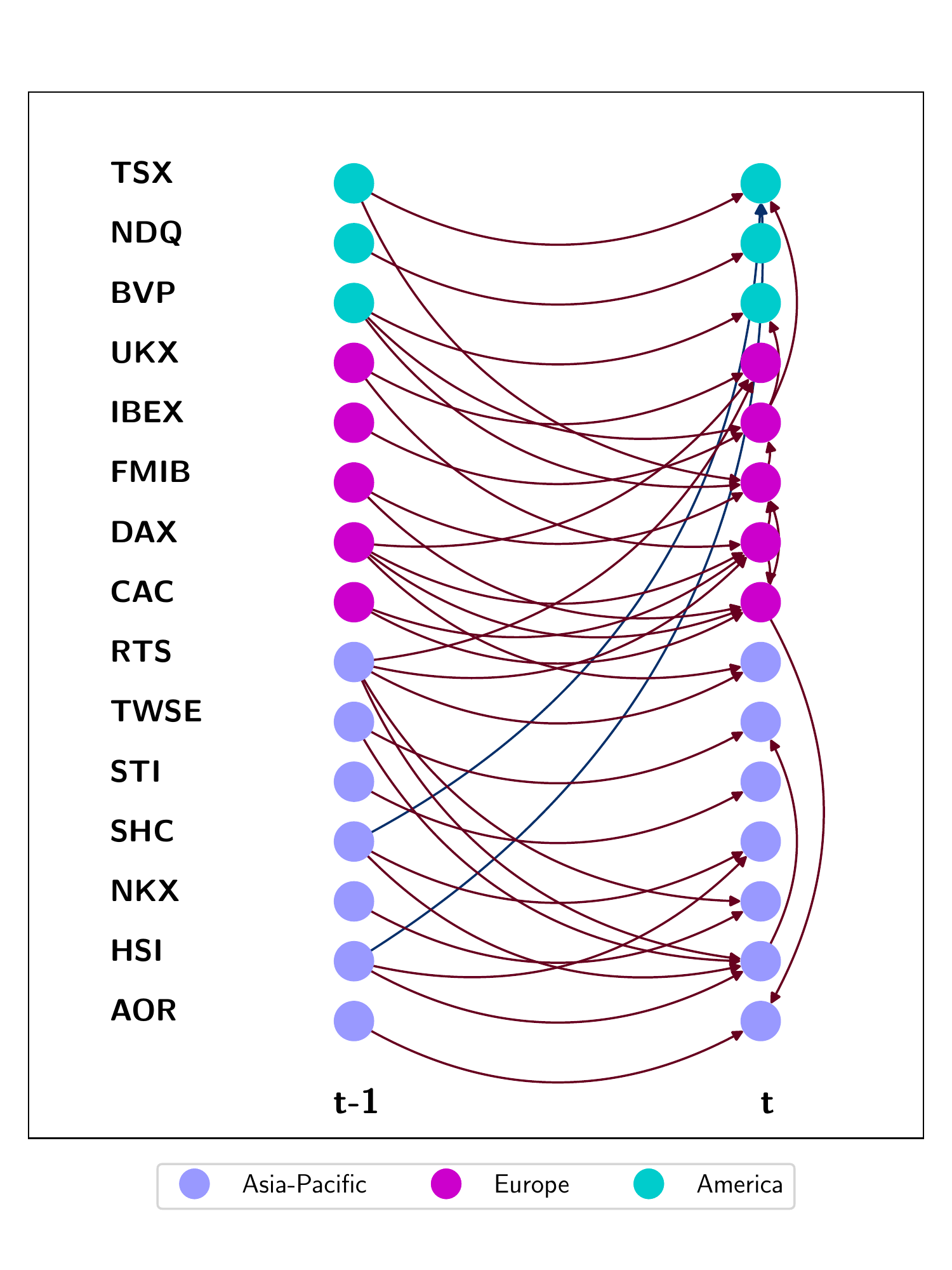}
        \caption{$\bar{c}=0.05$}
    \end{subfigure}
    \hfill
    \begin{subfigure}[b]{.32\textwidth}
        \centering
        \includegraphics[width=\textwidth]{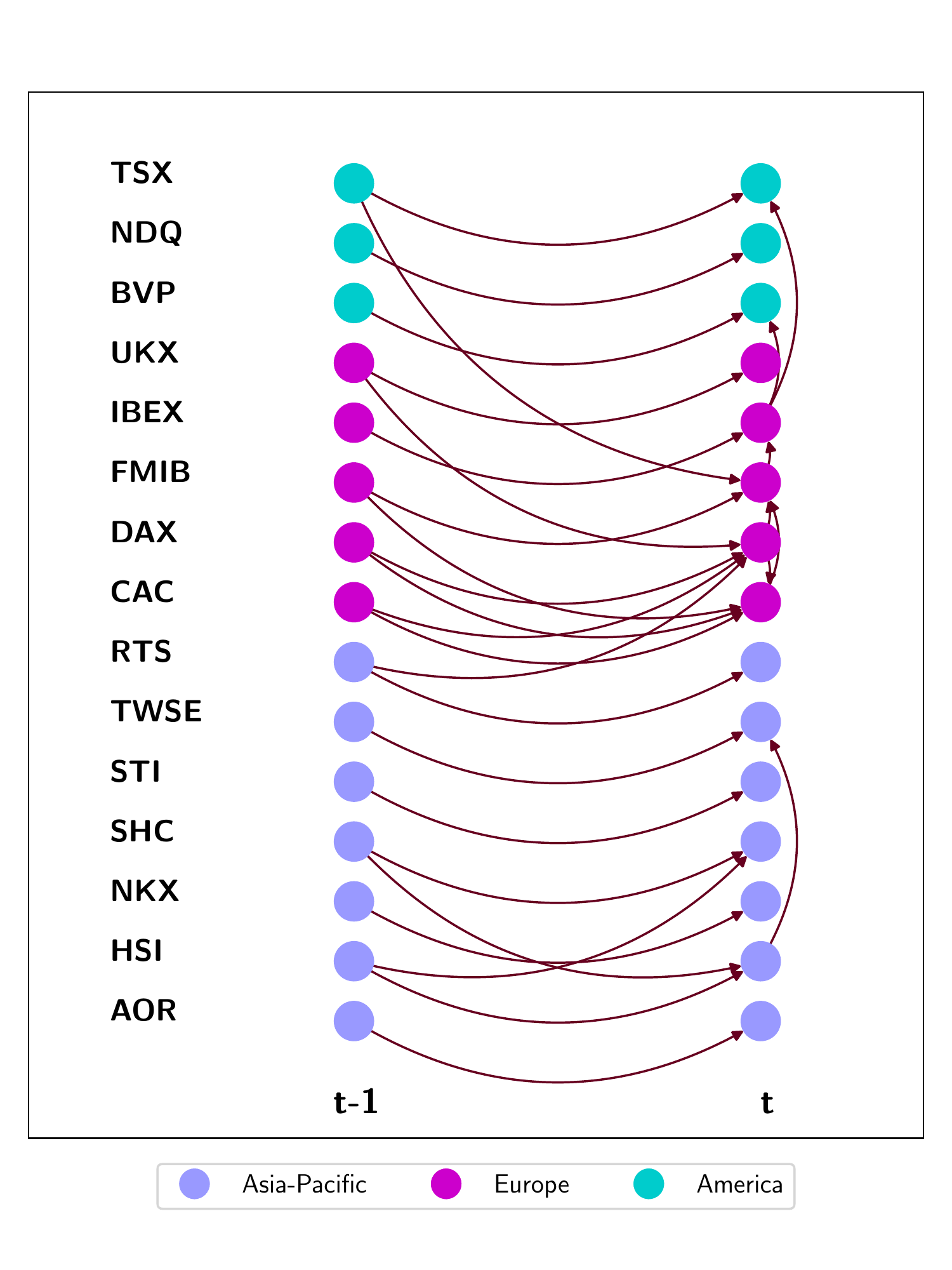}
        \caption{$\bar{c}=0.1$}
    \end{subfigure}
    \caption{Highly persistent causal matrix and corresponding SSCG for three different values of $\bar{c}$.}
    \label{fig:hp_time}
\end{figure}

\subsubsection{Temporal causal analysis}\label{sec:resultsTS}

First let us focus on learning causal graphs from time series using the aforementioned SS-CASTLE method. Then, \Cref{fig:hp_time} shows the signed causal matrix and the corresponding SSCG made up of persistent coefficients, where for readability reasons we only report three values of $\bar{c}$. In particular, the matrix representations in  \Cref{fig:hp_time} (a)-(c) are such that the rows represent the parents (sorted according to the timestamp), whereas the columns refer to the caused nodes. In this case, based on BIC criterion, we set $L=1$. Thus, the upper block of each matrix in \Cref{fig:hp_time} concerns lagged causal interactions, while the lower one is related to instantaneous causal effects. Each entry is red if the sign of the relation is positive; blue if negative and white if the edge is either absent or not persistent. Concerning the SSCG in  \Cref{fig:hp_time} (d)-(f), nodes are sorted and coloured according to the geographical area they belong to. Moreover, they are split based on the time lag. The rationale of edges color is the same as above. In addition, the greater the persistence of the causal relation, the thicker the corresponding arc. Looking at the causal matrices in \Cref{fig:hp_time}, we can notice how, as $\bar{c}$ grows from $0.01$ to $0.1$, a greater number of connections is pruned. In particular, arcs associated with negative causal relations get a weight lower than $0.1$ (in module); whereas, most of surviving edges correspond to autoregressive causal effects. Furthermore, we find denser causal connections among European and Asia-Pacific countries. At the same time, from the SSCGs in in \Cref{fig:hp_time}, we notice that persistent causal relationships between geographical areas are characterized by small weights. Overall, from this analysis, we cannot find nodes representing major risk drivers within the network, i.e., the considered equity markets show a similar number of outgoing arcs (i.e., out-degree).

\begin{figure}[p]
    \centering
    \begin{subfigure}[b]{.48\textwidth}
        \centering
        \includegraphics[width=\textwidth]{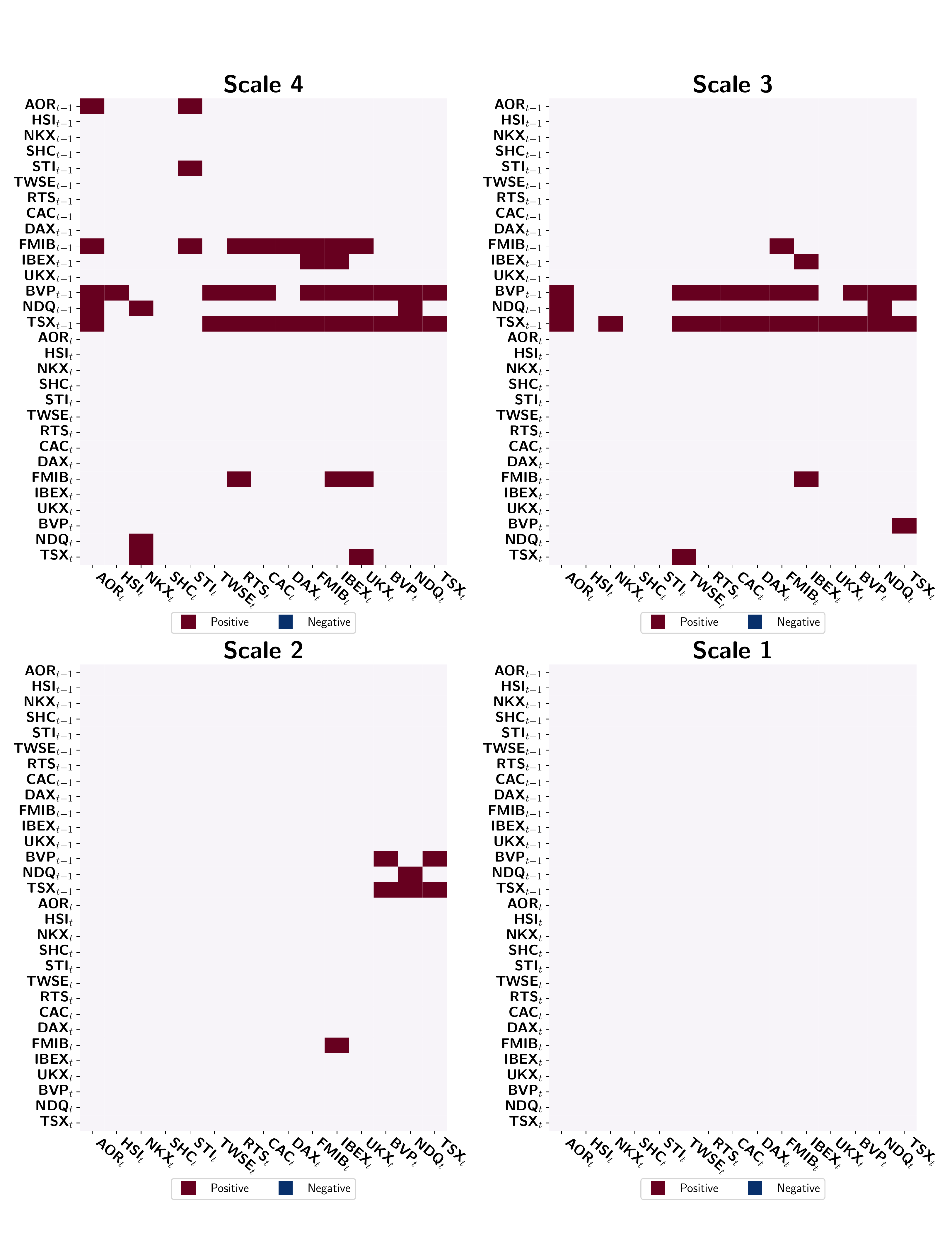}
        \caption{$\bar{c}=0.01$}
    \end{subfigure}
    \hfill
    \begin{subfigure}[b]{.48\textwidth}
        \centering
        \includegraphics[width=\textwidth]{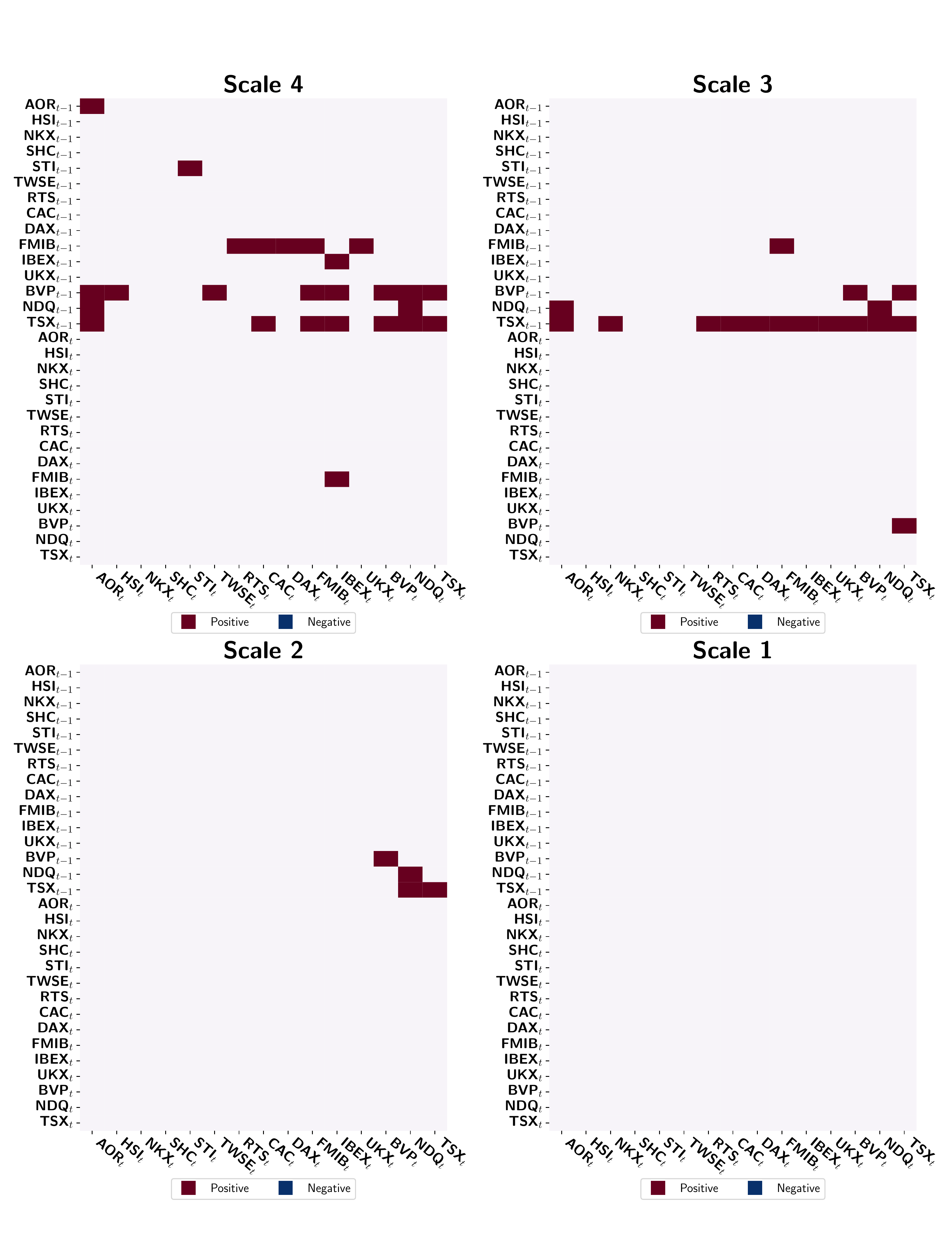}
        \caption{$\bar{c}=0.05$}
    \end{subfigure}
    \hfill
    \begin{subfigure}[b]{.48\textwidth}
        \centering
        \includegraphics[width=\textwidth]{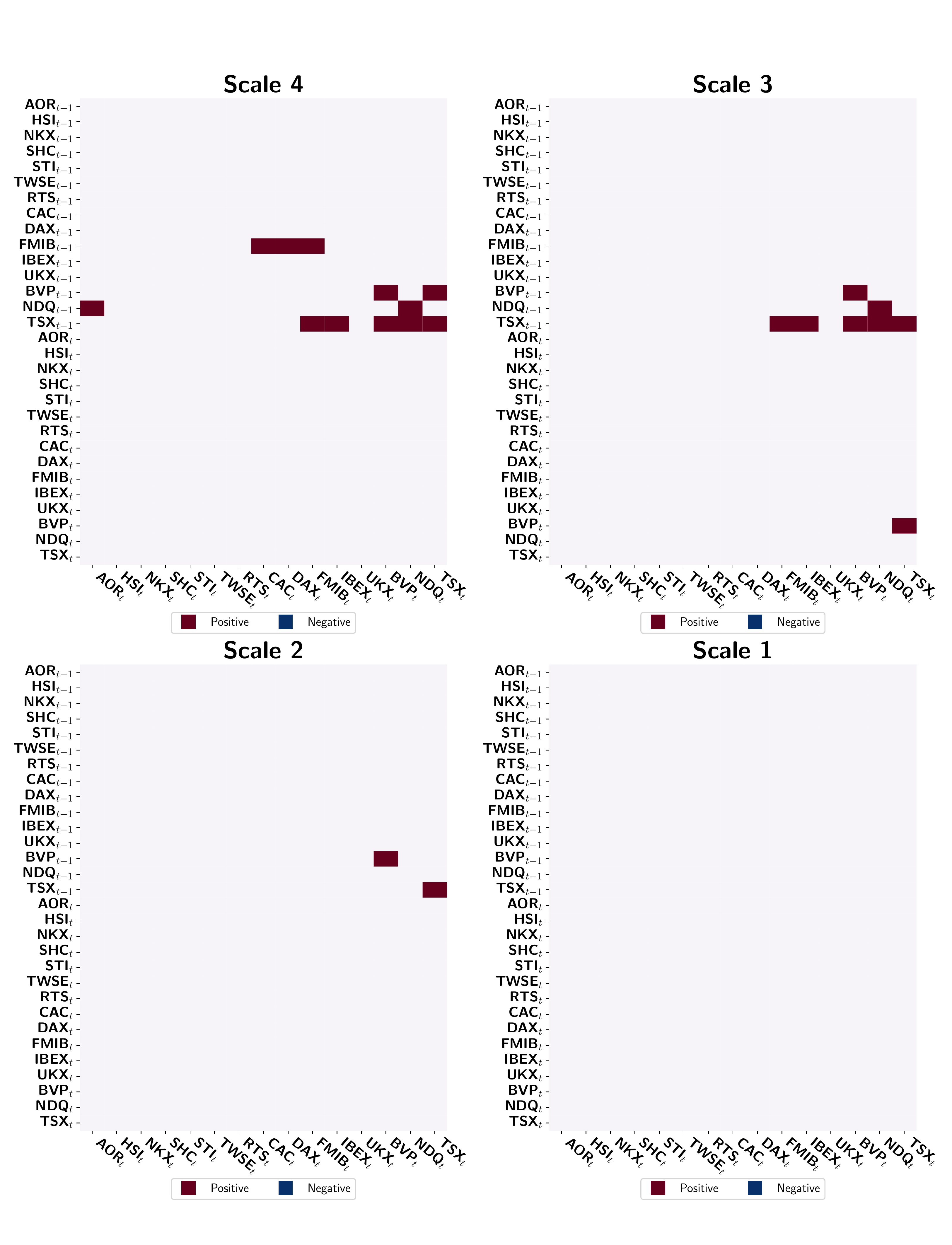}
        \caption{$\bar{c}=0.1$}
    \end{subfigure}
    \caption{Highly persistent multiscale causal matrix for three different values of $\bar{c}$.}
    \label{fig:hp_multiscale}
\end{figure}

\subsubsection{Multiscale causal analysis}\label{sec:resultsMS}

\begin{figure}[p]
    \centering
    \begin{subfigure}[b]{.48\textwidth}
        \centering
        \includegraphics[width=\textwidth]{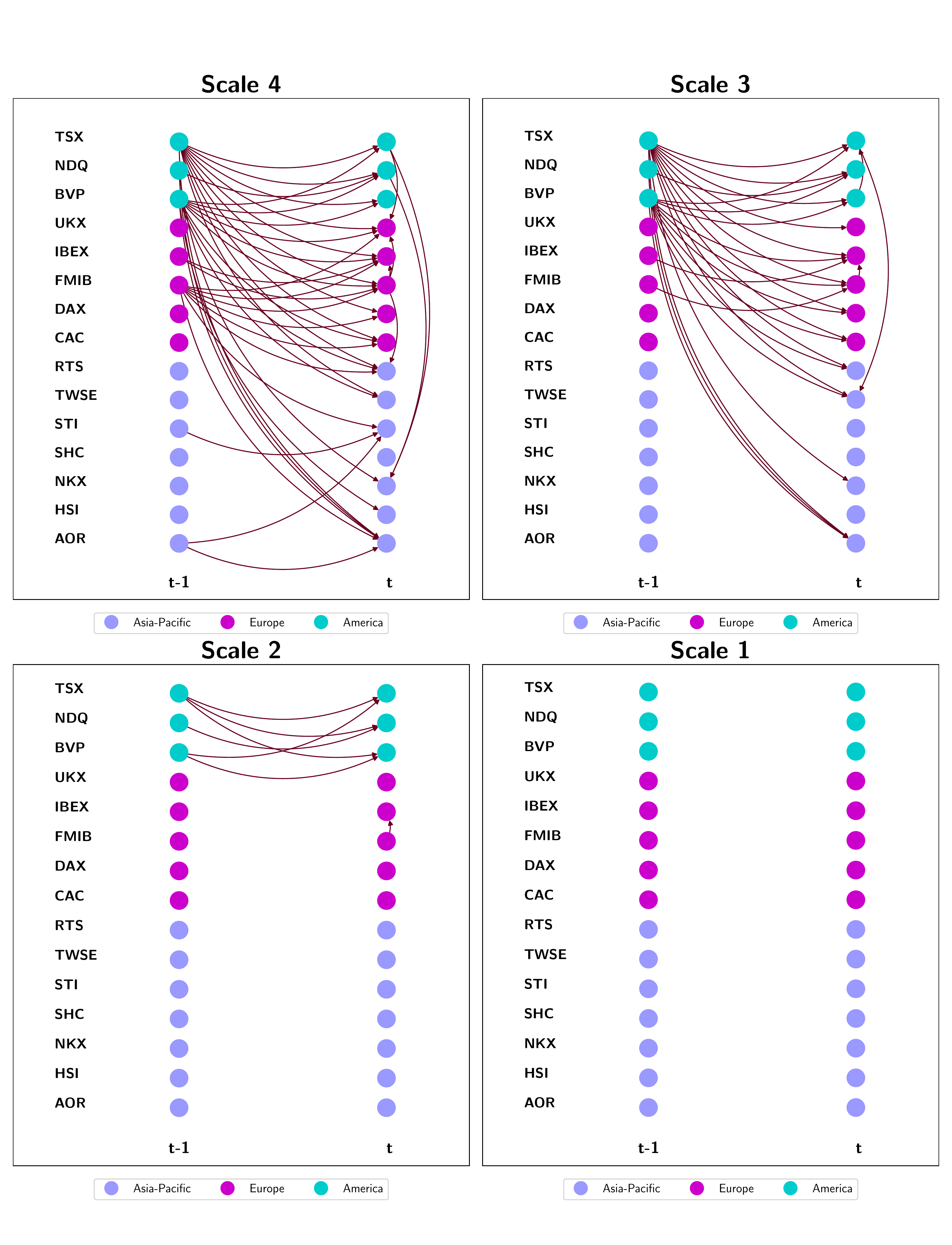}
        \caption{$\bar{c}=0.01$}
    \end{subfigure}
    \hfill
    \begin{subfigure}[b]{.48\textwidth}
        \centering
        \includegraphics[width=\textwidth]{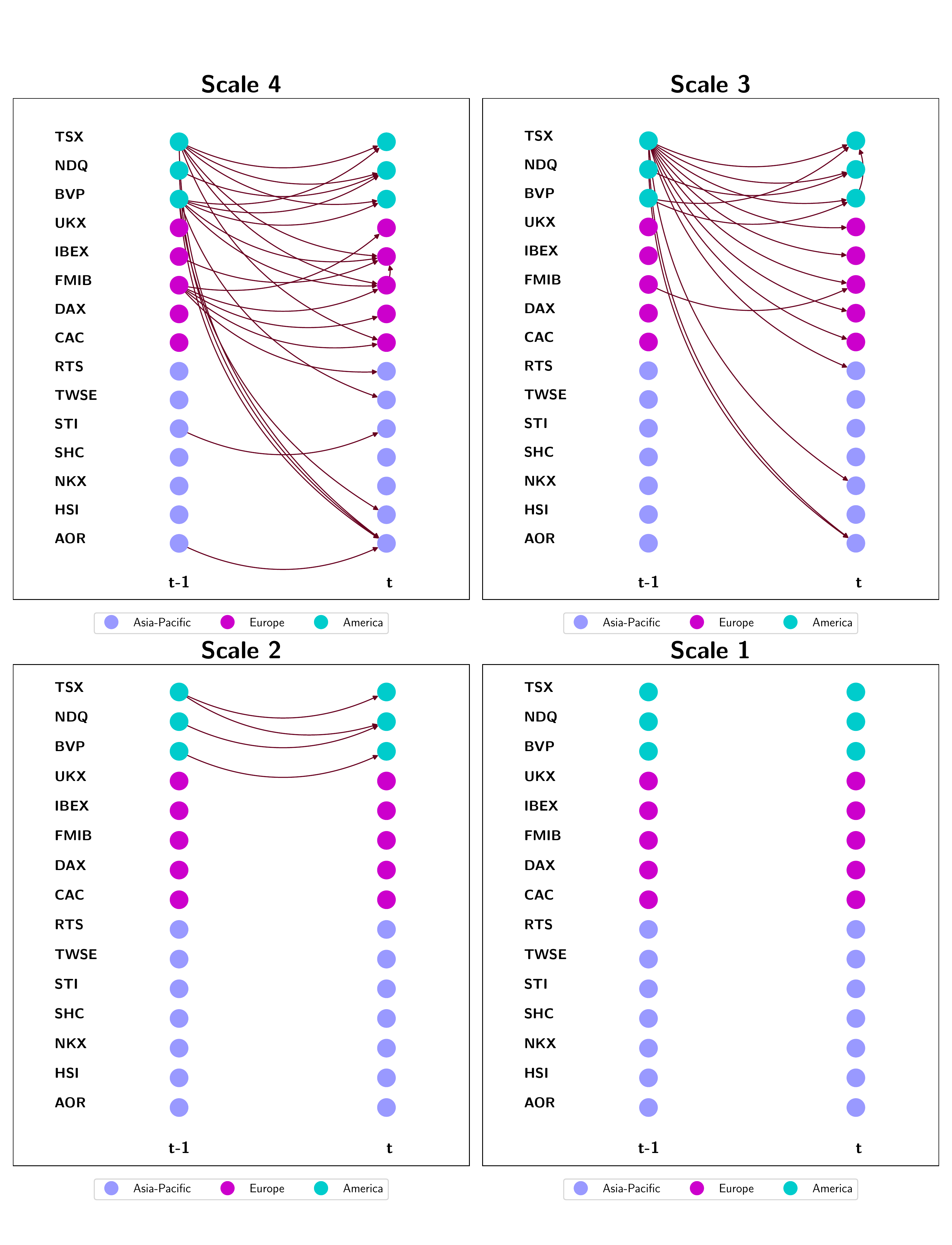}
        \caption{$\bar{c}=0.05$}
    \end{subfigure}
    \hfill
    \begin{subfigure}[b]{.48\textwidth}
        \centering
        \includegraphics[width=\textwidth]{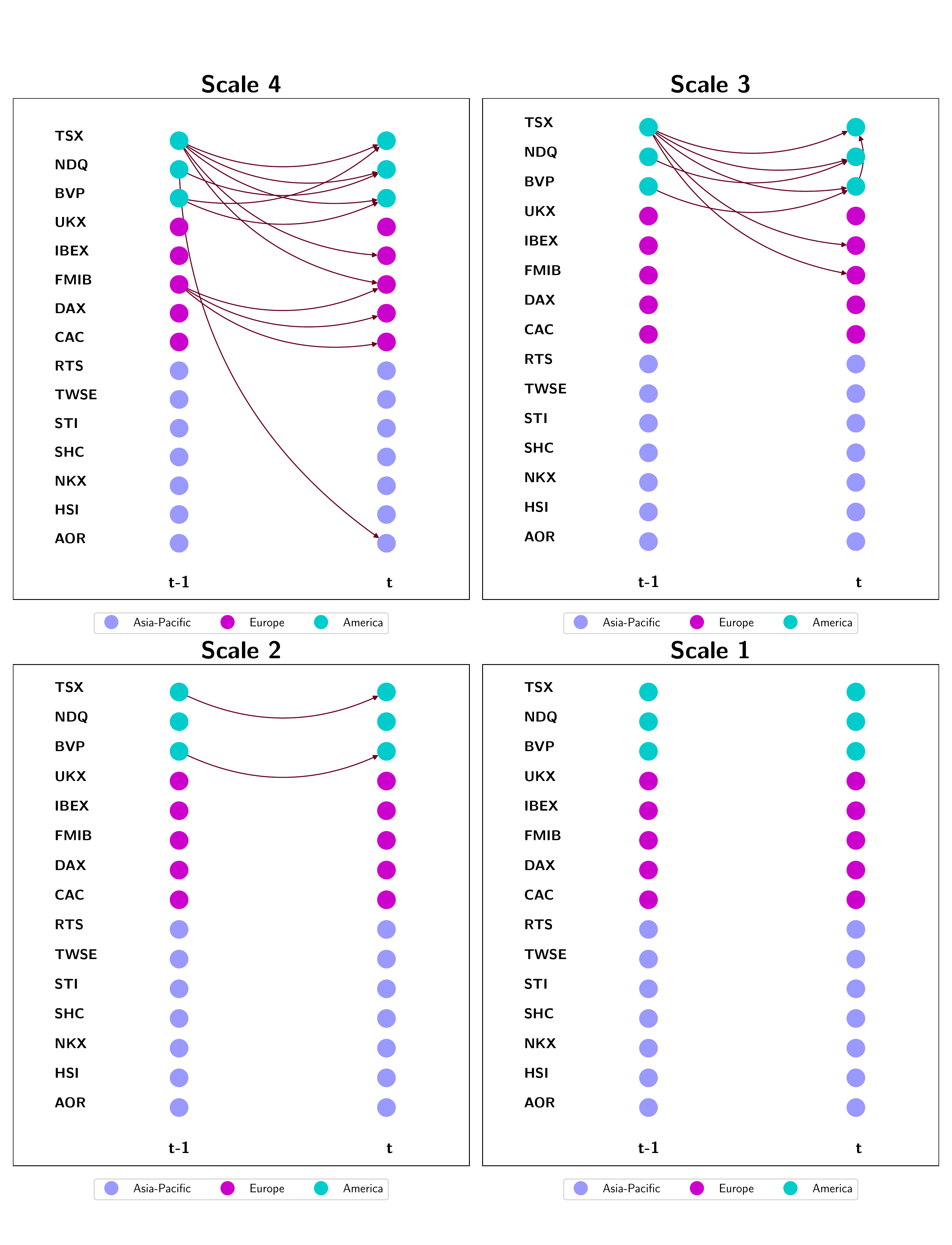}
        \caption{$\bar{c}=0.1$}
    \end{subfigure}
    \caption{Highly persistent MSCG for three different values of $\bar{c}$.}
    \label{fig:hp_multiscale_nets}
\end{figure}

We now focus on multiscale causal analysis, where we set the maximum number of scales equal to $D=4$, in accordance with the length of our data set, i.e., $T=336$ observations (see \cref{sec:MS-CASTLE}). Moreover, similarly to \citet{ren2021multiscale}, we use Daubechies least asymmetric wavelets with filter length equal to 8. Then, \Cref{fig:hp_multiscale} illustrates the highly persistent multiscale causal matrices obtained using the proposed MS-CASTLE method, for the same values of $\bar{c}$ analyzed above. For the sake of readability, for each threshold, we show separately the diagonal blocks of $\mathbf{\bar{W}}$ corresponding to different scales, i.e., the only elements of $\mathbf{\bar{W}}$ that can be different from zero in Equation \eqref{eq:MSCG} (since no interaction among scales actually takes place). From \Cref{fig:hp_multiscale}, we first notice that causal representations at different scales show a diverse level of sparsity and, furthermore, persistent causal relations assume only positive values. More in details, causal interactions are denser at mid-term scales (i.e., 3 and 4, which correspond to 8-16 and 16-32 days, respectively). On the contrary, causal effects turn out to be not persistent at scale 1, which represents a time resolution of 2-4 days. By looking across $\bar{c}$ values, we see that the strongest persistent connections appear at scale 3 and 4 and that the majority is lagged. Indeed, most instantaneous relations are associated to weights lower than 0.05 (in module). Finally, \Cref{fig:hp_multiscale_nets} depicts the corresponding MSCG for the aforementioned thresholds. From \Cref{fig:hp_multiscale_nets}, we can notice the following behaviors: i) apart from Australia, Asia-Pacific countries are isolated for $\bar{c}>0.05$; ii) the markets that drive the risk within the network are Brazil, Canada and Italy. The latter finding can be understood by looking at the number of nonzero entries per markets across columns, representing the out-degree of each node. More in details, the impacts of Brazil and Canada spread across all geographical areas, while Italy mainly drive the risk within the Eurozone. Finally, by looking at the multiscale matrix in \Cref{fig:hp_multiscale} corresponding to $\bar{c}=0.1$, we notice that US displays persistent lagged connections as well. 

\subsubsection{Comparison between temporal and multiscale analysis}\label{sec:discussion}

The results presented in \Cref{sec:results} illustrate that, in case of complex systems such as financial markets, temporal and multiscale analysis might lead to very different conclusions.
First of all, the inferred SSCG indicates a persistent causal structure at daily frequency, where the strongest connections are autoregressive lagged causal relations. On the contrary, empowered by information concerning the variation of the original signal at different scales, the MSCG shows that causal structures persist at mid-term scales (i.e., 3 and 4), while at short-term scale causal connections are absent. Thus, we can conclude that, in our case, the mere application of a multiscale-agnostic model leads to a noisy estimate of the causal structure, in which many of the relationships do not persist when decomposing the signal into different temporal resolutions.  In addition, in MSCG we do not observe negative causal coefficients as in case of SSCG, which are somehow difficult to justify during the considered period, since they indicate that an increase (decrease) in the volatility of a certain equity market causes a decrease (increase) in that of another market.

Finally, and most importantly, multiscale causal analysis allows us to identify the major risk drivers within the network of equity markets during covid-19 pandemic, i.e., Brazil, Canada and Italy. Interestingly, the US stock market, shows only an impact on Japan and Australia, together with an autoregressive effect. In particular, the importance of Canada within the network of stock markets has been underlined by \citet{ren2021multiscale} as well, who conducted a study in terms of partial correlation networks. However, since we deal with causation, our result has a stronger implications with respect to the aforementioned work. With regards to Brazil, we see that the corresponding stock index shows the highest volatility (see \Cref{sec:real-data}), and that its strongest connections (greater than $0.1$) are within the American area. Last, but not least, Italy has a high impact within the European area. 

\section{Conclusions and Future Research Directions}\label{sec:conclusion}

In this paper we have proposed a novel method to estimate the structure of linear causal relationships at different time scales. 
By relying upon SWT and non-convex optimization, MS-CASTLE takes explicitly into consideration behaviors of the system at hand spanning at diverse time resolutions.
Differently from existing causal inference methods, MS-CASTLE looks for linear causal relationships among variations of input signals within multiple frequency bands.
We illustrate that the multiscale-agnostic version of MS-CASTLE, named SS-CASTLE, improves in terms of computational efficiency, performance and robustness over the state of the art.

The study of the risk of 15 global equity markets, during covid-19 pandemic, shows that MS-CASTLE is able to provide useful information about the scales at which causal interactions occur (mid-term scales) and to identify major risk drivers within the system (Brazil, Canada and Italy). 
We highlight that the obtained results must be framed in the period of coronavirus outbreak. 
Our choice was conscious: given the nonstationary nature of financial markets~(\citealt{schmitt2013non}), we focused on a narrow period dominated by the pandemic emergency.
Thus, the use of different time windows may lead to the estimation of a different multiscale causal structure.

This observation highlights the need to work on the development of causal inference algorithms capable of handling both the multiscale nature of the analysed system and the nonstationarity of the underlying causal structure.
In this context, the application of gaussian processes to model the time dependence of the causal structure has lead to some advances (\citealt{huang2015identification}).
 
In addition, the proposed model considers only linear causal relationships: generalisation to nonlinear interactions represents further future work.
Here, kernel methods (\citealt{shen2016nonlinear}) and more recently non-linear ICA (\citealt{monti2020causal}) has been used to tackle the estimation task.
However, previous works only refer to the single scale case.

In this work we did not consider possible inter-scale cause-effect mechanisms.
However, we do not exclude that behaviors of signals at higher frequencies may impact those at lower frequencies and vice versa.
So, investigating the existence of such causal relationships represents an interesting future research direction.

Last but not least, it would be useful to develop a causal generative model that includes, as a special case, stationary models at a single temporal resolution, and that allows the modelling of multiscale, non-stationary and possibly nonlinear causal dynamics.
In this way, it would be possible to identify a common way of generating synthetic data sets on which to test the performance of causal structure learning algorithms.
Most importantly, such data sets could be generated to reflect the main features of time series from different fields of study, such as finance, neuroscience and climatology. 
To this end, it could be useful to exploit the mathematical modeling of Multivariate Locally Stationary Processes (\citealt{park2014estimating}).

Finally, we emphasize that the results of the case study show how MS-CASTLE can be used to support portfolio risk management. Indeed, depending on their investment horizon, investors could use the proposed methodology to make risk-aware decisions regarding their portfolios, from a causal perspective and without any prior assumption about the scale of analysis.






\vskip 0.2in
\bibliography{bibliography}

\end{document}